%% file: acl_latex.tex
\PassOptionsToPackage{usenames,dvipsnames,svgnames,table}{xcolor}
\PassOptionsToPackage{hyphens}{url}
\documentclass[11pt]{article}

\newif\ifpreprint
\input{variant}
\ifpreprint
  \usepackage[preprint]{acl}
\else
  \usepackage[review]{acl}
\fi

\usepackage{times}
\usepackage{latexsym}
\usepackage[T1]{fontenc}
\usepackage[utf8]{inputenc}
\usepackage{microtype}
\usepackage{inconsolata}
\usepackage{graphicx}
\usepackage{booktabs}
\usepackage{array}
\usepackage{longtable}
\usepackage{multirow}
\usepackage{amssymb}
\usepackage{amsmath}
\usepackage{xspace}
\usepackage{tikz}
\usetikzlibrary{positioning,calc,arrows.meta,fit}
\definecolor{pipeBlue}{HTML}{1864AB}
\definecolor{pipeBlueFill}{HTML}{E7F5FF}
\definecolor{pipeGray}{HTML}{6C757D}
\definecolor{pipeGrayFill}{HTML}{F1F3F5}

\usepackage{xcolor}
\usepackage[textsize=tiny]{todonotes}
\makeatletter
\newcommand*\iftodonotes{\if@todonotes@disabled\expandafter\@secondoftwo\else\expandafter\@firstoftwo\fi}
\makeatother

\newcommand{\jev}{Jev\xspace}
\newcommand{\jevopt}{\texttt{jev-opt}\xspace}
\newcommand{\suite}{TDM-Gauntlet}
\newcommand{\Uponacceptance}{Upon acceptance we}
\newcommand{\uponacceptance}{ upon acceptance}
\newcommand{\releaseat}{}

\newcommand{\nOtherVendors}{three}
\def\unsign#1{\expandafter\unsignx#1\relax}
\def\unsignx#1#2\relax{\ifx+#1#2\else#1#2\fi}
\def\absnum#1{\expandafter\absnumx#1\relax}
\def\absnumx#1#2\relax{\ifx+#1#2\else\ifx-#1#2\else#1#2\fi\fi}

\input{generated/numbers}

\input{generated/coverage}
\input{generated/framing}

\input{generated/confidence}

\input{generated/causes}

\input{generated/review}
\input{generated/budget}

\input{generated/budget_baseline}

\input{generated/offline}

\input{generated/frozen}

\input{generated/conventions}

\input{generated/clef}

\input{generated/offline_clef}

\input{generated/clef_rq3}

\input{generated/cal_refit}
\input{generated/luna}

\input{generated/luna_rq3}

\input{generated/frozen_seeds}

\input{generated/rq3_gap_interval}
\input{generated/label_curve}

\input{generated/h100_round2}

\input{generated/lora}

\input{generated/msd1}

\input{generated/msd1_rq3}

\input{generated/answer}

\title{Do We Still Need Fine-Tuned Classifiers?\\Swapping Typed Decision Models into \specCore{} ACL-Venue Papers}
\hypersetup{
  pdftitle={Do We Still Need Fine-Tuned Classifiers? Swapping Typed Decision Models into \specCore{} ACL-Venue Papers},
  pdfkeywords={Typed Decision Models; decision models; System One models; System 1 models; large language models; LLM; zero-shot classification;
    text classification; fine-tuned classifiers; fine-tuning; benchmark; prompt optimization; calibration; LoRA fine-tuning;
    annotation conventions; label efficiency; GEPA; TDM-Gauntlet; Jev; TypeSafe; Clef; Cloudflare; GPT-6 Luna Decisions; OpenAI; Microsoft-Decision-1; Qwen3.8-27B}}

\ifpreprint
  \input{preprint}
\else
  \author{Anonymous}
\fi

\begin{document}
\maketitle

\input{generated/body_numbers}
\input{generated/hero_numbers}
\input{generated/hero_steps}

\begin{abstract}
Typed Decision Models such as TypeSafe's \jev{}, also called System One or System~1 models, answer typed
questions about an input with a probability for each option, and they need no training. Many NLP papers
still fine-tune a task-specific classifier, and prompted large language models (LLMs) trail such classifiers
on most text classification benchmarks. We ask how many of these fine-tuned classifiers a Typed Decision
Model can replace under each paper's own evaluation, and what is needed where it cannot. We swap the
predictors of \specCore{} ACL-venue papers for
\jev{}, keep each paper's splits, label definitions and metric, and check the results with Cloudflare's Clef,
Microsoft-Decision-1 and OpenAI's GPT-6 Luna Decisions. The resulting \nDatasets{} English test sets, with
their questions and published scores, form the \suite{} benchmark. With a
zero-shot prompt, \jev{} directly replaced the paper's predictor on \nReplaces{} of the \nDatasets{} test sets,
mostly single judgments with plain labels, but rarely where the paper's method was trained on its
labels. Where \jev{} scored lower, we most often attributed the shortfall to an annotation convention
that the label name does not state. With \optTrainN{} training labels plus validation and no user-managed
GPU, calibration and GEPA-style prompt optimization closed a median \atOptGap\% of
the gap to the paper's method, about twice what off-the-shelf GEPA closed, for a median \$\frCostFqMedian{} per search. On the same labels, fine-tuned
encoders such as RoBERTa and DeBERTa closed \atEncGap\%, and LoRA fine-tuning of the
open LLM Qwen3.8-27B closed \atLoraGap\%. Fine-tuning remains the most accurate option, but a
zero-shot prompt is often enough for plain labels, and prompt optimization closes most of the remaining gap.
\end{abstract}

\input{sections/introduction}
\input{sections/setup}
\input{sections/results}
\input{sections/optimizing_decision_}
\input{sections/when}
\input{sections/related_work}
\input{sections/conclusion}
\input{sections/limitations}

\bibliography{papers25,refs,data}

\appendix
\input{sections/appendix}

\end{document}

%% file: variant.tex
\preprinttrue

%% file: generated/numbers.tex
\newcommand{\nDatasets}{46}
\newcommand{\nReplaces}{12}

\newcommand{\nPapersScored}{23}

\newcommand{\totalCost}{\$176}

\newcommand{\specInScope}{18,201}

\newcommand{\specFrom}{2024-01-01}
\newcommand{\specCutoff}{2026-09-26}

\newcommand{\specStress}{15}
\newcommand{\specCore}{24}

%% file: generated/framing.tex
\newcommand{\frAblATBDense}{65.5}
\newcommand{\frAblPTBDense}{68.9}
\newcommand{\frAblDTBDense}{+3.4}

\newcommand{\frAblAEZStance}{64.5}
\newcommand{\frAblPEZStance}{79.8}
\newcommand{\frAblDEZStance}{+15.3}

\newcommand{\frAblAImplicitHate}{57.0}
\newcommand{\frAblPImplicitHate}{62.4}
\newcommand{\frAblDImplicitHate}{+5.3}

\newcommand{\frAblAEditIntent}{67.2}
\newcommand{\frAblPEditIntent}{72.4}
\newcommand{\frAblDEditIntent}{+5.3}

\newcommand{\frAblAINLI}{79.0}
\newcommand{\frAblPINLI}{84.4}
\newcommand{\frAblDINLI}{+5.4}

\newcommand{\frAblADressContent}{33.5}
\newcommand{\frAblPDressContent}{59.7}
\newcommand{\frAblDDressContent}{+26.2}

\newcommand{\frAblADressOrganization}{31.9}
\newcommand{\frAblPDressOrganization}{55.0}
\newcommand{\frAblDDressOrganization}{+23.1}

\newcommand{\frAblADressLanguage}{45.2}
\newcommand{\frAblPDressLanguage}{60.4}
\newcommand{\frAblDDressLanguage}{+15.1}

\newcommand{\frAblNPos}{8}

\newcommand{\frAblNTasks}{8}

\newcommand{\frConvN}{16}
\newcommand{\frConvTotal}{24}

\newcommand{\frCostFqMedian}{1.66}

%% file: generated/confidence.tex
\newcommand{\cfSelNComparable}{13}

\newcommand{\cfSelNFull}{7}
\newcommand{\cfSelHalvings}{50}

\newcommand{\cfSelLttCovInli}{41}

\newcommand{\cfSelLttCovPinoseWebq}{87}

\newcommand{\cfRoEceZero}{0.206}

\newcommand{\cfRoEceTs}{0.048}

\newcommand{\cfRerunN}{6,324}
\newcommand{\cfRerunFlips}{173}

%% file: generated/causes.tex
\newcommand{\csN}{34}
\newcommand{\csNPapers}{17}

\newcommand{\csAgree}{27}
\newcommand{\csKappa}{0.67}
\newcommand{\csKappaLo}{0.44}
\newcommand{\csKappaHi}{0.86}
\newcommand{\csConv}{18}
\newcommand{\csConvPapers}{12}

\newcommand{\csLarge}{4}

\newcommand{\csCoarse}{1}

\newcommand{\csOther}{11}

%% file: generated/budget.tex
\newcommand{\optTrainN}{500}

\newcommand{\optValMin}{376}
\newcommand{\optValMax}{1,000}
\newcommand{\optLabelsMin}{876}
\newcommand{\optLabelsMax}{1,500}

%% file: generated/budget_baseline.tex
\newcommand{\bbPooledAllDiff}{+2.0}

\newcommand{\bbPooledNoTbdDiff}{-0.4}

\newcommand{\bbNBestFt}{12}
\newcommand{\bbNSets}{13}

\newcommand{\bbNUnclear}{nine}

\newcommand{\bbTbdFtMin}{23.1}
\newcommand{\bbTbdFtMax}{41.5}

\newcommand{\bbNBestReachPaper}{three}

\newcommand{\bbEqAllLo}{+0.6}
\newcommand{\bbEqAllHi}{+3.3}

\newcommand{\bbEqNoTbdLo}{-1.6}
\newcommand{\bbEqNoTbdHi}{+0.9}

\newcommand{\bbPipelineSigSets}{TB-Dense, EZ-STANCE A, MATRES}
\newcommand{\bbBaselineSigSets}{INLI}

%% file: generated/offline.tex
\newcommand{\ovRqOneB}{10,000}

\newcommand{\ovRqOneInconcl}{8}

\newcommand{\ovRqOneJevGeLlmZsTol}{22}
\newcommand{\ovRqOneLlmBestBelow}{22}
\newcommand{\ovRqOneLlmBestBetween}{12}

\newcommand{\ovRqOneLlmBestReplaces}{12}

\newcommand{\ovRqOneLlmZsBelow}{17}
\newcommand{\ovRqOneLlmZsBetween}{17}
\newcommand{\ovRqOneLlmZsReplaces}{12}

\newcommand{\ovRqOneMatchTrained}{4}

\newcommand{\ovRqOneNClustered}{12}
\newcommand{\ovRqOneNLlm}{30}

\newcommand{\ovRqOneNTrained}{38}

\newcommand{\ovRqOneNWidened}{15}

\newcommand{\ovRqOneParity}{0}

\newcommand{\ovRqOneSepAbove}{8}
\newcommand{\ovRqOneSepBelow}{30}
\newcommand{\ovRqOneTolPts}{1}

\newcommand{\ovRqThreePooledDiff}{+1.5}
\newcommand{\ovRqThreePooledHi}{+3.3}
\newcommand{\ovRqThreePooledLo}{$-$0.2}

%% file: generated/frozen.tex
\newcommand{\fzMatresPaper}{83.2}

\newcommand{\fzLifetoxPaper}{96.5}

\newcommand{\fzStorcPaper}{56.7}

\newcommand{\fzHatexplainPaper}{64.0}

\newcommand{\fzLatentPaper}{69.0}

\newcommand{\fzTbDenseZero}{54.3}
\newcommand{\fzTbDenseCal}{60.0}

\newcommand{\fzNewNSets}{5}

%% file: generated/conventions.tex
\newcommand{\cvPrograms}{20}

\newcommand{\cvSpans}{65}
\newcommand{\cvFrozenPos}{4}
\newcommand{\cvSeedPos}{16}

\newcommand{\cvDelAbs}{1.3}

\newcommand{\cvDelP}{0.092}

\newcommand{\cvCalAbs}{2.5}
\newcommand{\cvCalLo}{+0.9}
\newcommand{\cvCalHi}{+4.4}

\newcommand{\cvAZAbs}{14.4}

\newcommand{\cvJevCostAll}{\$9.21}

\newcommand{\cvPreDel}{+1.5}

%% file: generated/clef.tex
\newcommand{\clBetter}{11}
\newcommand{\clWithin}{15}
\newcommand{\clWorse}{19}
\newcommand{\clMedian}{$-$0.5}

\newcommand{\clConvN}{17}

\newcommand{\clFlashBetter}{8}
\newcommand{\clFlashWithin}{6}
\newcommand{\clFlashWorse}{31}
\newcommand{\clFlashMedian}{$-$4.4}

%% file: generated/offline_clef.tex
\newcommand{\ocNSets}{45}
\newcommand{\ocSameGroup}{39}

\newcommand{\ocRho}{0.87}

\newcommand{\ocFlashSameGroup}{35}

\newcommand{\ocPairNHigher}{9}
\newcommand{\ocPairNLower}{10}

%% file: generated/clef_rq3.tex
\newcommand{\clFlashPoolOpt}{+17.5}

\newcommand{\clFlashNholmOptJev}{0}
\newcommand{\clFlashNnegOptJev}{9}
\newcommand{\clFlashMedOptJev}{-4.6}

\newcommand{\clFlashGapAll}{67}
\newcommand{\clFlashGapCalAll}{32}

\newcommand{\clFlashEncAhead}{1}
\newcommand{\clFlashEncBehind}{6}

\newcommand{\clFlashEncMedDiff}{-3.7}

\newcommand{\clFlashGpuHours}{6.1}

\newcommand{\clClefNsets}{13}

\newcommand{\clClefPoolOpt}{+20.3}

\newcommand{\clClefNholmOptJev}{0}
\newcommand{\clClefNnegOptJev}{2}
\newcommand{\clClefMedOptJev}{-0.3}

\newcommand{\clClefGapAll}{77}
\newcommand{\clClefGapCalAll}{27}

\newcommand{\clClefEncAhead}{2}
\newcommand{\clClefEncBehind}{1}

\newcommand{\clClefEncMedDiff}{+0.2}

\newcommand{\clClefGpuHours}{11.0}

\newcommand{\clJevGapAll}{75}
\newcommand{\clJevGapCalAll}{23}

\newcommand{\clJevEncMedDiff}{+0.1}

%% file: generated/luna.tex
\newcommand{\luBetter}{6}
\newcommand{\luWithin}{14}
\newcommand{\luWorse}{25}
\newcommand{\luMedian}{$-$2.0}

\newcommand{\luSpend}{209.97}

\newcommand{\luSameGroup}{41}

\newcommand{\luRho}{0.96}

\newcommand{\luPairNHigher}{3}
\newcommand{\luPairNLower}{16}

%% file: generated/luna_rq3.tex
\newcommand{\luRqPoolOpt}{+14.6}

\newcommand{\luRqNholmOptJev}{0}
\newcommand{\luRqNnegOptJev}{7}
\newcommand{\luRqMedOptJev}{-4.4}
\newcommand{\luRqMedGapCal}{18}
\newcommand{\luRqMedGapOpt}{43}

\newcommand{\luRqEncAhead}{0}
\newcommand{\luRqEncBehind}{7}

\newcommand{\luRqEncMedDiff}{-3.3}

%% file: generated/frozen_seeds.tex
\newcommand{\fzsMatresSeedZero}{79.4}
\newcommand{\fzsMatresSeedOne}{78.9}
\newcommand{\fzsMatresSeedTwo}{79.6}
\newcommand{\fzsMatresMean}{79.3}

\newcommand{\fzsMatresGapMean}{65}

\newcommand{\fzsLifetoxSeedZero}{94.1}
\newcommand{\fzsLifetoxSeedOne}{91.3}
\newcommand{\fzsLifetoxSeedTwo}{93.8}
\newcommand{\fzsLifetoxMean}{93.1}

\newcommand{\fzsLifetoxGapMean}{81}

\newcommand{\fzsStorcSeedZero}{55.7}
\newcommand{\fzsStorcSeedOne}{56.6}
\newcommand{\fzsStorcSeedTwo}{54.5}
\newcommand{\fzsStorcMean}{55.6}

\newcommand{\fzsStorcGapMean}{72}

\newcommand{\fzsHatexplainSeedZero}{64.6}
\newcommand{\fzsHatexplainSeedOne}{64.7}
\newcommand{\fzsHatexplainSeedTwo}{64.2}
\newcommand{\fzsHatexplainMean}{64.5}

\newcommand{\fzsHatexplainGapMean}{106}

\newcommand{\fzsLatentSeedZero}{61.3}
\newcommand{\fzsLatentSeedOne}{61.1}
\newcommand{\fzsLatentSeedTwo}{62.6}
\newcommand{\fzsLatentMean}{61.7}

\newcommand{\fzsLatentGapMean}{33}

\newcommand{\fzsNewMedGapSetMean}{72}

%% file: generated/label_curve.tex
\newcommand{\lcNSetsNoDemo}{7}

\newcommand{\lcFullBudget}{400}

\newcommand{\lcTableNote}{$^\dagger$~8 of the 13 configurations; $^\S$~the 7 sets without labelled demonstrations; $^\ddagger$~without SciPrompt S2ORC, where no logistic regression could be fitted with 50 labels.}

%% file: generated/h100_round2.tex
\newcommand{\hrBaseNsets}{13}

\newcommand{\hrBasePoolOptZero}{+22.2}

\newcommand{\hrBasePoolOptClef}{-2.1}
\newcommand{\hrBasePoolOptClefLo}{-3.0}
\newcommand{\hrBasePoolOptClefHi}{-1.2}
\newcommand{\hrBaseMedOptClef}{-1.7}

\newcommand{\hrBaseNnegOptClef}{1}
\newcommand{\hrBaseGpuHours}{16.8}

\newcommand{\hrBaseGapCal}{25}
\newcommand{\hrBaseGapOpt}{65}

\newcommand{\hrNativeGpuHours}{2.3}

%% file: generated/lora.tex
\newcommand{\lrNSets}{13}

\newcommand{\lrPoolJev}{+3.7}
\newcommand{\lrPoolJevLo}{+2.6}
\newcommand{\lrPoolJevHi}{+5.1}
\newcommand{\lrAheadJev}{6}

\newcommand{\lrAheadEnc}{9}
\newcommand{\lrBehindEnc}{0}

\newcommand{\lrNReach}{6}
\newcommand{\lrNReachWord}{six}

\newcommand{\lrJevNReachWord}{three}

\newcommand{\lrGpuHours}{14.3}

\newcommand{\lrPeakGiB}{56.4}

\newcommand{\efSearchMinutesMed}{11}

%% file: generated/msd1.tex
\newcommand{\msBetter}{12}
\newcommand{\msWithin}{13}
\newcommand{\msWorse}{20}
\newcommand{\msMedian}{$-$0.4}

\newcommand{\msCiccClinc}{84.9}

\newcommand{\msCiccAcid}{20.5}

\newcommand{\msCiccAtis}{0.8}

\newcommand{\msCiccHwu}{72.5}

\newcommand{\msSpend}{62.55}

\newcommand{\msSameGroup}{37}

\newcommand{\msRho}{0.87}

\newcommand{\msPairNHigher}{7}
\newcommand{\msPairNLower}{9}

%% file: generated/msd1_rq3.tex
\newcommand{\msRqPoolOpt}{+16.3}

\newcommand{\msRqNholmOptJev}{0}
\newcommand{\msRqNnegOptJev}{1}
\newcommand{\msRqMedOptJev}{-1.0}
\newcommand{\msRqMedGapCal}{35}
\newcommand{\msRqMedGapOpt}{74}

\newcommand{\msRqEncAhead}{2}
\newcommand{\msRqEncBehind}{1}

\newcommand{\msRqEncMedDiff}{-0.9}

%% file: generated/answer.tex
\newcommand{\atN}{13}

\newcommand{\atNCfg}{8}

\newcommand{\atCalGapFrozen}{23}

\newcommand{\atCalGapNoDress}{22}

\newcommand{\atOptGap}{72}
\newcommand{\atOptGapFrozen}{72}
\newcommand{\atOptGapCfg}{72}
\newcommand{\atOptGapNoDress}{69}

\newcommand{\atOptGapFrozenMin}{33}
\newcommand{\atOptGapFrozenMax}{106}
\newcommand{\atEncGap}{59}

\newcommand{\atEncGapNoDress}{53}

\newcommand{\atLoraGap}{97}

\newcommand{\atLoraGapNoDress}{96}

\newcommand{\atOptGapAllLo}{41}
\newcommand{\atOptGapAllHi}{97}

\newcommand{\atXferClefGap}{64}
\newcommand{\atXferFlashGap}{54}
\newcommand{\atXferLunaGap}{28}
\newcommand{\atXferMsdGap}{78}
\newcommand{\atXferQwenGap}{60}
\newcommand{\atGepaTextGap}{35}
\newcommand{\atGepaProgGap}{40}
\newcommand{\atGepaDemoGap}{66}

\newcommand{\atGpuEnc}{0.37}
\newcommand{\atGpuLora}{1.03}

%% file: preprint.tex
\author{Marek \v{S}uppa \\
  Comenius University in Bratislava \\
  Cisco Systems
  \And
  Jaroslav Kop\v{c}an \\
  Kempelen Institute of \\
  Intelligent Technologies}
\hypersetup{pdfauthor={Marek \v{S}uppa, Jaroslav Kop\v{c}an}}

\renewcommand{\Uponacceptance}{We}
\renewcommand{\uponacceptance}{}
\renewcommand{\releaseat}{ at \url{https://github.com/NaiveNeuron/TDM-Gauntlet}}

%% file: generated/body_numbers.tex
\newcommand{\bodyNWithLLM}{30}
\newcommand{\bodyRhoLLM}{0.37}
\newcommand{\bodyNJevGeLLM}{22}
\newcommand{\bodyRhoK}{-0.48}
\newcommand{\bodyNSmallK}{29}
\newcommand{\bodyNLargeK}{8}
\newcommand{\bodyMedRelSmallK}{0.87}
\newcommand{\bodyMedRelLargeK}{0.61}
\newcommand{\bodyNLargeKBelow}{7}
\newcommand{\bodyMedRelAll}{0.86}
\newcommand{\bodyMedRelAllPct}{86}
\newcommand{\bodyCostPerKMedian}{\$0.05}
\newcommand{\bodyCostPerKMin}{\$0.015}
\newcommand{\bodyCostPerKMax}{\$101}
\newcommand{\bodyRhoCost}{-0.19}
\newcommand{\bodyCostPerKMedReplaces}{\$0.06}
\newcommand{\bodyLargeLabelCostShare}{85}
\newcommand{\bodyWikiCost}{\$105}
\newcommand{\bodyWikiTokensB}{2.50}
\newcommand{\bodyCiccMtodZero}{0.3}
\newcommand{\bodyCiccMtodAlpha}{1}
\newcommand{\bodyCiccMtodAcc}{98.2}
\newcommand{\bodyCiccHwuZero}{4.5}
\newcommand{\bodyCiccHwuAlpha}{5}
\newcommand{\bodyCiccHwuAcc}{81.5}
\newcommand{\bodyCiccClincZero}{1.8}
\newcommand{\bodyCiccClincAlpha}{1}
\newcommand{\bodyCiccClincAcc}{92.6}
\newcommand{\bodyCiccAcidZero}{2.4}
\newcommand{\bodyCiccAcidAlpha}{2}
\newcommand{\bodyCiccAcidAcc}{85.5}
\newcommand{\bodyCiccAtisZero}{3.5}
\newcommand{\bodyCiccAtisAlpha}{1}
\newcommand{\bodyCiccAtisAcc}{66.4}
\newcommand{\bodyCiccBankZero}{7.1}
\newcommand{\bodyCiccBankAlpha}{2}
\newcommand{\bodyCiccBankAcc}{78.9}
\newcommand{\bodyCiccNFull}{3}
\newcommand{\bodyCiccNSets}{5}
\newcommand{\bodyCalN}{11}
\newcommand{\bodyCalNOver}{11}
\newcommand{\bodyCalEceMin}{0.072}
\newcommand{\bodyCalEceMax}{0.287}
\newcommand{\bodyCalZeroMed}{35}
\newcommand{\bodyGapMin}{15}
\newcommand{\bodyGapMax}{107}
\newcommand{\bodyGapMedian}{31}
\newcommand{\bodyNTopoChanged}{3}
\newcommand{\bodyNProgramSig}{5}
\newcommand{\bodyNTextSig}{8}
\newcommand{\bodyGrowthText}{4.6}
\newcommand{\bodyGrowthProgram}{2.1}
\newcommand{\bodyReflCallsText}{1,499}
\newcommand{\bodyReflCallsProgram}{246}
\newcommand{\bodyReflCallsAll}{1,745}
\newcommand{\bodyReflCostAll}{\$105}
\newcommand{\bodyJevCostAll}{\$16.00}
\newcommand{\bodyTbdGapText}{30}
\newcommand{\bodyTbdValText}{67.1}
\newcommand{\bodyTbdTestText}{58.7}
\newcommand{\bodyTbdCharsText}{13,507}
\newcommand{\bodyTbdGapProgram}{46}
\newcommand{\bodyTbdValProgram}{66.4}
\newcommand{\bodyTbdTestProgram}{60.9}
\newcommand{\bodyTbdCharsProgram}{4,477}
\newcommand{\bodyTbdValZero}{61.6}
\newcommand{\bodyTbdTestZero}{54.4}
\newcommand{\bodyTbdCharsSeed}{3,221}
\newcommand{\bodyEzGapText}{43}
\newcommand{\bodyEzValText}{71.2}
\newcommand{\bodyEzTestText}{69.5}
\newcommand{\bodyEzCharsText}{15,398}
\newcommand{\bodyEzGapProgram}{74}
\newcommand{\bodyEzValProgram}{79.1}
\newcommand{\bodyEzTestProgram}{76.0}
\newcommand{\bodyEzCharsProgram}{3,934}
\newcommand{\bodyEzValZero}{63.1}
\newcommand{\bodyEzTestZero}{60.7}
\newcommand{\bodyEzCharsSeed}{2,021}
\newcommand{\bodySmGapText}{91}
\newcommand{\bodySmValText}{61.3}
\newcommand{\bodySmTestText}{58.4}
\newcommand{\bodySmCharsText}{17,116}
\newcommand{\bodySmGapProgram}{107}
\newcommand{\bodySmValProgram}{61.2}
\newcommand{\bodySmTestProgram}{61.3}
\newcommand{\bodySmCharsProgram}{5,665}
\newcommand{\bodySmValZero}{45.0}
\newcommand{\bodySmTestZero}{41.9}
\newcommand{\bodySmCharsSeed}{2,562}
\newcommand{\bodyEdGapText}{23}
\newcommand{\bodyEdValText}{70.4}
\newcommand{\bodyEdTestText}{70.5}
\newcommand{\bodyEdCharsText}{14,509}
\newcommand{\bodyEdGapProgram}{8}
\newcommand{\bodyEdValProgram}{70.9}
\newcommand{\bodyEdTestProgram}{67.7}
\newcommand{\bodyEdCharsProgram}{6,092}
\newcommand{\bodyEdValZero}{65.3}
\newcommand{\bodyEdTestZero}{66.3}
\newcommand{\bodyEdCharsSeed}{4,177}
\newcommand{\bodyInliGapText}{15}
\newcommand{\bodyInliValText}{79.8}
\newcommand{\bodyInliTestText}{79.2}
\newcommand{\bodyInliCharsText}{8,625}
\newcommand{\bodyInliGapProgram}{15}
\newcommand{\bodyInliValProgram}{81.2}
\newcommand{\bodyInliTestProgram}{79.3}
\newcommand{\bodyInliCharsProgram}{6,129}
\newcommand{\bodyInliValZero}{78.6}
\newcommand{\bodyInliTestZero}{76.9}
\newcommand{\bodyInliCharsSeed}{1,709}
\newcommand{\bodyDrCGapText}{23}
\newcommand{\bodyDrCValText}{38.9}
\newcommand{\bodyDrCTestText}{37.5}
\newcommand{\bodyDrCCharsText}{5,408}
\newcommand{\bodyDrCGapProgram}{-2}
\newcommand{\bodyDrCValProgram}{31.9}
\newcommand{\bodyDrCTestProgram}{29.1}
\newcommand{\bodyDrCCharsProgram}{5,589}
\newcommand{\bodyDrCValZero}{29.7}
\newcommand{\bodyDrCTestZero}{29.8}
\newcommand{\bodyDrCCharsSeed}{1,810}
\newcommand{\bodyDrOGapText}{20}
\newcommand{\bodyDrOValText}{47.3}
\newcommand{\bodyDrOTestText}{49.1}
\newcommand{\bodyDrOCharsText}{12,511}
\newcommand{\bodyDrOGapProgram}{25}
\newcommand{\bodyDrOValProgram}{49.4}
\newcommand{\bodyDrOTestProgram}{51.0}
\newcommand{\bodyDrOCharsProgram}{4,917}
\newcommand{\bodyDrOValZero}{34.5}
\newcommand{\bodyDrOTestZero}{42.1}
\newcommand{\bodyDrOCharsSeed}{2,150}
\newcommand{\bodyDrLGapText}{36}
\newcommand{\bodyDrLValText}{48.4}
\newcommand{\bodyDrLTestText}{44.1}
\newcommand{\bodyDrLCharsText}{6,430}
\newcommand{\bodyDrLGapProgram}{-34}
\newcommand{\bodyDrLValProgram}{43.6}
\newcommand{\bodyDrLTestProgram}{27.7}
\newcommand{\bodyDrLCharsProgram}{3,634}
\newcommand{\bodyDrLValZero}{40.3}
\newcommand{\bodyDrLTestZero}{35.6}
\newcommand{\bodyDrLCharsSeed}{2,052}
\newcommand{\bodyTbdVagueGoldPct}{45}
\newcommand{\bodyTbdVaguePredPct}{22}
\newcommand{\bodyTbdVagueRecall}{30}
\newcommand{\bodyEzNoneF}{46.0}
\newcommand{\bodyEzAgainstF}{71.7}
\newcommand{\bodySmIncF}{20.4}
\newcommand{\bodySmIncPred}{85}
\newcommand{\bodySmStereoPred}{370}
\newcommand{\bodySmPerClass}{150}
\newcommand{\bodyInliImpliedAcc}{61.9}
\newcommand{\bodyInliNeutralAcc}{87.4}
\newcommand{\audPinoseTrueFalseJev}{96.9}
\newcommand{\audPinoseTrueFalsePaper}{86.6}
\newcommand{\audPinoseTrueFalseK}{2}
\newcommand{\audPinoseNqJev}{88.0}
\newcommand{\audPinoseNqPaper}{80.4}
\newcommand{\audPinoseNqK}{2}
\newcommand{\audPinoseTriviaqaJev}{96.4}
\newcommand{\audPinoseTriviaqaPaper}{83.9}
\newcommand{\audPinoseTriviaqaK}{2}
\newcommand{\audPinoseWebqJev}{88.4}
\newcommand{\audPinoseWebqPaper}{83.3}
\newcommand{\audPinoseWebqK}{2}
\newcommand{\audPrpGraphTrecCovidJev}{82.1}
\newcommand{\audPrpGraphTrecCovidPaper}{78.8}
\newcommand{\audPrpGraphTrecCovidK}{100}
\newcommand{\audXstestJev}{87.0}
\newcommand{\audXstestPaper}{80.4}
\newcommand{\audXstestLLM}{80.4}
\newcommand{\audXstestK}{3}
\newcommand{\audFactpicoJev}{53.4}
\newcommand{\audFactpicoPaper}{47.4}
\newcommand{\audFactpicoLLM}{47.5}
\newcommand{\audFactpicoK}{5}
\newcommand{\audFeniceJev}{77.4}
\newcommand{\audFenicePaper}{65.7}
\newcommand{\audFeniceK}{2}
\newcommand{\audBiorcEightZeroZeroJev}{90.8}
\newcommand{\audBiorcEightZeroZeroPaper}{90.8}
\newcommand{\audBiorcEightZeroZeroLLM}{47.6}
\newcommand{\audBiorcEightZeroZeroK}{6}
\newcommand{\audScipromptArxivJev}{79.0}
\newcommand{\audScipromptArxivPaper}{70.2}
\newcommand{\audScipromptArxivLLM}{54.9}
\newcommand{\audScipromptArxivK}{53}
\newcommand{\audScipromptSTwoorcJev}{52.9}
\newcommand{\audScipromptSTwoorcPaper}{56.7}
\newcommand{\audScipromptSTwoorcLLM}{49.6}
\newcommand{\audScipromptSTwoorcK}{19}
\newcommand{\audScipromptSdpraJev}{91.3}
\newcommand{\audScipromptSdpraPaper}{90.7}
\newcommand{\audScipromptSdpraLLM}{81.2}
\newcommand{\audScipromptSdpraK}{7}
\newcommand{\audScipromptEmergingNlpJev}{95.0}
\newcommand{\audScipromptEmergingNlpPaper}{89.3}
\newcommand{\audScipromptEmergingNlpLLM}{8.6}
\newcommand{\audScipromptEmergingNlpK}{21}
\newcommand{\audScalarConstructsImmigrationFearPairwiseJev}{78.7}
\newcommand{\audScalarConstructsImmigrationFearPairwisePaper}{91.0}
\newcommand{\audScalarConstructsImmigrationFearPairwiseLLM}{74.0}
\newcommand{\audScalarConstructsImmigrationFearPairwiseK}{2}
\newcommand{\audScalarConstructsAdNegativityPairwiseJev}{87.1}
\newcommand{\audScalarConstructsAdNegativityPairwisePaper}{91.0}
\newcommand{\audScalarConstructsAdNegativityPairwiseLLM}{86.0}
\newcommand{\audScalarConstructsAdNegativityPairwiseK}{2}
\newcommand{\audScalarConstructsGrandstandingPairwiseJev}{58.6}
\newcommand{\audScalarConstructsGrandstandingPairwisePaper}{78.0}
\newcommand{\audScalarConstructsGrandstandingPairwiseLLM}{54.0}
\newcommand{\audScalarConstructsGrandstandingPairwiseK}{2}
\newcommand{\audScalarConstructsImmigrationFearPointwiseJev}{77.4}
\newcommand{\audScalarConstructsImmigrationFearPointwisePaper}{91.0}
\newcommand{\audScalarConstructsImmigrationFearPointwiseLLM}{81.0}
\newcommand{\audScalarConstructsImmigrationFearPointwiseK}{9}
\newcommand{\audScalarConstructsAdNegativityPointwiseJev}{89.4}
\newcommand{\audScalarConstructsAdNegativityPointwisePaper}{91.0}
\newcommand{\audScalarConstructsAdNegativityPointwiseLLM}{91.0}
\newcommand{\audScalarConstructsAdNegativityPointwiseK}{9}
\newcommand{\audScalarConstructsGrandstandingPointwiseJev}{70.0}
\newcommand{\audScalarConstructsGrandstandingPointwisePaper}{78.0}
\newcommand{\audScalarConstructsGrandstandingPointwiseLLM}{60.0}
\newcommand{\audScalarConstructsGrandstandingPointwiseK}{9}
\newcommand{\audLifetoxJev}{78.4}
\newcommand{\audLifetoxPaper}{96.5}
\newcommand{\audLifetoxLLM}{74.4}
\newcommand{\audLifetoxK}{2}
\newcommand{\audEditIntentJev}{66.7}
\newcommand{\audEditIntentPaper}{84.3}
\newcommand{\audEditIntentLLM}{37.0}
\newcommand{\audEditIntentK}{5}
\newcommand{\audEzstanceSubtaskAJev}{60.7}
\newcommand{\audEzstanceSubtaskAPaper}{81.2}
\newcommand{\audEzstanceSubtaskALLM}{49.9}
\newcommand{\audEzstanceSubtaskAK}{3}
\newcommand{\audEzstanceSubtaskBJev}{59.5}
\newcommand{\audEzstanceSubtaskBPaper}{78.3}
\newcommand{\audEzstanceSubtaskBLLM}{49.5}
\newcommand{\audEzstanceSubtaskBK}{3}
\newcommand{\audContempoMatresJev}{72.3}
\newcommand{\audContempoMatresPaper}{83.2}
\newcommand{\audContempoMatresLLM}{56.7}
\newcommand{\audContempoMatresK}{4}
\newcommand{\audContempoTbdenseJev}{54.3}
\newcommand{\audContempoTbdensePaper}{68.6}
\newcommand{\audContempoTbdenseLLM}{67.6}
\newcommand{\audContempoTbdenseK}{6}
\newcommand{\audMismatchedJev}{71.2}
\newcommand{\audMismatchedPaper}{78.2}
\newcommand{\audMismatchedLLM}{55.6}
\newcommand{\audMismatchedK}{4}
\newcommand{\audInliJev}{76.8}
\newcommand{\audInliPaper}{92.4}
\newcommand{\audInliLLM}{77.0}
\newcommand{\audInliK}{4}
\newcommand{\audMedreadmeJev}{71.4}
\newcommand{\audMedreadmePaper}{83.3}
\newcommand{\audMedreadmeLLM}{72.1}
\newcommand{\audMedreadmeK}{6}
\newcommand{\audNavUnknownClincOneFiveZeroJev}{86.0}
\newcommand{\audNavUnknownClincOneFiveZeroPaper}{95.7}
\newcommand{\audNavUnknownClincOneFiveZeroLLM}{83.7}
\newcommand{\audNavUnknownClincOneFiveZeroK}{76}
\newcommand{\audNavUnknownBankingSevenSevenJev}{71.2}
\newcommand{\audNavUnknownBankingSevenSevenPaper}{83.2}
\newcommand{\audNavUnknownBankingSevenSevenLLM}{69.4}
\newcommand{\audNavUnknownBankingSevenSevenK}{39}
\newcommand{\audMadeJev}{32.9}
\newcommand{\audMadePaper}{54.0}
\newcommand{\audMadeLLM}{43.0}
\newcommand{\audMadeK}{1,154}
\newcommand{\audSmarterHatexplainJev}{55.6}
\newcommand{\audSmarterHatexplainPaper}{64.0}
\newcommand{\audSmarterHatexplainLLM}{56.0}
\newcommand{\audSmarterHatexplainK}{3}
\newcommand{\audSmarterLatentHateJev}{58.0}
\newcommand{\audSmarterLatentHatePaper}{69.0}
\newcommand{\audSmarterLatentHateLLM}{60.0}
\newcommand{\audSmarterLatentHateK}{3}
\newcommand{\audSmarterImplicitHateJev}{41.9}
\newcommand{\audSmarterImplicitHatePaper}{60.0}
\newcommand{\audSmarterImplicitHateLLM}{58.0}
\newcommand{\audSmarterImplicitHateK}{6}
\newcommand{\audDressNewJev}{20.6}
\newcommand{\audDressNewPaper}{69.1}
\newcommand{\audDressNewLLM}{30.4}
\newcommand{\audDressNewK}{9}
\newcommand{\audTtpTramJev}{44.0}
\newcommand{\audTtpTramPaper}{77.0}
\newcommand{\audTtpTramK}{737}
\newcommand{\audTtpProceduresPlusJev}{45.9}
\newcommand{\audTtpProceduresPlusPaper}{75.7}
\newcommand{\audTtpProceduresPlusK}{737}
\newcommand{\audTtpExpertJev}{54.8}
\newcommand{\audTtpExpertPaper}{73.1}
\newcommand{\audTtpExpertK}{737}
\newcommand{\audLegalsegJev}{52.0}
\newcommand{\audLegalsegPaper}{77.0}
\newcommand{\audLegalsegLLM}{8.0}
\newcommand{\audLegalsegK}{7}
\newcommand{\audRelembWikiwdJev}{54.1}
\newcommand{\audRelembWikiwdPaper}{89.1}
\newcommand{\audRelembWikiwdK}{352}
\newcommand{\audRelembNytfbJev}{19.6}
\newcommand{\audRelembNytfbPaper}{94.9}
\newcommand{\audRelembNytfbK}{56}
\newcommand{\audCiccMtodJev}{97.3}
\newcommand{\audCiccMtodPaper}{100.0}
\newcommand{\audCiccMtodK}{12}
\newcommand{\audCiccHwuSixFourJev}{47.9}
\newcommand{\audCiccHwuSixFourPaper}{82.0}
\newcommand{\audCiccHwuSixFourK}{68}
\newcommand{\audCiccCOneFiveZeroJev}{0.0}
\newcommand{\audCiccCOneFiveZeroPaper}{97.0}
\newcommand{\audCiccCOneFiveZeroK}{150}
\newcommand{\audCiccAcidJev}{0.0}
\newcommand{\audCiccAcidPaper}{87.0}
\newcommand{\audCiccAcidK}{175}
\newcommand{\audCiccAtisJev}{0.0}
\newcommand{\audCiccAtisPaper}{98.0}
\newcommand{\audCiccAtisK}{26}
\newcommand{\optNTasks}{8}
\newcommand{\optNSeeds}{3}
\newcommand{\optDeltaMin}{+4.8}
\newcommand{\optDeltaMax}{+34.0}
\newcommand{\optNSig}{8}
\newcommand{\optSeedDeltaMin}{+4.1}
\newcommand{\optNBeatLLM}{7}
\newcommand{\optNBeatPaper}{2}
\newcommand{\optSeedSdMin}{0.7}
\newcommand{\optSeedSdMax}{3.2}
\newcommand{\optSeedRangeMax}{6.2}
\newcommand{\optNVtwoInRange}{5}
\newcommand{\optNVtwoBelowRange}{3}
\newcommand{\optGapMin}{31}
\newcommand{\optGapMax}{106}
\newcommand{\optGapMedian}{68}
\newcommand{\optNWithinPaper}{2}
\newcommand{\optNValAbove}{6}
\newcommand{\optValGapMin}{1.4}
\newcommand{\optValGapMax}{4.6}
\newcommand{\optValBelowMax}{6.1}
\newcommand{\optNNativeBetter}{3}
\newcommand{\optNNativeWorse}{1}
\newcommand{\optNTailoredBest}{2}
\newcommand{\optTestCostMin}{\$0.02}
\newcommand{\optTestCostMax}{\$0.24}
\newcommand{\optTestCostPerKMedian}{\$0.05}
\newcommand{\optTestCostPerKMin}{\$0.02}
\newcommand{\optTestCostPerKMax}{\$0.38}
\newcommand{\optTbdMean}{61.1}
\newcommand{\optTbdSd}{1.4}
\newcommand{\optTbdMeanDelta}{+6.7}
\newcommand{\optTbdSeedMin}{59.7}
\newcommand{\optTbdSeedMax}{62.4}
\newcommand{\optTbdSeedTokensMin}{615}
\newcommand{\optTbdSeedTokensMax}{1,125}
\newcommand{\optEzMean}{77.0}
\newcommand{\optEzSd}{1.1}
\newcommand{\optEzMeanDelta}{+16.4}
\newcommand{\optEzSeedMin}{75.8}
\newcommand{\optEzSeedMax}{77.7}
\newcommand{\optEzSeedTokensMin}{481}
\newcommand{\optEzSeedTokensMax}{736}
\newcommand{\optSmMean}{61.2}
\newcommand{\optSmSd}{3.2}
\newcommand{\optSmMeanDelta}{+19.3}
\newcommand{\optSmSeedMin}{58.5}
\newcommand{\optSmSeedMax}{64.7}
\newcommand{\optSmSeedTokensMin}{1,482}
\newcommand{\optSmSeedTokensMax}{2,582}
\newcommand{\optEdMean}{73.9}
\newcommand{\optEdSd}{1.0}
\newcommand{\optEdMeanDelta}{+7.6}
\newcommand{\optEdSeedMin}{72.8}
\newcommand{\optEdSeedMax}{74.8}
\newcommand{\optEdSeedTokensMin}{615}
\newcommand{\optEdSeedTokensMax}{1,050}
\newcommand{\optInliMean}{81.8}
\newcommand{\optInliSd}{0.7}
\newcommand{\optInliMeanDelta}{+4.8}
\newcommand{\optInliSeedMin}{81.1}
\newcommand{\optInliSeedMax}{82.4}
\newcommand{\optInliSeedTokensMin}{518}
\newcommand{\optInliSeedTokensMax}{582}
\newcommand{\optDrCMean}{63.8}
\newcommand{\optDrCSd}{2.1}
\newcommand{\optDrCMeanDelta}{+34.0}
\newcommand{\optDrCSeedMin}{61.9}
\newcommand{\optDrCSeedMax}{66.0}
\newcommand{\optDrCSeedTokensMin}{4,476}
\newcommand{\optDrCSeedTokensMax}{9,164}
\newcommand{\optDrOMean}{62.0}
\newcommand{\optDrOSd}{2.2}
\newcommand{\optDrOMeanDelta}{+19.9}
\newcommand{\optDrOSeedMin}{59.7}
\newcommand{\optDrOSeedMax}{64.1}
\newcommand{\optDrOSeedTokensMin}{1,206}
\newcommand{\optDrOSeedTokensMax}{8,597}
\newcommand{\optDrLMean}{57.0}
\newcommand{\optDrLSd}{2.9}
\newcommand{\optDrLMeanDelta}{+21.4}
\newcommand{\optDrLSeedMin}{53.8}
\newcommand{\optDrLSeedMax}{59.1}
\newcommand{\optDrLSeedTokensMin}{4,245}
\newcommand{\optDrLSeedTokensMax}{4,808}
\newcommand{\optTbdZero}{54.4}
\newcommand{\optTbdZeroVal}{61.6}
\newcommand{\optTbdZeroTokens}{1,030}
\newcommand{\optTbdNative}{57.4}
\newcommand{\optTbdNativeVal}{64.4}
\newcommand{\optTbdNativeDelta}{+3.1}
\newcommand{\optTbdNativeLo}{-0.5}
\newcommand{\optTbdNativeHi}{+6.4}
\newcommand{\optTbdNativeTokens}{721}
\newcommand{\optTbdVoneText}{58.7}
\newcommand{\optTbdVoneTextVal}{67.1}
\newcommand{\optTbdVoneTextDelta}{+4.3}
\newcommand{\optTbdVoneTextLo}{-0.3}
\newcommand{\optTbdVoneTextHi}{+8.5}
\newcommand{\optTbdVoneTextTokens}{3,526}
\newcommand{\optTbdVoneProg}{60.9}
\newcommand{\optTbdVoneProgVal}{66.4}
\newcommand{\optTbdVoneProgDelta}{+6.5}
\newcommand{\optTbdVoneProgLo}{+2.6}
\newcommand{\optTbdVoneProgHi}{+10.1}
\newcommand{\optTbdVoneProgTokens}{3,529}
\newcommand{\optTbdVtwo}{60.2}
\newcommand{\optTbdVtwoVal}{67.2}
\newcommand{\optTbdVtwoDelta}{+5.8}
\newcommand{\optTbdVtwoLo}{+1.6}
\newcommand{\optTbdVtwoHi}{+9.1}
\newcommand{\optTbdVtwoTokens}{3,088}
\newcommand{\optTbdVthree}{62.4}
\newcommand{\optTbdVthreeVal}{68.9}
\newcommand{\optTbdVthreeDelta}{+8.0}
\newcommand{\optTbdVthreeLo}{+3.1}
\newcommand{\optTbdVthreeHi}{+12.2}
\newcommand{\optTbdVthreeTokens}{615}
\newcommand{\optTbdTailored}{63.0}
\newcommand{\optTbdTailoredVal}{67.8}
\newcommand{\optTbdTailoredDelta}{+8.6}
\newcommand{\optTbdTailoredLo}{+6.1}
\newcommand{\optTbdTailoredHi}{+11.4}
\newcommand{\optTbdTailoredTokens}{721}
\newcommand{\optTbdPaper}{68.6}
\newcommand{\optTbdLLM}{67.6}
\newcommand{\optTbdVthreeVtwoDelta}{+2.2}
\newcommand{\optEzZero}{60.7}
\newcommand{\optEzZeroVal}{63.1}
\newcommand{\optEzZeroTokens}{510}
\newcommand{\optEzNative}{68.6}
\newcommand{\optEzNativeVal}{70.9}
\newcommand{\optEzNativeDelta}{+7.9}
\newcommand{\optEzNativeLo}{+7.2}
\newcommand{\optEzNativeHi}{+8.7}
\newcommand{\optEzNativeTokens}{656}
\newcommand{\optEzVoneText}{69.5}
\newcommand{\optEzVoneTextVal}{71.2}
\newcommand{\optEzVoneTextDelta}{+8.9}
\newcommand{\optEzVoneTextLo}{+8.1}
\newcommand{\optEzVoneTextHi}{+9.7}
\newcommand{\optEzVoneTextTokens}{2,581}
\newcommand{\optEzVoneProg}{76.0}
\newcommand{\optEzVoneProgVal}{79.1}
\newcommand{\optEzVoneProgDelta}{+15.3}
\newcommand{\optEzVoneProgLo}{+14.3}
\newcommand{\optEzVoneProgHi}{+16.3}
\newcommand{\optEzVoneProgTokens}{717}
\newcommand{\optEzVtwo}{74.9}
\newcommand{\optEzVtwoVal}{78.1}
\newcommand{\optEzVtwoDelta}{+14.2}
\newcommand{\optEzVtwoLo}{+13.3}
\newcommand{\optEzVtwoHi}{+15.2}
\newcommand{\optEzVtwoTokens}{1,511}
\newcommand{\optEzVthree}{77.7}
\newcommand{\optEzVthreeVal}{79.8}
\newcommand{\optEzVthreeDelta}{+17.0}
\newcommand{\optEzVthreeLo}{+16.0}
\newcommand{\optEzVthreeHi}{+18.1}
\newcommand{\optEzVthreeTokens}{736}
\newcommand{\optEzTailored}{76.3}
\newcommand{\optEzTailoredVal}{77.5}
\newcommand{\optEzTailoredDelta}{+15.6}
\newcommand{\optEzTailoredLo}{+14.7}
\newcommand{\optEzTailoredHi}{+16.7}
\newcommand{\optEzTailoredTokens}{931}
\newcommand{\optEzPaper}{81.2}
\newcommand{\optEzLLM}{49.9}
\newcommand{\optEzVthreeVtwoDelta}{+2.8}
\newcommand{\optSmZero}{41.9}
\newcommand{\optSmZeroVal}{45.0}
\newcommand{\optSmZeroTokens}{765}
\newcommand{\optSmNative}{53.2}
\newcommand{\optSmNativeVal}{56.2}
\newcommand{\optSmNativeDelta}{+11.3}
\newcommand{\optSmNativeLo}{+8.8}
\newcommand{\optSmNativeHi}{+14.1}
\newcommand{\optSmNativeTokens}{610}
\newcommand{\optSmVoneText}{58.4}
\newcommand{\optSmVoneTextVal}{61.3}
\newcommand{\optSmVoneTextDelta}{+16.5}
\newcommand{\optSmVoneTextLo}{+13.3}
\newcommand{\optSmVoneTextHi}{+20.0}
\newcommand{\optSmVoneTextTokens}{4,068}
\newcommand{\optSmVoneProg}{61.3}
\newcommand{\optSmVoneProgVal}{61.2}
\newcommand{\optSmVoneProgDelta}{+19.4}
\newcommand{\optSmVoneProgLo}{+15.9}
\newcommand{\optSmVoneProgHi}{+23.1}
\newcommand{\optSmVoneProgTokens}{1,295}
\newcommand{\optSmVtwo}{61.2}
\newcommand{\optSmVtwoVal}{63.3}
\newcommand{\optSmVtwoDelta}{+19.3}
\newcommand{\optSmVtwoLo}{+15.9}
\newcommand{\optSmVtwoHi}{+22.9}
\newcommand{\optSmVtwoTokens}{3,342}
\newcommand{\optSmVthree}{60.4}
\newcommand{\optSmVthreeVal}{65.7}
\newcommand{\optSmVthreeDelta}{+18.5}
\newcommand{\optSmVthreeLo}{+14.9}
\newcommand{\optSmVthreeHi}{+22.1}
\newcommand{\optSmVthreeTokens}{1,482}
\newcommand{\optSmTailored}{57.1}
\newcommand{\optSmTailoredVal}{59.1}
\newcommand{\optSmTailoredDelta}{+15.3}
\newcommand{\optSmTailoredLo}{+12.1}
\newcommand{\optSmTailoredHi}{+18.5}
\newcommand{\optSmTailoredTokens}{1,021}
\newcommand{\optSmPaper}{60.0}
\newcommand{\optSmLLM}{58.0}
\newcommand{\optSmVthreeVtwoDelta}{-0.8}
\newcommand{\optEdZero}{66.3}
\newcommand{\optEdZeroVal}{65.3}
\newcommand{\optEdZeroTokens}{737}
\newcommand{\optEdNative}{68.8}
\newcommand{\optEdNativeVal}{71.3}
\newcommand{\optEdNativeDelta}{+2.5}
\newcommand{\optEdNativeLo}{+0.7}
\newcommand{\optEdNativeHi}{+4.9}
\newcommand{\optEdNativeTokens}{1,001}
\newcommand{\optEdVoneText}{70.5}
\newcommand{\optEdVoneTextVal}{70.4}
\newcommand{\optEdVoneTextDelta}{+4.2}
\newcommand{\optEdVoneTextLo}{+1.7}
\newcommand{\optEdVoneTextHi}{+6.7}
\newcommand{\optEdVoneTextTokens}{2,385}
\newcommand{\optEdVoneProg}{67.7}
\newcommand{\optEdVoneProgVal}{70.9}
\newcommand{\optEdVoneProgDelta}{+1.4}
\newcommand{\optEdVoneProgLo}{-1.0}
\newcommand{\optEdVoneProgHi}{+3.9}
\newcommand{\optEdVoneProgTokens}{969}
\newcommand{\optEdVtwo}{73.0}
\newcommand{\optEdVtwoVal}{74.5}
\newcommand{\optEdVtwoDelta}{+6.7}
\newcommand{\optEdVtwoLo}{+3.7}
\newcommand{\optEdVtwoHi}{+9.2}
\newcommand{\optEdVtwoTokens}{2,500}
\newcommand{\optEdVthree}{74.0}
\newcommand{\optEdVthreeVal}{74.2}
\newcommand{\optEdVthreeDelta}{+7.7}
\newcommand{\optEdVthreeLo}{+4.5}
\newcommand{\optEdVthreeHi}{+10.7}
\newcommand{\optEdVthreeTokens}{686}
\newcommand{\optEdTailored}{76.2}
\newcommand{\optEdTailoredVal}{78.7}
\newcommand{\optEdTailoredDelta}{+9.9}
\newcommand{\optEdTailoredLo}{+7.1}
\newcommand{\optEdTailoredHi}{+12.7}
\newcommand{\optEdTailoredTokens}{1,758}
\newcommand{\optEdPaper}{84.3}
\newcommand{\optEdLLM}{37.0}
\newcommand{\optEdVthreeVtwoDelta}{+0.9}
\newcommand{\optInliZero}{77.0}
\newcommand{\optInliZeroVal}{78.6}
\newcommand{\optInliZeroTokens}{660}
\newcommand{\optInliNative}{77.3}
\newcommand{\optInliNativeVal}{79.2}
\newcommand{\optInliNativeDelta}{+0.4}
\newcommand{\optInliNativeLo}{-0.2}
\newcommand{\optInliNativeHi}{+1.0}
\newcommand{\optInliNativeTokens}{593}
\newcommand{\optInliVoneText}{79.2}
\newcommand{\optInliVoneTextVal}{79.8}
\newcommand{\optInliVoneTextDelta}{+2.3}
\newcommand{\optInliVoneTextLo}{+1.3}
\newcommand{\optInliVoneTextHi}{+3.3}
\newcommand{\optInliVoneTextTokens}{2,104}
\newcommand{\optInliVoneProg}{79.3}
\newcommand{\optInliVoneProgVal}{81.2}
\newcommand{\optInliVoneProgDelta}{+2.4}
\newcommand{\optInliVoneProgLo}{+1.5}
\newcommand{\optInliVoneProgHi}{+3.3}
\newcommand{\optInliVoneProgTokens}{1,420}
\newcommand{\optInliVtwo}{81.8}
\newcommand{\optInliVtwoVal}{84.1}
\newcommand{\optInliVtwoDelta}{+4.8}
\newcommand{\optInliVtwoLo}{+3.7}
\newcommand{\optInliVtwoHi}{+6.0}
\newcommand{\optInliVtwoTokens}{4,892}
\newcommand{\optInliVthree}{82.4}
\newcommand{\optInliVthreeVal}{84.4}
\newcommand{\optInliVthreeDelta}{+5.4}
\newcommand{\optInliVthreeLo}{+4.2}
\newcommand{\optInliVthreeHi}{+6.7}
\newcommand{\optInliVthreeTokens}{582}
\newcommand{\optInliTailored}{80.0}
\newcommand{\optInliTailoredVal}{81.4}
\newcommand{\optInliTailoredDelta}{+3.1}
\newcommand{\optInliTailoredLo}{+2.0}
\newcommand{\optInliTailoredHi}{+4.1}
\newcommand{\optInliTailoredTokens}{1,337}
\newcommand{\optInliPaper}{92.4}
\newcommand{\optInliLLM}{77.0}
\newcommand{\optInliVthreeVtwoDelta}{+0.6}
\newcommand{\optDrCZero}{29.8}
\newcommand{\optDrCZeroVal}{29.7}
\newcommand{\optDrCZeroTokens}{1,063}
\newcommand{\optDrCNative}{22.9}
\newcommand{\optDrCNativeVal}{20.3}
\newcommand{\optDrCNativeDelta}{-6.8}
\newcommand{\optDrCNativeLo}{-19.4}
\newcommand{\optDrCNativeHi}{+2.9}
\newcommand{\optDrCNativeTokens}{1,087}
\newcommand{\optDrCVoneText}{37.5}
\newcommand{\optDrCVoneTextVal}{38.9}
\newcommand{\optDrCVoneTextDelta}{+7.7}
\newcommand{\optDrCVoneTextLo}{+1.7}
\newcommand{\optDrCVoneTextHi}{+11.7}
\newcommand{\optDrCVoneTextTokens}{2,044}
\newcommand{\optDrCVoneProg}{29.1}
\newcommand{\optDrCVoneProgVal}{31.9}
\newcommand{\optDrCVoneProgDelta}{-0.6}
\newcommand{\optDrCVoneProgLo}{-10.6}
\newcommand{\optDrCVoneProgHi}{+6.4}
\newcommand{\optDrCVoneProgTokens}{1,874}
\newcommand{\optDrCVtwo}{59.5}
\newcommand{\optDrCVtwoVal}{52.9}
\newcommand{\optDrCVtwoDelta}{+29.7}
\newcommand{\optDrCVtwoLo}{+19.4}
\newcommand{\optDrCVtwoHi}{+38.2}
\newcommand{\optDrCVtwoTokens}{26,849}
\newcommand{\optDrCVthree}{61.9}
\newcommand{\optDrCVthreeVal}{59.7}
\newcommand{\optDrCVthreeDelta}{+32.2}
\newcommand{\optDrCVthreeLo}{+20.8}
\newcommand{\optDrCVthreeHi}{+42.5}
\newcommand{\optDrCVthreeTokens}{4,476}
\newcommand{\optDrCTailored}{37.7}
\newcommand{\optDrCTailoredVal}{38.5}
\newcommand{\optDrCTailoredDelta}{+8.0}
\newcommand{\optDrCTailoredLo}{+0.8}
\newcommand{\optDrCTailoredHi}{+12.8}
\newcommand{\optDrCTailoredTokens}{1,439}
\newcommand{\optDrCPaper}{63.1}
\newcommand{\optDrCLLM}{31.0}
\newcommand{\optDrCVthreeVtwoDelta}{+2.4}
\newcommand{\optDrOZero}{42.1}
\newcommand{\optDrOZeroVal}{34.5}
\newcommand{\optDrOZeroTokens}{1,119}
\newcommand{\optDrONative}{41.3}
\newcommand{\optDrONativeVal}{32.9}
\newcommand{\optDrONativeDelta}{-0.9}
\newcommand{\optDrONativeLo}{-7.9}
\newcommand{\optDrONativeHi}{+8.0}
\newcommand{\optDrONativeTokens}{1,179}
\newcommand{\optDrOVoneText}{49.1}
\newcommand{\optDrOVoneTextVal}{47.3}
\newcommand{\optDrOVoneTextDelta}{+7.0}
\newcommand{\optDrOVoneTextLo}{-1.7}
\newcommand{\optDrOVoneTextHi}{+16.4}
\newcommand{\optDrOVoneTextTokens}{3,892}
\newcommand{\optDrOVoneProg}{51.0}
\newcommand{\optDrOVoneProgVal}{49.4}
\newcommand{\optDrOVoneProgDelta}{+8.9}
\newcommand{\optDrOVoneProgLo}{+1.0}
\newcommand{\optDrOVoneProgHi}{+18.5}
\newcommand{\optDrOVoneProgTokens}{1,600}
\newcommand{\optDrOVtwo}{61.8}
\newcommand{\optDrOVtwoVal}{59.9}
\newcommand{\optDrOVtwoDelta}{+19.6}
\newcommand{\optDrOVtwoLo}{+11.9}
\newcommand{\optDrOVtwoHi}{+30.3}
\newcommand{\optDrOVtwoTokens}{17,882}
\newcommand{\optDrOVthree}{62.2}
\newcommand{\optDrOVthreeVal}{55.0}
\newcommand{\optDrOVthreeDelta}{+20.1}
\newcommand{\optDrOVthreeLo}{+11.0}
\newcommand{\optDrOVthreeHi}{+32.4}
\newcommand{\optDrOVthreeTokens}{1,206}
\newcommand{\optDrOTailored}{50.3}
\newcommand{\optDrOTailoredVal}{53.5}
\newcommand{\optDrOTailoredDelta}{+8.2}
\newcommand{\optDrOTailoredLo}{+0.3}
\newcommand{\optDrOTailoredHi}{+18.7}
\newcommand{\optDrOTailoredTokens}{1,187}
\newcommand{\optDrOPaper}{77.1}
\newcommand{\optDrOLLM}{32.2}
\newcommand{\optDrOVthreeVtwoDelta}{+0.5}
\newcommand{\optDrLZero}{35.6}
\newcommand{\optDrLZeroVal}{40.3}
\newcommand{\optDrLZeroTokens}{1,101}
\newcommand{\optDrLNative}{7.6}
\newcommand{\optDrLNativeVal}{11.1}
\newcommand{\optDrLNativeDelta}{-28.0}
\newcommand{\optDrLNativeLo}{-34.8}
\newcommand{\optDrLNativeHi}{-18.5}
\newcommand{\optDrLNativeTokens}{998}
\newcommand{\optDrLVoneText}{44.1}
\newcommand{\optDrLVoneTextVal}{48.4}
\newcommand{\optDrLVoneTextDelta}{+8.5}
\newcommand{\optDrLVoneTextLo}{-0.2}
\newcommand{\optDrLVoneTextHi}{+14.3}
\newcommand{\optDrLVoneTextTokens}{2,200}
\newcommand{\optDrLVoneProg}{27.7}
\newcommand{\optDrLVoneProgVal}{43.6}
\newcommand{\optDrLVoneProgDelta}{-7.9}
\newcommand{\optDrLVoneProgLo}{-14.7}
\newcommand{\optDrLVoneProgHi}{-2.7}
\newcommand{\optDrLVoneProgTokens}{1,315}
\newcommand{\optDrLVtwo}{53.7}
\newcommand{\optDrLVtwoVal}{57.6}
\newcommand{\optDrLVtwoDelta}{+18.1}
\newcommand{\optDrLVtwoLo}{+12.3}
\newcommand{\optDrLVtwoHi}{+24.0}
\newcommand{\optDrLVtwoTokens}{16,955}
\newcommand{\optDrLVthree}{58.2}
\newcommand{\optDrLVthreeVal}{60.4}
\newcommand{\optDrLVthreeDelta}{+22.6}
\newcommand{\optDrLVthreeLo}{+14.2}
\newcommand{\optDrLVthreeHi}{+32.4}
\newcommand{\optDrLVthreeTokens}{4,245}
\newcommand{\optDrLTailored}{35.8}
\newcommand{\optDrLTailoredVal}{44.1}
\newcommand{\optDrLTailoredDelta}{+0.2}
\newcommand{\optDrLTailoredLo}{-3.6}
\newcommand{\optDrLTailoredHi}{+3.5}
\newcommand{\optDrLTailoredTokens}{1,352}
\newcommand{\optDrLPaper}{58.9}
\newcommand{\optDrLLLM}{23.1}
\newcommand{\optDrLVthreeVtwoDelta}{+4.4}
\newcommand{\optTbdValPaperAsWritten}{61.6}
\newcommand{\optTbdValNativeAsWritten}{64.4}
\newcommand{\optTbdValSearchPaper}{65.5}
\newcommand{\optTbdValSearchNative}{65.4}
\newcommand{\optTbdValHeadOnly}{65.5}
\newcommand{\optEzValPaperAsWritten}{63.1}
\newcommand{\optEzValNativeAsWritten}{70.9}
\newcommand{\optEzValSearchPaper}{63.2}
\newcommand{\optEzValSearchNative}{80.0}
\newcommand{\optEzValHeadOnly}{64.5}
\newcommand{\optSmValPaperAsWritten}{45.0}
\newcommand{\optSmValNativeAsWritten}{56.2}
\newcommand{\optSmValSearchPaper}{59.6}
\newcommand{\optSmValSearchNative}{61.5}
\newcommand{\optSmValHeadOnly}{57.0}
\newcommand{\optEdValPaperAsWritten}{65.3}
\newcommand{\optEdValNativeAsWritten}{71.3}
\newcommand{\optEdValSearchPaper}{67.2}
\newcommand{\optEdValSearchNative}{67.6}
\newcommand{\optEdValHeadOnly}{67.2}
\newcommand{\optInliValPaperAsWritten}{78.6}
\newcommand{\optInliValNativeAsWritten}{79.2}
\newcommand{\optInliValSearchPaper}{79.0}
\newcommand{\optInliValSearchNative}{75.5}
\newcommand{\optInliValHeadOnly}{79.0}
\newcommand{\optDrCValPaperAsWritten}{25.1}
\newcommand{\optDrCValNativeAsWritten}{20.3}
\newcommand{\optDrCValSearchPaper}{57.6}
\newcommand{\optDrCValSearchNative}{36.1}
\newcommand{\optDrCValHeadOnly}{33.5}
\newcommand{\optDrOValPaperAsWritten}{32.8}
\newcommand{\optDrOValNativeAsWritten}{32.9}
\newcommand{\optDrOValSearchPaper}{47.4}
\newcommand{\optDrOValSearchNative}{44.5}
\newcommand{\optDrOValHeadOnly}{31.9}
\newcommand{\optDrLValPaperAsWritten}{8.5}
\newcommand{\optDrLValNativeAsWritten}{11.1}
\newcommand{\optDrLValSearchPaper}{60.2}
\newcommand{\optDrLValSearchNative}{33.5}
\newcommand{\optDrLValHeadOnly}{45.2}
\newcommand{\optTbdVagueGold}{45}
\newcommand{\optTbdVaguePredZero}{22}
\newcommand{\optTbdVagueRecallZero}{29}
\newcommand{\optTbdBalAccZero}{52.5}
\newcommand{\optTbdMacroFZero}{40.9}
\newcommand{\optTbdVaguePredNative}{5}
\newcommand{\optTbdVagueRecallNative}{8}
\newcommand{\optTbdBalAccNative}{59.3}
\newcommand{\optTbdMacroFNative}{38.8}
\newcommand{\optTbdVaguePredVthree}{61}
\newcommand{\optTbdVagueRecallVthree}{80}
\newcommand{\optTbdBalAccVthree}{40.0}
\newcommand{\optTbdMacroFVthree}{44.1}
\newcommand{\optTbdVaguePredSeedMin}{61}
\newcommand{\optTbdVaguePredSeedMax}{66}
\newcommand{\optTbdBalAccSeedMean}{38.8}
\newcommand{\optTbdReadoutParams}{42}
\newcommand{\optFitN}{400}
\newcommand{\levTtpTram}{47.4}
\newcommand{\levTtpTramPrev}{44.0}
\newcommand{\levTtpTramPaper}{77.0}
\newcommand{\levTtpProc}{60.7}
\newcommand{\levTtpProcPrev}{45.9}
\newcommand{\levTtpProcPaper}{75.7}
\newcommand{\levTtpExpert}{58.6}
\newcommand{\levTtpExpertPrev}{54.8}
\newcommand{\levTtpExpertPaper}{73.1}
\newcommand{\levTtpExpertFull}{47.6}
\newcommand{\levTtpExpertFullPrev}{39.0}
\newcommand{\levTtpExpertFullPaper}{48.4}
\newcommand{\levWiki}{69.6}
\newcommand{\levWikiPrev}{54.1}
\newcommand{\levWikiPaper}{89.1}
\newcommand{\levMade}{39.9}
\newcommand{\levMadePrev}{32.9}
\newcommand{\levMadePaper}{54.0}
\newcommand{\levNewCost}{\$5.68}
\newcommand{\optSpendVone}{\$105}
\newcommand{\optCallsVone}{1,745}
\newcommand{\optSpendVtwo}{\$66}
\newcommand{\optCallsVtwo}{1,440}
\newcommand{\optSpendVthree}{\$17}
\newcommand{\optCallsVthree}{96}
\newcommand{\optSpendTailored}{\$21}
\newcommand{\optCallsTailored}{265}
\newcommand{\optDiscrepancies}{0}
\newcommand{\hdN}{5}
\newcommand{\hdDeltaMin}{+1.4}
\newcommand{\hdDeltaMax}{+15.4}
\newcommand{\hdParamsMin}{2}
\newcommand{\hdParamsMax}{42}
\newcommand{\hdNSig}{5}
\newcommand{\hdDevCost}{\$0.38}
\newcommand{\hdTestCost}{\$0.00}
\newcommand{\hdMatresZero}{72.3}
\newcommand{\hdMatresHead}{79.1}
\newcommand{\hdMatresDelta}{+6.8}
\newcommand{\hdMatresParams}{20}
\newcommand{\hdMatresPaper}{83.2}
\newcommand{\hdTbdZero}{54.3}
\newcommand{\hdTbdHead}{62.6}
\newcommand{\hdTbdDelta}{+8.3}
\newcommand{\hdTbdParams}{5}
\newcommand{\hdTbdPaper}{68.6}
\newcommand{\hdHxZero}{55.6}
\newcommand{\hdHxHead}{61.1}
\newcommand{\hdHxDelta}{+5.5}
\newcommand{\hdHxParams}{2}
\newcommand{\hdHxPaper}{64.0}
\newcommand{\hdLatZero}{58.0}
\newcommand{\hdLatHead}{59.4}
\newcommand{\hdLatDelta}{+1.4}
\newcommand{\hdLatParams}{12}
\newcommand{\hdLatPaper}{69.0}
\newcommand{\hdImpZero}{41.9}
\newcommand{\hdImpHead}{57.2}
\newcommand{\hdImpDelta}{+15.4}
\newcommand{\hdImpParams}{42}
\newcommand{\hdImpPaper}{60.0}
\newcommand{\hdTbdVaguePred}{46}
\newcommand{\bodyRerunN}{6,324}
\newcommand{\bodyRerunFlips}{173}
\newcommand{\bodyRerunIdentical}{6,324}
\newcommand{\bodyRerunFlipPct}{2.7}
\newcommand{\bodyRerunChangedPct}{74}
\newcommand{\bodyRerunMedDiff}{0.02}
\newcommand{\bodyGeoBinaryN}{6}
\newcommand{\bodyGeoBinaryMatch}{5}
\newcommand{\bodyGeoBinaryBelow}{0}
\newcommand{\bodyGeoFewN}{12}
\newcommand{\bodyGeoFewMatch}{2}
\newcommand{\bodyGeoFewBelow}{3}
\newcommand{\bodyGeoManyN}{6}
\newcommand{\bodyGeoManyMatch}{2}
\newcommand{\bodyGeoManyBelow}{1}
\newcommand{\bodyGeoOrdinalN}{6}
\newcommand{\bodyGeoOrdinalMatch}{1}
\newcommand{\bodyGeoOrdinalBelow}{4}
\newcommand{\bodyGeoPairN}{4}
\newcommand{\bodyGeoPairMatch}{1}
\newcommand{\bodyGeoPairBelow}{0}
\newcommand{\bodyGeoSeqN}{2}
\newcommand{\bodyGeoSeqMatch}{1}
\newcommand{\bodyGeoSeqBelow}{0}
\newcommand{\bodyGeoLargeN}{5}
\newcommand{\bodyGeoLargeMatch}{0}
\newcommand{\bodyGeoLargeBelow}{5}
\newcommand{\bodyGeoConformalN}{5}
\newcommand{\bodyGeoConformalMatch}{0}
\newcommand{\bodyGeoConformalBelow}{4}

%% file: generated/hero_numbers.tex
\newcommand{\heroMarkers}{Half-filled: the rephrased question, chosen on training data. Filled: with training data, pre-registered and tested once (the eight optimized tasks: the optimized questions, mean over 3 optimizer seeds; MATRES, SMARTER HateXplain and SMARTER Latent: calibration or a logistic regression on the unchanged zero-shot answers; TTP, Wiki-WD, MADE: candidate reranking and fitted thresholds).}

%% file: generated/hero_steps.tex
\newcommand{\hsNSets}{13}
\newcommand{\hsJevPrompt}{0}
\newcommand{\hsJevPromptLo}{0}
\newcommand{\hsJevPromptHi}{0}
\newcommand{\hsJevCal}{23}
\newcommand{\hsJevCalLo}{8}
\newcommand{\hsJevCalHi}{40}
\newcommand{\hsJevOpt}{72}
\newcommand{\hsJevOptLo}{48}
\newcommand{\hsJevOptHi}{97}
\newcommand{\hsEnc}{59}
\newcommand{\hsEncLo}{44}
\newcommand{\hsEncHi}{88}
\newcommand{\hsLora}{97}
\newcommand{\hsLoraLo}{83}
\newcommand{\hsLoraHi}{107}

%% file: sections/introduction.tex
\section{Introduction}
\label{sec:intro}

\begin{figure}[t]
\centering
\includegraphics[width=\columnwidth]{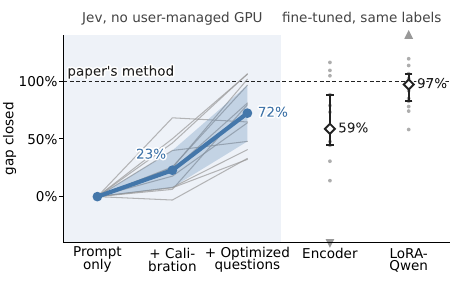}
\caption{Share of the gap between prompt-only \jev{} and the paper's method that calibration, optimized
questions and fine-tuning close on the \hsNSets{} configurations of \S\ref{sec:opt}. LoRA-Qwen is
Qwen3.8-27B fine-tuned with LoRA on the same labels. The blue line and band show the median and
interquartile range, and grey lines show individual configurations.}
\label{fig:page1}
\end{figure}

\input{sections/fig_pipeline}

Typed Decision Models, also called System One models \citep{typesafe2026jev}, are a new class of model
interface. A Typed Decision Model takes a JSON state and typed questions and returns a probability
for every option. TypeSafe's \jev{}, Cloudflare's open-weight Clef and Clef-flash, OpenAI's GPT-6 Luna Decisions (Luna for short) and
Microsoft-Decision-1 offer this interface
\citep{typesafe2026jev,cloudflare2026clef,openai2026luna,microsoft2026decision1}. Reading label
probabilities from a language model is not new. PET and LM-BFF turned cloze prompts into few-shot
classifiers by training on a few labels \citep{schick-schutze-2021-exploiting,gao-etal-2021-making},
while Typed Decision Models return such probabilities through an API, without any training. Many recent NLP
papers fine-tune a task-specific model that picks a label from a fixed set, places an item on a scale or
ranks given candidates, and prompted LLMs trail such models on most classification benchmarks
\citep{qin-etal-2023-chatgpt,kocon2023chatgpt,ziems2024css,bucher2024finetuned}. Typed Decision Models
raise the question of whether these tasks still need fine-tuned classifiers.

We take \specCore{} papers from ACL, EMNLP, NAACL and EACL 2024--2026 and replace each paper's \emph{predictor},
the model that outputs its label, score or ranking, with \jev{} 1.13, keeping the paper's splits, label definitions and metric. We send the same requests to the Typed Decision
Models of \nOtherVendors{} other vendors. We score \nDatasets{} English test sets from \nPapersScored{} of
the papers (Explanation ranking has no comparable test set, Table~\ref{tab:deviations}). We call these test
sets, together with their questions, metrics and published scores, \emph{\suite{}}. The experiments with
labels use \atN{} \emph{configurations} from 11 of these test sets, because DREsS is scored separately for each
of its three rubrics. The model itself is never fine-tuned
(Figure~\ref{fig:pipeline}). When labels are available, we add calibration and optimized questions whose
answers serve as features for a small logistic regression, as in Text Bottleneck Models
\citep{ludan2023tbm}. To match the papers' effort on their own methods, we optimize the questions
with a GEPA-style search \citep{agrawal2025gepa} over typed questions. We
compare against the papers' methods and, on the same labels, against fine-tuned encoders and LoRA
fine-tuning \citep{hu2022lora} of an open LLM. Throughout, we measure each step by the share
of the gap between prompt-only \jev{} and the paper's method that it closes (Figure~\ref{fig:page1}).

\textbf{A prompt alone is enough on about a quarter of the test sets.} We
say that a decision model \emph{directly replaces} a predictor when, with a prompt and no labels, it
matches or beats the paper's method on the paper's own metric and test split (\S\ref{sec:protocol}).
\jev{} directly replaced the predictor on \nReplaces{} of the \nDatasets{} test sets
(\S\ref{sec:results}). These are mostly single judgments with \emph{plain labels}, whose names are enough
to apply them, and rarely test sets where the paper's method was trained on the paper's labels.

\textbf{Where a prompt falls short, the label name most often leaves out an annotation convention.} We
assign this cause on \csConv{} of the \csN{} test sets where \jev{} scores below the paper's method
(\S\ref{sec:failures}). \citet{rao2026jev} observed the same for rating conventions. In SMARTER Implicit Hate, for
example, annotators use ``Incitement to Violence'' for posts that promote in-group supremacy and hate groups, which the label name does not say. Clef, Luna and Microsoft-Decision-1 mostly fall into the same outcome group as \jev{},
and they are also below the paper's method on the test sets
where we attributed the shortfall to conventions.

\textbf{With \optTrainN{} training labels and a validation split, optimized questions close most of the gap and LoRA closes almost all of it.}
The median share of the gap closed is \atOptGap\% for calibration and optimized questions, which need no
user-managed GPU, and \atLoraGap\% for LoRA fine-tuning of the open LLM Qwen3.8-27B \citep{qwen2026models}
(\S\ref{sec:budget}). An encoder fine-tuned on the same labels does about as well as the optimized questions. The
search closes about twice as much of the gap as off-the-shelf GEPA, and differs from it most in that it
scores a question after fitting a logistic regression on its answers (\S\ref{sec:fqs}). The same
questions improve every other Typed Decision Model we tested, and Qwen3.8-27B, scored from its option
log-probabilities, comes close to Clef, the decision model post-trained from it (\S\ref{sec:opt-clef}).

\Uponacceptance{} will release \suite{} with the code, prompts, test manifests and responses, and the optimizer as \jevopt{}\releaseat.

\section{Background: Two Pipelines}
\label{sec:background}

In the \emph{fine-tuning pipeline}, a task-specific model learns the label set, input format and
annotation conventions from the paper's training split, and baselines, often a prompted LLM, show how
much this training adds. In the \emph{decision-model pipeline}, the model is never fine-tuned. It starts from a \emph{base question} that we write from the paper's own definitions (\S\ref{sec:protocol}), then adds \emph{calibration}, which fits per-class biases or a temperature,
and then \emph{optimized questions}, rewritten questions whose answers serve as features
(\S\ref{sec:opt}). \jev{} takes a JSON \emph{state} and questions of three types. A \textsc{Choice}
question has up to 255 named options, each with a criterion, a \textsc{Score} question has 2--10
ordered levels, and a \textsc{Noul} is a yes/no question answered with $P(\text{yes})$. Questions in one
request share the state and are answered independently. Probabilities are rounded to two decimals, and
\textsc{Choice} and \textsc{Score} answers also report a confidence. We use \jev{} 1.13 (snapshot
2026-09-17 for every request) through OpenRouter, billed per input token. Clef accepts the same requests,
and the yes/no, choice and score questions of Luna and Microsoft-Decision-1 map onto these types.

%% file: sections/fig_pipeline.tex
\begin{figure*}[t]
\centering
\begingroup
\newcommand{\kv}[2]{\begin{tabular}{@{}l@{\ }p{#1}@{}}#2\end{tabular}}
\newcommand{\key}[1]{\textcolor{black!60}{#1:}}
\hyphenpenalty=10000 \exhyphenpenalty=10000
\begin{tikzpicture}[x=1mm, y=1mm, >={Stealth[length=3pt]},
  font=\fontsize{7.5}{9}\selectfont,
  box/.style={draw, rounded corners=1.5pt, align=left, inner xsep=2pt, inner ysep=2pt, anchor=north west},
  ft/.style={box, draw=pipeGray, fill=pipeGrayFill, line width=0.5pt},
  dmA/.style={box, draw=pipeBlue, fill=pipeBlueFill, line width=0.5pt},
  dmB/.style={box, draw=pipeBlue, fill=pipeBlue!14, line width=0.7pt},
  dmC/.style={box, draw=pipeBlue, fill=pipeBlue!24, line width=0.9pt},
  head/.style={align=left, anchor=north west, inner sep=0pt, font=\fontsize{8}{9.5}\selectfont},
  arr/.style={->, line width=0.5pt, draw=black!75}]
  \node[head, text=pipeGray!70!black] at (0,0) {\textbf{Fine-tuning}\\\textbf{pipeline}\\(the papers)};
  \node[ft, text width=25.5mm] (f1) at (20.5,0) {\textbf{Baselines}\\incl.\ a prompted LLM};
  \node[ft, text width=88.5mm] (f2) at (51,0) {\textbf{Fine-tuning}\\
    \kv{80mm}{\key{labels} & the paper's full training split\\ \key{trains} & the model's weights}};
  \node[head, text=pipeBlue] at (0,-14) {\textbf{Decision-model}\\\textbf{pipeline}\\(\jev{} is never\\fine-tuned)};
  \node[dmA, text width=25.5mm] (d1) at (20.5,-14) {\textbf{1 Prompt}\\
    \kv{16mm}{\key{labels} & none\\ \key{fits} & nothing}};
  \node[dmB, text width=40mm] (d2) at (51,-14) {\textbf{2 + Calibration}\\
    \kv{32.5mm}{\key{labels} & \optFitN{} (100 suffice for calibration error)\\ \key{fits} & class biases or a temperature}};
  \node[dmC, text width=43.5mm] (d3) at (96,-14) {\textbf{3 + Optimized questions}\\
    \kv{35mm}{\key{labels} & \optTrainN{} and a validation split\\
      \key{search} & guided by the labels\\
      \key{fits} & a small logistic regression on \jev{}'s answers\\
      \key{cost} & median \$\frCostFqMedian{} per search}};
  \node[draw=black!70, fill=white, rounded corners=1.5pt, line width=0.6pt, inner sep=0pt,
    fit={(144.5,0) (144.5,0 |- d3.south) (160,0)}] (t) {};
  \node[align=center, text width=14mm, inner sep=0pt] at (t.center) {\textbf{The paper's test set and metric}};
  \draw[arr] (f1.east) -- (f1.east -| f2.west);
  \draw[arr] (f2.east) -- (f2.east -| t.west);
  \draw[arr] (d1.east) -- (d1.east -| d2.west);
  \draw[arr] (d2.east) -- (d2.east -| d3.west);
  \draw[arr] (d3.east) -- (d3.east -| t.west);
\end{tikzpicture}
\endgroup
\caption{The two pipelines compared in this paper: the papers' fine-tuned methods and \jev{}, which is never
fine-tuned and uses labels only for calibration and question optimization.}
\label{fig:pipeline}
\end{figure*}
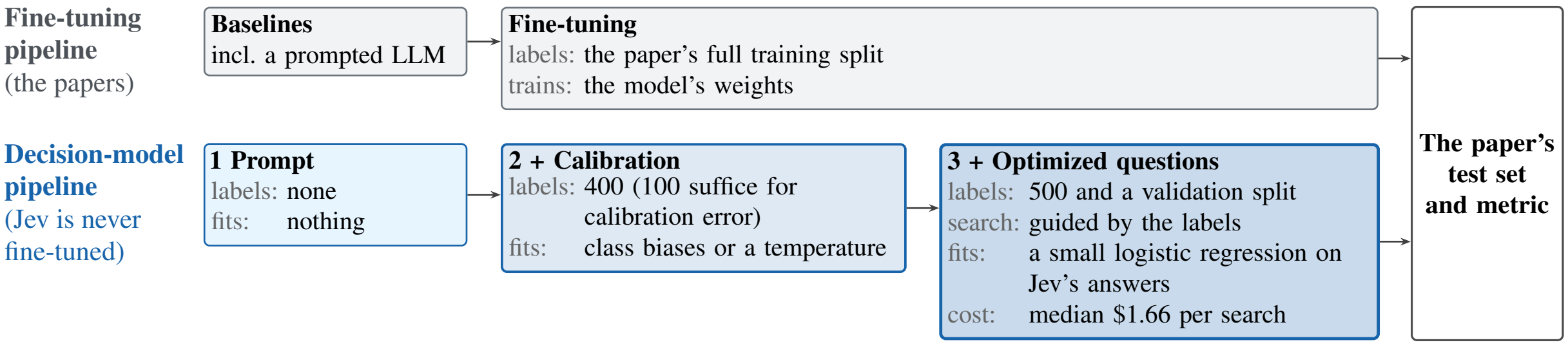

%% file: sections/setup.tex
\section{Protocol: Swapping the Predictor}
\label{sec:protocol}

We fixed each English test set and its question before querying the model on any test item. We run every
test item, score it with the paper's primary metric, and take each published number from a cited table and
page. Table~\ref{tab:deviations} lists every deviation from a paper's protocol, and we recorded all raw
responses.

\input{generated/body_tasks}
\input{generated/main_table}

\paragraph{Sample.}
We screened the titles, abstracts, PDFs and releases of \specInScope{} main-conference and Findings
papers of ACL, EMNLP, NAACL and EACL from January 2024 to September 2026
(Appendix~\ref{app:selection}). A paper is eligible if the output we replace comes from a fixed label
set, scale or candidate set and replacing it tests a substantive prediction claim. We must be able to use
or closely reconstruct its split and metric, and its labels, inputs and test data must be described
and released well enough to prompt a model. We fixed the \specCore{} papers before seeing any model
output and chose them to cover different output types, domains and levels of difficulty
(Table~\ref{tab:tasks}), with \specStress{} of them chosen as likely hard cases. All rows sent to the model are English, since FENICE, CICC
and Navigating the Unknown are restricted to their English subsets and we removed rows that language
identification flagged (at most 1.7\% of a test set).

\paragraph{Questions from the paper's own definitions.}
For each task we write one question that asks for the label, and we call it the \emph{base question}. Its instructions come from the paper's
zero-shot LLM prompt or its annotation guidelines, its option criteria quote the paper's label
definitions verbatim, and we record the source of every sentence. Questions are
zero-shot, even where the paper's LLM baselines were few-shot. Nothing is tuned on test labels, and
thresholds are fixed in advance (0.5 for \textsc{Noul}s) or fit on the development split where the
paper does the same (BIORC800, FENICE). The model gets the same information as the predictor it
replaces, for example label names only where that predictor saw none.

\paragraph{Outputs that are not one question.}
\label{sec:fitting}
Where the predictor's output does not fit one question, we decompose it. Wiki-WD's \audRelembWikiwdK{}
relations are asked in two halves plus a final question, MADE's \audMadeK{} codes are asked top-down
along the code hierarchy, and TTP mapping ranks all \audTtpTramK{} ATT\&CK techniques by
$P(\text{yes})$. For PRP-Graph, the paper's tournament code aggregates pairwise answers, and LegalSeg and
BIORC800 ask one question per sentence with the whole document in the state.

\paragraph{How we compare.}
We place each test set in one of three outcome groups by comparing the scores as published, on the paper's
primary metric ($\times$100). In the first group, \jev{} \emph{matches or beats} the paper's method,
meaning that it scores higher or at most 0.5 points lower. Only then do we say that it directly replaces
the predictor. In the second group, \jev{} is \emph{between} the paper's prompted-LLM baseline and the
paper's method. It falls short of the paper's method and, by the same 0.5-point rule, matches or beats the LLM
baseline. That baseline is the paper's best zero-shot prompted LLM, or its best few-shot one if it reports no
zero-shot one. Some papers report no LLM baseline. For those, \jev{} is in the second group when it falls
short of the paper's method and reaches at least 80\% of its score. In the third group, \jev{} is \emph{below both}.
Of the \nDatasets{} test sets, \ovRqOneLlmZsReplaces{} fall into the first group, \ovRqOneLlmZsBetween{} into the second and
\ovRqOneLlmZsBelow{} into the third (Appendix~\ref{app:rq1} gives the counts for another choice of LLM
baseline).

\paragraph{Significance.} The papers publish their scores without intervals, so the groups compare point
estimates and ignore sampling noise. We therefore also resample \jev{}'s test items (\ovRqOneB{}
replicates, by cluster where items share a source) and test each difference from the published score.
\jev{} is \emph{significantly above or below} the paper's method when the difference is
significant after Holm correction over the \nDatasets{} test sets, and at \emph{parity} when an equivalence
test with a \ovRqOneTolPts-point margin is significant. The remaining test sets are \emph{inconclusive}.
The groups ask whether \jev{}'s score is as high as the published one, and the tests ask whether the
difference can be told apart from noise. A test set can therefore be in the first group without being at
parity. The tests are
adapted from \citet{rao2026jev}, who use a margin of 5 points. Where the published number was computed on
test items other than ours (Table~\ref{tab:deviations}), we treat it as uncertain too and assume it has
\jev{}'s sampling variance (Appendix~\ref{app:rq1}).

%% file: generated/body_tasks.tex
\begin{table}[t]
\centering\scriptsize
\setlength{\tabcolsep}{2.2pt}
\resizebox{\ifdim\width>\columnwidth\columnwidth\else\width\fi}{!}{%
\begin{tabular}{@{}lcccc@{}}
\toprule
Paper & Domain & Output & \jev{} interface & Test \\
\midrule
PiNose~\citeyearpar{zhang-etal-2024-transferable} & factuality & binary & Noul & R \\
FENICE~\citeyearpar{scire-etal-2024-fenice} & factuality & binary & Noul per claim & P \\
FactPICO~\citeyearpar{joseph-etal-2024-factpico} & medicine & ordinal 0--4 & Score each & R \\
XSTest~\citeyearpar{rottger-etal-2024-xstest} & safety & 3-way & Choice & R \\
LifeTox~\citeyearpar{kim-etal-2024-lifetox} & safety & binary & Choice & R \\
PRP-Graph~\citeyearpar{luo-etal-2024-prp} & biomed.\ IR & rank 100 & Choice, tournament & E \\
Scalar Constructs~\citeyearpar{licht-etal-2025-measuring} & politics & pairwise; 1--9 & Choice; Score & R \\
BIORC800~\citeyearpar{lan-etal-2024-multi} & biomed. & 6-label seq. & Noul, sent.\,$\times$\,label & R \\
SciPrompt~\citeyearpar{you-etal-2024-sciprompt} & science & 7--53-way & Choice & E \\
Re3-Sci edit intent~\citeyearpar{ruan-etal-2024-large} & science & 5-way & Choice, routed & R \\
EZ-STANCE~\citeyearpar{zhao-caragea-2024-ez} & social media & 3-way & Choice, routed & E \\
ConTempo~\citeyearpar{niu-etal-2024-contempo} & news & 4-/6-way & Choice & I \\
MISMATCHED~\citeyearpar{shaik-etal-2025-mismatched} & science & 4-way & Choice & R \\
INLI~\citeyearpar{havaldar-etal-2025-entailed} & general & 4-way & Choice & R \\
MedReadMe~\citeyearpar{jiang-xu-2024-medreadme} & medicine & ordinal 1--6 & Score & R \\
\begin{tabular}[t]{@{}l@{}}Navigating the\\\quad Unknown~\citeyearpar{sali-toraman-2025-navigating}\end{tabular} & dialogue & 39--76-way & Choice & C \\
MADE~\citeyearpar{agarwal-etal-2026-made} & medicine & 1,154-label set & Noul, top-down & R \\
SMARTER~\citeyearpar{nghiem-etal-2026-smarter} & social media & 3-/6-way & Choice & C \\
DREsS~\citeyearpar{yoo-etal-2025-dress} & education & 3 ordinal & Score per rubric & C \\
TTP mapping~\citeyearpar{nguyen-etal-2024-noise} & security & rank 737 & Noul each & R \\
LegalSeg~\citeyearpar{nigam-etal-2025-legalseg} & law & 7-way seq. & Choice per sent. & E \\
\begin{tabular}[t]{@{}l@{}}Entity or Relation\\\quad Embeddings~\citeyearpar{mtumbuka-schockaert-2024-entity}\end{tabular} & general & 56-/352-way & Choice, split & E \\
CICC~\citeyearpar{hengst-etal-2024-conformal} & dialogue & conformal set & Choice & R \\
\begin{tabular}[t]{@{}l@{}}Explanation\\\quad ranking~\citeyearpar{bailleux-etal-2026-explanation}\end{tabular} & NLI expl. & rank 5 & Score each & C \\
\bottomrule
\end{tabular}}
\caption{The \specCore{} papers in our sample and how \jev{} replaces each paper's predictor (Appendix~\ref{app:catalogue} describes the data of each paper with an example). In the Test column, R is the released test set, E the released set without non-English rows, P a partial release, I a rebuilt input pipeline and C a rebuilt test set or split.}
\label{tab:tasks}
\end{table}

%% file: generated/main_table.tex
\begin{table*}[t]
\centering\footnotesize
\setlength{\tabcolsep}{4pt}
\resizebox{\ifdim\width>\textwidth\textwidth\else\width\fi}{!}{%
\begin{tabular}{lccccccc}
\toprule
Dataset & Output & $N$ & Metric & Jev & Paper's method & LLM & $\Delta$ \\
\midrule
\multicolumn{8}{l}{\textbf{\textit{Jev matches or beats the paper's method}}} \\
FENICE~\citep{scire-etal-2024-fenice} & binary & 279 & bal.\ acc. & \cellcolor{green!15}\textbf{77.4} & 65.7 & \textcolor{gray}{--} & \textcolor{ForestGreen}{+11.7} \\
PiNose True-False~\citep{zhang-etal-2024-transferable} & binary & 1,000 & AUC & \cellcolor{green!15}\textbf{96.9} & 86.6 & \textcolor{gray}{--} & \textcolor{ForestGreen}{+10.3} \\
SciPrompt arXiv~\citep{you-etal-2024-sciprompt} & 53-way & 5,295 & acc. & \cellcolor{green!15}\textbf{79.0} & 70.2 & 54.9 & \textcolor{ForestGreen}{+8.9} \\
XSTest (judge)~\citep{rottger-etal-2024-xstest} & 3-way & 2,250 & acc. & \cellcolor{green!15}\textbf{87.0} & 80.4 & 80.4 & \textcolor{ForestGreen}{+6.6} \\
FactPICO~\citep{joseph-etal-2024-factpico} & ordinal & 345 & Kendall $\tau_b$ & \cellcolor{green!15}\textbf{53.4} & 47.4 & 47.5 & \textcolor{ForestGreen}{+6.0} \\
PRP-Graph TREC-COVID~\citep{luo-etal-2024-prp} & ranking & 50 & nDCG@10 & \cellcolor{green!15}\textbf{82.1}$^\dagger$ & 78.8 & \textcolor{gray}{--} & \textcolor{ForestGreen}{+3.3} \\
BIORC800~\citep{lan-etal-2024-multi} & multi-label & 160 & micro-F1 & \cellcolor{green!15}\textbf{90.8}$^\dagger$ & \textbf{90.8} & 47.6 & \textcolor{ForestGreen}{0.0} \\
\midrule
\multicolumn{8}{l}{\textbf{\textit{Jev between the prompted LLM and the paper's method}}} \\
CICC MTOD~\citep{hengst-etal-2024-conformal} & conformal set & 8,665 & Single & 97.3 & \textbf{100.0} & \textcolor{gray}{--} & \textcolor{BrickRed}{$-$2.7} \\
MISMATCHED~\citep{shaik-etal-2025-mismatched} & 4-way & 2,400 & macro-F1 & 71.2 & \textbf{78.2} & 55.6 & \textcolor{BrickRed}{$-$7.0} \\
SMARTER HateXplain~\citep{nghiem-etal-2026-smarter} & 3-way & 1,193 & macro-F1 & 55.6 & \textbf{64.0} & 56.0 & \textcolor{BrickRed}{$-$8.4} \\
\begin{tabular}[t]{@{}l@{}}Navigating the Unknown\\\quad CLINC150 (ID+OOD)~\citep{sali-toraman-2025-navigating}\end{tabular} & 76-way & 1,500 & macro-F1 & 86.0 & \textbf{95.7} & 83.7 & \textcolor{BrickRed}{$-$9.7} \\
ConTempo MATRES~\citep{niu-etal-2024-contempo} & 4-way & 837 & micro-F1 & 72.3 & \textbf{83.2} & 56.7 & \textcolor{BrickRed}{$-$10.9} \\
Scalar Constructs: imm.\ fear (pair)~\citep{licht-etal-2025-measuring} & pairwise & 3,265 & Spearman $\rho$ & 78.7 & \textbf{91.0} & 74.0 & \textcolor{BrickRed}{$-$12.3} \\
INLI~\citep{havaldar-etal-2025-entailed} & 4-way & 4,000 & acc. & 76.8 & \textbf{92.4} & 77.0 & \textcolor{BrickRed}{$-$15.7} \\
Re3-Sci edit intent~\citep{ruan-etal-2024-large} & 5-way & 2,312 & macro-F1 & 66.7 & \textbf{84.3} & 37.0 & \textcolor{BrickRed}{$-$17.6} \\
LifeTox~\citep{kim-etal-2024-lifetox} & binary & 3,000 & macro-F1 & 78.4 & \textbf{96.5} & 74.4 & \textcolor{BrickRed}{$-$18.1} \\
EZ-STANCE A~\citep{zhao-caragea-2024-ez} & 3-way & 7,666 & macro-F1 & 60.7 & \textbf{81.2} & 49.9 & \textcolor{BrickRed}{$-$20.5} \\
LegalSeg~\citep{nigam-etal-2025-legalseg} & 7-way seq. & 149,875 & macro-F1 & 52.0 & \textbf{77.0} & 8.0 & \textcolor{BrickRed}{$-$25.0} \\
\midrule
\multicolumn{8}{l}{\textbf{\textit{Jev below both}}} \\
MedReadMe~\citep{jiang-xu-2024-medreadme} & ordinal & 1,140 & Pearson $r$ & 71.4 & \textbf{83.3} & 72.1 & \textcolor{BrickRed}{$-$11.9} \\
MADE~\citep{agarwal-etal-2026-made} & 1,154 multi-label & 10,288 & macro-F1 & 32.9 & \textbf{54.0} & 43.0 & \textcolor{BrickRed}{$-$21.1} \\
TTP mapping TRAM~\citep{nguyen-etal-2024-noise} & 737 ranking & 725 & P@1 & 44.0 & \textbf{77.0} & \textcolor{gray}{--} & \textcolor{BrickRed}{$-$33.0} \\
\begin{tabular}[t]{@{}l@{}}Entity or Relation Embeddings\\\quad Wiki-WD~\citep{mtumbuka-schockaert-2024-entity}\end{tabular} & 352-way & 739,977 & micro-F1 & 54.1 & \textbf{89.1} & \textcolor{gray}{--} & \textcolor{BrickRed}{$-$35.0} \\
DREsS~\citep{yoo-etal-2025-dress} & ordinal & 401 & QWK & 20.6 & \textbf{69.1} & 30.4 & \textcolor{BrickRed}{$-$48.5} \\
\bottomrule
\end{tabular}}
\caption{Zero-shot Jev, prompted with the paper's label definitions, compared with each paper's method and its prompted-LLM baseline (LLM) on one test set per paper. Rows are grouped by outcome as defined in \S\ref{sec:protocol}, and Table~\ref{tab:full} lists all test sets. $^\dagger$ Scored under the paper's protocol.}
\label{tab:main}
\end{table*}

%% file: sections/results.tex
\section{Which Predictors a Prompt Replaces}
\label{sec:results}

With the 0.5-point rule of \S\ref{sec:protocol}, \jev{} directly replaced the predictor on \nReplaces{}
of the \nDatasets{} test sets (Table~\ref{tab:main}, all test sets in Table~\ref{tab:full}). Tested
against the published scores, \jev{} is significantly above the paper's method on \ovRqOneSepAbove{} test sets,
at parity on \ovRqOneParity, significantly below on \ovRqOneSepBelow{} and inconclusive on
\ovRqOneInconcl{} (Appendix~\ref{app:rq1}). It replaced the predictor on \bodyGeoBinaryMatch{} of the \bodyGeoBinaryN{} test sets with two labels and on
none of the \bodyGeoLargeN{} with over 300 possible labels. On the median test set, \jev{} reaches
\bodyMedRelAllPct\% of the paper's score. Its median cost is \bodyCostPerKMedian{} per 1,000 test items.

\begin{figure*}[t]
\centering
\includegraphics[width=\textwidth]{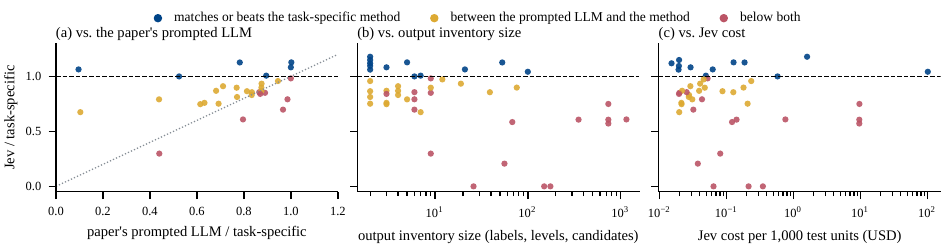}
\caption{\jev{}'s score as a fraction of the score of the paper's method (\emph{task-specific} in the axis labels) on the \nDatasets{} test sets.
Panels (a)--(c) plot it against the same fraction for the prompted LLM, the size of the output inventory
and \jev{}'s cost.}
\label{fig:factors}
\end{figure*}

\paragraph{Which methods a prompt can replace.}
Most matches are against methods that were not trained on the task's human labels. Of the
\ovRqOneNTrained{} test sets whose method was trained on the paper's labelled data, \jev{} replaced
\ovRqOneMatchTrained{}, among them two SciPrompt test sets with few-shot prompt-tuned models
(Appendix~\ref{app:rq1}). \jev{} is better than the paper's prompted LLM, or at most 0.5 points below it, on
\ovRqOneJevGeLlmZsTol{} of \ovRqOneNLlm{} test sets. The predictors it matched judge factuality,
refusal, pairwise relevance or topics with human-readable names.

\paragraph{Typed Decision Models from other vendors.}
To check the pattern, we sent \jev{}'s requests unchanged to Clef
\citep{cloudflare2026clef,chen2026clef}, Cloudflare's open 27B decision model built for \jev{}'s
interface, to GPT-6 Luna Decisions \citep{openai2026luna} and to Microsoft-Decision-1
\citep{microsoft2026decision1}. We skipped Wiki-WD because of its cost. On the other \ocNSets{} test sets,
Clef falls into the same outcome group as \jev{} on \ocSameGroup, Luna on \luSameGroup{} and
Microsoft-Decision-1 on \msSameGroup. All three stay below the paper's method on
every convention test set we ran them on (\S\ref{sec:failures}, Appendix~\ref{app:vendors}).

\subsection{What a Prompt Misses}
\label{sec:failures}

For each of the \csN{} test sets where \jev{} scores below the paper's method, we assigned one main
cause (Appendix~\ref{app:causes}). An LLM drafted the causes and an author checked them (agreement
$\kappa=\csKappa$, 95\% interval \csKappaLo--\csKappaHi). The most common cause, on \csConv{} test sets
from \csConvPapers{} of the \csNPapers{} papers, is an annotation convention that the label name does not
state. Large label spaces account for \csLarge{} test sets, coarse probabilities for \csCoarse{} and other
or unclear causes for \csOther.

\paragraph{(a) Labels that encode annotation conventions.}
Some labels mean more than their names say, because the annotators followed a rule that the paper does
not state. \jev{} sees only the label name, so it misses such rules. In the SMARTER Implicit Hate example
of \S\ref{sec:intro}, \jev{} reads ``Incitement to Violence'' literally and gives it to \bodySmIncPred{} test
posts, while \bodySmPerClass{} carry it in the gold labels. TB-Dense's VAGUE class and EZ-STANCE's neutral
class fail in the same way. Two known habits of \jev{} may add to these errors. It tends to compress
ordinal scales \citep{gao2026choices}, which may matter for the DREsS essay scores. It can also follow an
option's name rather than the rubric attached to it \citep{sun2026typesafe}, like some open models with
the same interface \citep{azizi2026labels}.

\paragraph{(b) Large opaque label spaces.}
\jev{} also does worse when it must choose among many answers that it knows only by name. Its share of
the paper's score falls as the number of possible answers grows ($\rho=\bodyRhoK$,
Figure~\ref{fig:factors}b), from a median of \bodyMedRelSmallK{} with at most ten answers to
\bodyMedRelLargeK{} with a hundred or more. In TTP mapping, for example, the model must rank
\audTtpTramK{} ATT\&CK attack techniques, which \jev{} knows only from their names and descriptions,
while a trained model learns them from labelled examples.

\paragraph{(c) Probabilities too coarse for the paper's method.}
Some methods need very small probabilities, which \jev{} cannot give because it rounds every probability
to two decimals. CICC uses conformal prediction \citep{angelopoulos2021gentle} to turn a model's
probabilities into a set of possible intents, and its ``Single'' metric is the share of requests left
with one intent. This works only if the true intent gets a probability above zero on almost all
calibration items. When \jev{} is confidently wrong, rounding gives the true intent a probability of
zero. On CLINC150 this happens on \bodyCiccClincZero\% of calibration items, more than the
\bodyCiccClincAlpha\% that the method allows, so every set contains all intents and the ``Single'' metric is 0,
although top-1 accuracy is \bodyCiccClincAcc. Microsoft-Decision-1 does not round to two decimals.
It leaves one intent on \msCiccClinc\% of CLINC150 items, while on ACID and ATIS, where \jev{}'s sets
also contain all intents, it does so for few items. Rounding is therefore the main cause only on CLINC150. On ACID the main cause
is the large label space, and on ATIS an unstated convention (Table~\ref{tab:causes}).

\paragraph{Testing the convention account.}
If unstated conventions cause errors, stating them should help. We tested the \cvSpans{} sentences that
state a convention in \cvPrograms{} of the optimized question sets of \S\ref{sec:opt}
(Appendix~\ref{app:conventions}). Adding them to the base question raises its calibrated score by
\cvCalAbs{} points (95\% interval \cvCalLo{} to \cvCalHi), so stating conventions helps beyond what
calibration to the class frequencies does. Deleting them from the
optimized questions lowers the score by \cvDelAbs{} points, which is not significant after Holm correction
($p=\cvDelP$) and does not differ significantly from deleting the same amount of other text. The gain
from adding them is also small
next to the \cvAZAbs{} points that the full optimized questions add (\S\ref{sec:opt}).

\subsection{Answering Only Confident Items}
\label{sec:confidence}

\jev{} reports a confidence with each answer, so it can answer only the items where its confidence passes
a threshold and skip the rest. Learn-then-Test \citep{angelopoulos2021ltt} picks the lowest threshold at
which the accuracy of the answered items is above the paper's number with high probability. We choose the
threshold on one random half of each test set and score the other half, over \cfSelHalvings{} halvings.
We use the \cfSelNComparable{} test sets where the paper reports accuracy or P@1. For PiNose WebQ we
therefore compare accuracy, not the AUC of Table~\ref{tab:full}. When it answers every item, \jev{} is
below the paper's method on \number\numexpr\cfSelNComparable-\cfSelNFull\relax{} of these test sets. On three of
them (INLI, SciPrompt S2ORC and PiNose WebQ), answering only confident items brings \jev{} to the paper's
number in most halvings, while it still answers \cfSelLttCovInli--\cfSelLttCovPinoseWebq\% of the items.
On the three TTP sets, Learn-then-Test finds no such threshold.

%% file: sections/optimizing_decision_.tex
\section{What Labels Add, and at What Cost}
\label{sec:opt}

We test how much \optTrainN{} training labels plus validation improve \jev{}, and compare with
fine-tuning on the same labels (Table~\ref{tab:answer}). Of the \atN{} configurations we use, \optNTasks{} were chosen for large convention-driven
gaps (Table~\ref{tab:opt}), and the other \fzNewNSets{} were selected by a rule after the search was final
(\S\ref{sec:frozen}). Each runs with \optNSeeds{} optimizer seeds on the paper's validation split and fixed
test set (Appendix~\ref{app:opt-data}).

\paragraph{Calibration.}
Calibration helps where the gap comes from a class prior. Per-class biases, fit on \optFitN{} training
examples without changing the base question, raise TB-Dense from \fzTbDenseZero{} to
\fzTbDenseCal{} (Table~\ref{tab:frozen}). A single temperature \citep{guo2017calibration} fit on \optFitN{} items lowers the
expected calibration error from \cfRoEceZero{} to \cfRoEceTs.

\subsection{Prompt Optimization for Typed Decision Models}
\label{sec:fqs}

The papers tuned their own methods, so a fair comparison also optimizes the decision model's
questions. Off-the-shelf prompt optimizers score a prompt by the label it elicits
\citep{khattab2024dspy,agrawal2025gepa}. With a probability for every option, the most
useful question is often one whose answers separate the labels, whether or not \jev{}'s own choice is
right. We therefore build a GEPA-style search for Typed Decision Models in \jevopt{}
(Appendix~\ref{app:opt-ir}). Unlike GEPA, which tests one candidate at a time, it asks many candidate
questions in one request. Questions are typed programs of \textsc{Choice}, \textsc{Score} and
\textsc{Noul} nodes, and a proposed question can be of another type. Every candidate is scored after
fitting calibration or a logistic regression on its answers, like the concept layer of Text Bottleneck
Models \citep{ludan2023tbm}. A question may also show \jev{} 8 labelled training examples similar to the
input, balanced across labels. With optimizer seed 0, the selected questions do so on
\number\numexpr\atN-\lcNSetsNoDemo\relax{} of the \atN{} configurations. The
search (Figure~\ref{fig:fqs}) starts from the base question and from questions that an LLM wrote from
the task definition alone, with short, literal criteria that separate neighbouring options. The LLM
either rewrites the base question or changes its question type or structure. We call the best
rewritten question, run as written, the \emph{rephrased question}.

Fitting before scoring is the most important difference from GEPA. Off-the-shelf GEPA scores the label
the model picks. On the \optNTasks{} configurations, its text search closes a median \atGepaTextGap\% of
the gap and its program search \atGepaProgGap\%, against \atOptGapCfg\% for the optimized questions. GEPA
with demonstrations and calibration (GEPA+d/c in the tables), which
also fits before scoring, closes \atGepaDemoGap\% and is not
significantly different from the optimized questions (pooled difference \ovRqThreePooledDiff{} points,
95\% interval \ovRqThreePooledLo{} to \ovRqThreePooledHi, Appendix~\ref{app:opt-others}).

\begin{figure}[t]
\centering
\fbox{\begin{minipage}{0.96\columnwidth}\scriptsize
\textbf{Input:} the base question, questions an LLM rewrote or restructured from the task definition
alone, \optTrainN{} labelled training examples (\optFitN{} for fitting), validation split.
\begin{list}{\arabic{enumi}.}{\usecounter{enumi}\setlength{\leftmargin}{1.1em}\setlength{\labelwidth}{0.9em}%
  \setlength{\labelsep}{0.2em}\setlength{\itemsep}{0pt}\setlength{\parsep}{0pt}\setlength{\topsep}{1pt}}
\item Ask each question on the \optFitN{} fitting examples, with the questions of one example in one request.
\item For each question, fit a logistic regression on its answers and score it by 5-fold cross-validation.
\item For 4 rounds, an LLM sees the 4 best questions, a per-class summary and the 8 worst out-of-fold
  examples and proposes 20 different variants, each of which may ask for 8 similar labelled training
  examples per input. Ask them all, packed as in step~1, then fit and score.
\item Score combinations of questions offline from out-of-fold answers. The 4 best sets of
  questions, and stacks of 2--4 of them, go to validation.
\item Keep a candidate only if it gains at least 1 validation point, and test the chosen one once.
\end{list}
\end{minipage}}
\caption{The question optimization procedure. Appendix~\ref{app:opt} summarizes the protocol, and
Table~\ref{tab:opt-settings} lists the settings.}
\label{fig:fqs}
\end{figure}

\input{generated/body_opt}

\paragraph{Results.}\label{sec:opt-results}
The optimized questions improve every configuration over the base question
(Table~\ref{tab:opt}). When both feed the same logistic regression, they still beat the base question
on validation on \frAblNPos{} of \frAblNTasks{} configurations (Table~\ref{tab:ablation}), and they beat
the paper's prompted LLM on \optNBeatLLM. Of the \frConvTotal{} selected question sets (\optNSeeds{} seeds
on each of the \optNTasks{} configurations), \frConvN{} state
a convention of the dataset that the base question does not.

\subsection{Rule-Selected Sets and Labels}
\label{sec:opt-what}

\paragraph{All eligible shortfall test sets.}\label{sec:frozen}
With the search final, we ran it with three optimizer seeds on every test set where \jev{}
is below the paper's method and that meets the conditions of Appendix~\ref{app:frozen}, such as at most 40
labels and a released training set. These are ConTempo MATRES, LifeTox,
SciPrompt S2ORC, SMARTER HateXplain and SMARTER Latent Hate. On these five,
calibration closes a median \atCalGapFrozen\% of the gap and the optimized questions again
\atOptGapFrozen\%, ranging from \atOptGapFrozenMin\% to \atOptGapFrozenMax\%.

\paragraph{Labels.}
The search fits on \optFitN{} of the \optTrainN{} training labels, takes its labelled
examples from all \optTrainN{}, and selects on \optValMin--\optValMax{} validation labels, which makes
\optLabelsMin--\optLabelsMax{} labels in all. Once the
questions are fixed, calibration gains little beyond about 100 labels (Table~\ref{tab:curve} in
Appendix~\ref{app:opt}).

\subsection{Fine-Tuning with the Same Labels}\label{sec:budget}

On all \atN{} configurations, calibration and optimized questions, which need no user-managed GPU, close a
median \atOptGap\% of the gap (95\% interval \atOptGapAllLo--\atOptGapAllHi), a fine-tuned encoder
\atEncGap\% and LoRA fine-tuning \atLoraGap\% (Table~\ref{tab:answer}).

\begin{table}[t]
\centering\small
\setlength{\tabcolsep}{3pt}
\resizebox{\columnwidth}{!}{\input{generated/answer_table}}
\caption{Labels, build cost and results of each way to build a classifier on the \atN{} configurations of
\S\ref{sec:opt}. Gap closed is the median share of the gap to the paper's method, measured from the
prompt-only scores in the Zero column of Table~\ref{tab:frozen}, on all \atN{} and on the \fzNewNSets{}
sets added by the rule of \S\ref{sec:frozen}. Reach is the number of configurations above that method or at most 0.5 points below it, as in \S\ref{sec:protocol}. API costs are medians over the \atNCfg{}
configurations of Table~\ref{tab:opt}. On DREsS, the prompt-only scores fit no Score cutoffs, so they are
lower than the Base q.\ column of Table~\ref{tab:opt}.}
\label{tab:answer}
\end{table}

\begin{table}[t]
\centering\small
\setlength{\tabcolsep}{3pt}
\resizebox{\columnwidth}{!}{%
\begin{tabular}{@{}l*{2}{>{\centering\arraybackslash}p{2.9em}}cc@{}}
\toprule
& \multicolumn{2}{c}{Gap closed (\%)} & \multicolumn{2}{c}{Median difference} \\
\cmidrule(lr){2-3}\cmidrule(l){4-5}
\multirow{-2}{*}[0.5ex]{Model} & Cal. & Opt. & vs.\ \jev{} & vs.\ encoder \\
\midrule
\jev{} 1.13 & \clJevGapCalAll & \clJevGapAll & -- & $\clJevEncMedDiff$ \\
Clef (27B) & \clClefGapCalAll & \clClefGapAll & $\clClefMedOptJev$ & $\clClefEncMedDiff$ \\
Clef-flash (9B) & \clFlashGapCalAll & \clFlashGapAll & $\clFlashMedOptJev$ & $\clFlashEncMedDiff$ \\
GPT-6 Luna Decisions & \luRqMedGapCal & \luRqMedGapOpt & $\luRqMedOptJev$ & $\luRqEncMedDiff$ \\
Microsoft-Decision-1 & \msRqMedGapCal & \msRqMedGapOpt & $\msRqMedOptJev$ & $\msRqEncMedDiff$ \\
Qwen3.8-27B & \hrBaseGapCal & \hrBaseGapOpt & -- & -- \\
\bottomrule
\end{tabular}}
\caption{Other models answering the optimized questions that optimizer seed 0 found for \jev{}, on the \clClefNsets{} configurations of
\S\ref{sec:opt}, with the logistic regression and the choice of questions refitted for each model. Gap
closed is measured from each model's own zero-shot score (Appendix~\ref{app:rq1}). The median differences,
in points, compare the model with these questions against \jev{} with the same questions and against the
best encoder trained on the same labels (for \jev{}, its 3-seed mean). Negative means below. All columns
are medians over the configurations, and Table~\ref{tab:vendors} gives the significance tests. With these
seed 0 questions, \jev{} closes \clJevGapAll\% of the gap, against \atOptGap\% for its 3-seed mean in
Table~\ref{tab:answer}.}
\label{tab:models}
\end{table}

\paragraph{Encoders.}
The best of three trained baselines on validation is a fine-tuned RoBERTa-large or DeBERTa-v3-large
on \bbNBestFt{} of \bbNSets{} configurations. It is behind the optimized questions on
\bbPipelineSigSets, ahead on \bbBaselineSigSets{} and not significantly different on the other
\bbNUnclear{} (Holm). Pooled over the configurations other than TB-Dense, where fine-tuning was unstable,
the two are equivalent within $\pm$2 points (TOST, Appendix~\ref{app:budget}).

\paragraph{LoRA on an open LLM.}
We fine-tuned Qwen3.8-27B, the LLM that Clef is post-trained from, with LoRA on the same \optTrainN{}
training labels (LoRA-Qwen, seed 0). Because LoRA has one run, we compare it with \jev{}'s optimized
questions from optimizer seed 0. Pooled over configurations, LoRA-Qwen is more accurate by \unsign{\lrPoolJev} points (95\% interval \unsign{\lrPoolJevLo} to \unsign{\lrPoolJevHi}).
After Holm correction, it is ahead on \lrAheadJev{} of \lrNSets{} configurations and behind on none
(Appendix~\ref{app:lora}). It is also ahead of the best same-label encoder on \lrAheadEnc{} and behind on
\lrBehindEnc, and reaches the paper's method on \lrNReach{} configurations (Table~\ref{tab:answer}).

\paragraph{Accuracy and effort.}
A Typed Decision Model can classify from a prompt alone, at a median cost of \bodyCostPerKMedian{} per
1,000 test items. With labels, one search over its questions cost a median \$\frCostFqMedian{} and took
\efSearchMinutesMed{} minutes, not counting the starting questions or the development of the search.
Fine-tuning needs a user-managed GPU and training code from the start. LoRA-Qwen took a median of
\atGpuLora{} GPU-hours of training per configuration on an H100, with peak memory of \lrPeakGiB\,GiB.

\subsection{Other Models}\label{sec:opt-clef}

To check that the optimized questions transfer to other models, Clef, Clef-flash, Luna and
Microsoft-Decision-1 answered the questions that optimizer seed 0 found for \jev{}, with the logistic regression and the
choice of questions refitted for each model (Table~\ref{tab:models}, Appendix~\ref{app:vendors}). The
questions improve every model, by \unsign{\luRqPoolOpt} to \unsign{\clClefPoolOpt} points pooled
(Table~\ref{tab:vendors}).
Qwen3.8-27B, scored from its option log-probabilities, is a median \absnum{\hrBaseMedOptClef} points below
Clef (Appendix~\ref{app:h100}). The optimized questions need probabilities for all options, and
Qwen3.8-27B's log-probabilities provide them.

%% file: generated/body_opt.tex
\begin{table*}[t]
\centering\small
\setlength{\tabcolsep}{3.6pt}
\resizebox{\textwidth}{!}{%
\begin{tabular}{@{}lccccccc>{\columncolor{gray!10}}c>{\columncolor{gray!10}}cccccc@{}}
\toprule
& & \multicolumn{2}{c}{Zero-shot} & \multicolumn{4}{c}{Prompt search (training data)} & \multicolumn{2}{c}{Optimized q.\ (ours)} & \multicolumn{2}{c}{Other model} & \multicolumn{1}{c}{Same labels} & \multicolumn{2}{c}{Paper} \\
\cmidrule(lr){3-4}\cmidrule(lr){5-8}\cmidrule(lr){9-10}\cmidrule(lr){11-12}\cmidrule(lr){13-13}\cmidrule(l){14-15}
& & & & \multicolumn{1}{c}{GEPA} & \multicolumn{1}{c}{GEPA} & \multicolumn{1}{c}{GEPA} & \multicolumn{1}{c}{Jev-} & & & \multicolumn{2}{c}{same q.} & & & \\
\multirow{-3}{*}[0.5ex]{Configuration} & \multicolumn{1}{c}{\multirow{-3}{*}[0.5ex]{$N$}} & \multicolumn{1}{c}{\multirow{-2}{*}{Base q.}} & \multicolumn{1}{c}{\multirow{-2}{*}{Rephr. q.}} & \multicolumn{1}{c}{text} & \multicolumn{1}{c}{prog.} & \multicolumn{1}{c}{+d/c} & \multicolumn{1}{c}{adapted} & \multicolumn{1}{>{\columncolor{gray!10}}c}{\multirow{-2}{*}{mean}} & \multicolumn{1}{>{\columncolor{gray!10}}c}{\multirow{-2}{*}{$\Delta$}} & \multicolumn{1}{c}{Clef} & \multicolumn{1}{c}{Flash} & \multicolumn{1}{c}{\multirow{-2}{*}{Encoder}} & \multicolumn{1}{c}{\multirow{-2}{*}{Method}} & \multicolumn{1}{c}{\multirow{-2}{*}{LLM}} \\
\midrule
ConTempo TB-Dense & 1,427 & 54.4 & 57.4$^\circ$ & 58.7$^\circ$ & 60.9 & 60.2 & \underline{63.0} & \textbf{61.1}{\tiny\,$\pm$1.4} & \tiny{+6.7} & 56.6$^\circ$ & 51.7 & 30.8$^\ast${\tiny\,$\pm$9.5} & 68.6 & 67.6 \\
EZ-STANCE A & 7,666 & 60.7 & 68.6 & 69.5 & 76.0 & 74.9 & 76.3 & \textbf{\underline{77.0}}{\tiny\,$\pm$1.1} & \tiny{+16.4} & 71.8 & 66.1 & 69.8$^\ast${\tiny\,$\pm$2.0} & 81.2 & 49.9 \\
SMARTER Implicit Hate & 900 & 41.9 & 53.2 & 58.4 & \underline{61.3} & 61.2 & 57.1 & \textbf{61.2}{\tiny\,$\pm$3.2} & \tiny{+19.3} & 62.9 & 58.9 & 56.2{\tiny\,$\pm$0.8} & 60.0 & 58.0 \\
Re3-Sci edit intent & 2,312 & 66.3 & 68.8 & 70.5 & 67.7$^\circ$ & 73.0 & \underline{76.2} & \textbf{73.9}{\tiny\,$\pm$1.0} & \tiny{+7.6} & 74.9 & 68.1 & 75.1{\tiny\,$\pm$0.3} & 84.3 & 37.0 \\
INLI & 4,000 & 77.0 & 77.3$^\circ$ & 79.2 & 79.3 & 81.8 & 80.0 & \textbf{\underline{81.8}}{\tiny\,$\pm$0.7} & \tiny{+4.8} & 76.5 & 74.1 & 85.9$^\ast${\tiny\,$\pm$1.4} & 92.4 & 77.0 \\
DREsS content & 401 & 29.8$^\dagger$ & 22.9$^\circ$ & 37.5 & 29.1$^\circ$ & 59.5 & 37.7 & \textbf{\underline{63.8}}{\tiny\,$\pm$2.1} & \tiny{+34.0} & 65.6 & 47.8 & 69.7{\tiny\,$\pm$1.4} & 63.1 & 31.0 \\
DREsS organization & 401 & 42.1$^\dagger$ & 41.3$^\circ$ & 49.1$^\circ$ & 51.0 & 61.8 & 50.3 & \textbf{\underline{62.0}}{\tiny\,$\pm$2.2} & \tiny{+19.9} & 66.7 & 58.9 & 65.9{\tiny\,$\pm$1.2} & 77.1 & 32.2 \\
DREsS language & 401 & 35.6$^\dagger$ & 7.6 & 44.1$^\circ$ & 27.7 & 53.7 & 35.8$^\circ$ & \textbf{\underline{57.0}}{\tiny\,$\pm$2.9} & \tiny{+21.4} & 62.1 & 33.8 & 64.0{\tiny\,$\pm$1.3} & 58.9 & 23.1 \\
\bottomrule
\end{tabular}}
\caption{Test scores ($\times$100) on the \optNTasks{} configurations chosen for large convention-driven gaps. The optimized questions (bold) are a mean $\pm$ sd over 3 seeds, $\Delta$ is their gain over the base question, and underline marks the best \jev{} score in the row. Jev-adapted is GEPA text search changed for \jev{}, which fits class biases or Score cutoffs but no logistic regression before scoring. $^\dagger$ marks Score cutoffs fit on training data, $^\circ$ a 95\% CI of the difference to the base question that includes 0, and $^\ast$ Holm $p<0.05$ against the optimized questions.}
\label{tab:opt}
\end{table*}

%% file: generated/answer_table.tex
\begin{tabular}{@{}lccccc@{}}
\toprule
& & & \multicolumn{2}{c}{Gap closed (\%)} & \\
\cmidrule(lr){4-5}
\multirow{-2}{*}[0.5ex]{Classifier} & \multicolumn{1}{c}{\multirow{-2}{*}[0.5ex]{Labels}} & \multicolumn{1}{c}{\multirow{-2}{*}[0.5ex]{Build cost}} & \multicolumn{1}{c}{all 13} & \multicolumn{1}{c}{5 by rule} & \multicolumn{1}{c}{\multirow{-2}{*}[0.5ex]{Reach}} \\
\midrule
\jev{}, prompt & none & none & 0 & 0 & 0 \\
\quad + calibration & 400 & \$0.04 & 23 & 23 & 0 \\
\quad + opt.\ q. & 500+val. & \$1.66 & 72 & 72 & 3 \\
Encoder & 500+val. & 0.37 GPU-h & 59 & 58 & 3 \\
LoRA-Qwen & 500+val. & 1.03 GPU-h & 97 & 100 & 6 \\
Paper's method & full & -- & 100 & 100 & 13 \\
\bottomrule
\end{tabular}

%% file: sections/when.tex
\section{What to Use When}
\label{sec:when}

\begin{table}[t]
\centering\small
\setlength{\tabcolsep}{4pt}
\begin{tabular}{@{}>{\raggedright\arraybackslash}p{0.49\columnwidth}>{\raggedright\arraybackslash}p{0.46\columnwidth}@{}}
\toprule
Situation & Result in our sample \\
\midrule
\multicolumn{2}{@{}l@{}}{\textit{No labels, a prompt only (\bodyCostPerKMedian{} per 1,000 items)}} \\
Binary judgment & replaced on \bodyGeoBinaryMatch{} of \bodyGeoBinaryN{} test sets \\
Method not trained on the paper's labels & replaced on \number\numexpr\nReplaces-\ovRqOneMatchTrained\relax{} of \number\numexpr\nDatasets-\ovRqOneNTrained\relax \\
Method trained on the paper's labels & replaced on \ovRqOneMatchTrained{} of \ovRqOneNTrained \\
Over 300 possible labels & replaced on \bodyGeoLargeMatch{} of \bodyGeoLargeN \\
\midrule
\multicolumn{2}{@{}l@{}}{\textit{\optTrainN{} training labels plus validation}} \\
Calibration and optimized questions, no GPU & closes \atOptGap\% of the gap (\$\frCostFqMedian, \efSearchMinutesMed{} min) \\
Fine-tuned encoder & closes \atEncGap\% (\atGpuEnc{} GPU-h) \\
LoRA on Qwen3.8-27B & closes \atLoraGap\% (\atGpuLora{} GPU-h) \\
\bottomrule
\end{tabular}
\caption{When a prompt was enough and what \optTrainN{} training labels added. \emph{Replaced} counts the
test sets on which prompt-only \jev{} matches or beats the paper's method (\S\ref{sec:protocol}), and a test
set can fall under more than one of the first four rows. The last three rows give the median share of the
gap to the paper's method closed on the \atN{} configurations and the median build cost
(Table~\ref{tab:answer}).}
\label{tab:when}
\end{table}

In our sample, fine-tuning stayed ahead in two cases (Table~\ref{tab:when}). The first is large opaque label spaces. On all
\bodyGeoLargeN{} test sets with over 300 possible labels, zero-shot \jev{} was below both
(\S\ref{sec:protocol}). Only MADE has an LLM baseline, and \jev{} was below it too. The second is accuracy with the same labels, where LoRA-Qwen reached the paper's method on
\lrNReachWord{} of the \atN{} configurations, \jev{}'s optimized questions on \lrJevNReachWord{} and the same-label
encoder on \bbNBestReachPaper{} (Table~\ref{tab:answer}).

We could not tell from a task's description whether its labels are plain. Zero-shot \jev{} matched the paper's
method on some test sets whose label definitions cite annotation guidelines and was below it on many whose
definitions do not.

%% file: sections/related_work.tex
\section{Related Work}
\label{sec:related}

\paragraph{Prompted LLMs against fine-tuned models.}
Analyses of prompted LLMs on classification find that they can rival crowd workers
\citep{gilardi2023chatgpt} but trail fine-tuned models on most benchmarks
\citep{qin-etal-2023-chatgpt,kocon2023chatgpt,ziems2024css,lehman2023clinical,bucher2024finetuned}. With few labels, SetFit \citep{tunstall2022setfit}
and logistic regression on small-LLM embeddings \citep{buckmann2024logistic} are strong, and prompt
optimization and fine-tuning work best together \citep{soylu-etal-2024-fine}. The conclusions of such
analyses depend on the benchmarks they choose \citep{edwards-camacho-collados-2024-language}.

\paragraph{Typed Decision Models.}
\citet{rao2026jev} find that \jev{} matches LLM rubric judges \citep{zheng2023judging} at a far lower
cost and that unstated rating conventions may explain the errors both make. We extend this
explanation to published predictors. Concurrent work compares \jev{} with option probabilities from open
models \citep{deusser2026jev}, with LLMs and other Typed Decision Models
\citep{ibrahim2026decision,duan2026general}, and with trained classifiers. Trained classifiers lead on
intent detection when they see the task's labels \citep{rafe2026benchmarking} and stay competitive as
guardrails \citep{geada2026redhat}, a setting where prompts did not transfer to \jev{} from Laya, an open-weight Typed Decision Model
\citep{laya2026}.
Other work studies scale compression, option probabilities from any LLM, fallbacks and selective
thresholds \citep{gao2026choices,zhang2026anyjev,li2026llmasjev,ibrahim2026decision,li2026jevjudge,geifman2017selective,liu2026systemone,li2026fast,lu2026rejection}.
Unlike these studies, we swap the predictor inside published evaluations.

\paragraph{Reproducing published comparisons.}
Many NLP results cannot be reproduced from what the papers release \citep{belz-etal-2021-systematic},
and \citet{belz-etal-2023-missing} find that most of the human evaluations they assessed lacked the information or conditions needed to repeat them. Comparisons also depend on tuning budgets
\citep{dodge-etal-2019-show} and contamination \citep{sainz-etal-2023-nlp}. We therefore keep each
paper's test set and metric and report every number we could not reproduce (Table~\ref{tab:deviations}).

\paragraph{Techniques we build on.}
Prompt optimizers \citep{zhou2023ape,pryzant-etal-2023-automatic,khattab2024dspy,agrawal2025gepa}
score a prompt by the answer it elicits. Concept bottleneck
models predict the label from concepts \citep{koh2020concept}, which in text can be LLM answers to
expert-written questions \citep{mcinerney-etal-2023-chill} or selected yes/no questions \citep{benara2024qaemb}. The Text Bottleneck Model
\citep{ludan2023tbm} lets an LLM propose concepts from misclassified training examples and fits a
linear layer on their scores. Boosting and feature-selection methods also combine answers to many prompts
\citep{hou2023promptboosting,pitis2023boosted,zhang2024prefer,morris-etal-2023-tree,zhong2024explaining}, and a concurrent SemEval-2026 system
fits 488 coefficients over \jev{}'s answers \citep{zhang2026decide}.
The search in \jevopt{} follows GEPA but asks many typed questions in one request and scores them after
fitting (\S\ref{sec:fqs}).

%% file: sections/conclusion.tex
\section{Conclusion}
\label{sec:conclusion}

In our sample, a prompt alone was enough to replace the paper's predictor on about a quarter of the test
sets, mostly single judgments with plain labels. Where it fell short, the label name most often left out
an annotation convention, and the other Typed Decision Models fell short on the same test sets. With
\optTrainN{} training labels and a validation split, calibration and optimized questions closed most of
the gap without training the model, about as well as a fine-tuned encoder. LoRA fine-tuning of an open
LLM on the same labels closed almost all of it.

Typed Decision Models are still less accurate than fine-tuning, but users can build a classifier with
them from a prompt alone. A prompt is cheap enough to try first, because we could not tell in
advance which labels are plain. Where it falls short, a short search over the questions is the next step.
Fine-tuning stays ahead when the label space is large or the highest accuracy matters. The optimized
questions also improved the other Typed Decision Models.

%% file: sections/limitations.tex
\section*{Limitations}

We chose the \specCore{} papers by hand and included more likely hard cases than a random sample
would, so the share of each outcome holds for this sample only. All test data are English. We take
each paper's published score as fixed, so our intervals do not include how much the paper's score would vary
with other training runs or model choices. We had to rebuild or use partial releases of some test sets (SMARTER, Navigating
the Unknown, DREsS, FENICE and ConTempo). Except for LoRA on SciPrompt S2ORC, a method trained on our
\optTrainN{} labels beats the paper's method only on test sets that we reconstructed or whose inputs we
rebuilt (Table~\ref{tab:deviations}).

We developed the search over several rounds on the same eight configurations, including their test
splits, and the final optimizer settings came out of this work. For this reason we recorded every test score we computed and will release them all. Three of the \atN{} configurations with labels are DREsS rubrics
that score the same essays. Without them, the medians of the gap closed are \atCalGapNoDress\% for calibration, \atOptGapNoDress\% for optimized questions, \atEncGapNoDress\% for the encoder and \atLoraGapNoDress\% for LoRA. We tuned the
trained baselines only lightly. Fine-tuning was unstable on TB-Dense, and LoRA used one seed and
often picked its last epoch, so it might do better with more training. An LLM drafted the cause table
and the convention spans and an author checked them, so they are not an independent annotation.

The other vendors' models answered questions that were optimized for \jev{}. Clef's developers report
results on BANKING77 and CLINC150, which are the data behind several sets where Clef does much better
than \jev{} \citep{cloudflare2026clef,chen2026clef}, and we did not test any model for contamination.
\jev{}, Luna and Microsoft-Decision-1 are commercial APIs that their vendors can change, and the same
request does not always return the same label. When we sent identical requests to \jev{} twice, \cfRerunFlips{} of \cfRerunN{} labels changed, almost all of them at low confidence. The results can therefore be reproduced from the responses we recorded, which we will release\uponacceptance, but not exactly from new calls. We did not measure inference latency.

\section*{Ethics Statement}

We use only publicly released datasets, and we took the LDC-derived NYT-FB, TimeBank and ATIS texts
from public research releases (RECON, TempEval-3 and the ATIS copy that CICC uses). Where licence or access terms restrict
redistribution (LDC-derived text, CORD-19 with its mixed source licences, the consent-restricted DREsS essays and the
access-restricted MADE release), we will release row identifiers, hashes and rebuild scripts instead of text. The released material will be intended for research reproduction only. We did not scan the data
for personal information and rely on the sources' anonymization. No annotators or crowdworkers were
recruited. Several datasets contain hate speech, unsafe advice or harmful requests (SMARTER, LifeTox,
XSTest).

Test items went through OpenRouter to \jev{} and Microsoft-Decision-1 on zero-retention endpoints. Requests to
GPT-6 Luna Decisions also went through OpenRouter but were served by OpenAI, which keeps abuse-monitoring logs for up to 30 days. The open-weight
models and baselines ran on our own GPUs. Seven of the source papers already ran OpenAI models on the same data, as did earlier work on NYT, ATIS and TREC-COVID
\citep{wadhwa-etal-2023-revisiting,pan2023preliminary,zhu-etal-2024-zero,sun-etal-2023-chatgpt}.

A match on a paper's test set does not show that a decision model is safe to deploy in that paper's
application. Several tasks come from high-stakes settings (adverse medical-device events, legal
judgments, hate-speech moderation), and on these our results show large shortfalls. Building a classifier
from a prompt also makes profiling and large-scale moderation cheaper, and it sends the classified
text to a third-party API.

None of the authors is affiliated with TypeSafe, Cloudflare, OpenAI or Microsoft. We paid list prices
for the API models, whose terms permit research use, and we will release the raw responses under each
provider's terms. Clef, Clef-flash and Qwen3.8-27B are open-weight models
under Apache-2.0 that we ran at pinned revisions. The training data of all these models are not public.

\section*{Generative AI Use Disclosure}

Generative AI tools assisted at different stages of preparing this paper. The authors have checked
their outputs and take complete responsibility for any resulting inaccuracies.

%% file: sections/appendix.tex
\input{generated/compute_budget}

\section*{Appendix overview}
The appendices follow the order of the main text. They cover the sample (\ref{app:selection}), the
statistics (\ref{app:rq1}), the cause of each shortfall (\ref{app:causes}), the search
(\ref{app:opt}), fine-tuning on the same labels (\ref{app:budget}), the other Typed Decision Models
(\ref{app:vendors}), the convention ablation (\ref{app:conventions}), cost, compute, data and licences
(\ref{app:repro}), an optimized question set (\ref{app:opt-programs}) and the data of each paper
(\ref{app:catalogue}).

\section{Sample and test sets}\label{app:selection}
Candidates came from the ACL Anthology export of the main-conference and Findings volumes of ACL, EMNLP,
NAACL and EACL from \specFrom{} to the \specCutoff{} cutoff (\specInScope{} records).
The eligibility rule and the choice of the \specCore{} papers are described in \S\ref{sec:protocol}. We flagged
non-English rows with language identification \citep{lui-baldwin-2012-langid} and checked them by hand
before removing them. Table~\ref{tab:full} lists every scored test set, and Table~\ref{tab:deviations} gives every
deviation from the papers' protocols. Appendix~\ref{app:catalogue} describes the data of each paper.

\input{generated/full_table}

\begin{table*}[!tp]
\centering\footnotesize
\setlength{\tabcolsep}{4pt}
\begin{tabular}{@{}lp{2.2cm}p{9.3cm}@{}}
\toprule
Paper & \multicolumn{1}{c}{Test set} & \multicolumn{1}{c}{Deviations from the paper's protocol} \\
\midrule
LifeTox & complete & none \\
PiNose & complete & accuracy uses a fixed threshold, where the paper tunes it (AUC, the primary metric, is unaffected) \\
FENICE & partial release & different test rows from the paper. The GPT-3.5 claims behind its best number are not released, so we compare with FENICE on T5 claims. Stories are rebuilt from public editions \\
FactPICO & complete & expert span ratings are not released, so only the per-summary score is compared \\
CICC & complete & the published BANKING77 result cannot be reproduced at the printed coverage level, so we do not compare on it \\
Re3-Sci edit intent & complete & additions and deletions get a three-label question, and scoring still uses all five labels \\
EZ-STANCE & English subset & the paper's numbers include the non-English rows \\
ConTempo & complete, inputs rebuilt & the paper's CleanMATRES version of MATRES cannot be scored, and our construction of the inputs replaces the paper's \\
MISMATCHED & complete & none \\
INLI & complete & none \\
Entity or Relation Embeddings & English subset & 13 non-English Wiki-WD sentences are removed, and the options carry only the relation names \\
PRP-Graph & English subset & 46 non-English documents are removed from the candidate lists. The paper interpolates with BM25, with the weight chosen on test, so we compare with its uninterpolated number \\
Scalar Constructs & complete & the pairwise set is compared with the paper's DeBERTa-v3-large reward
model \citep{licht-etal-2025-measuring} \\
DREsS & reconstructed & our seeded split, without empty essays. The paper's numbers come from the authors' own split \\
MedReadMe & complete & none \\
Explanation ranking & reconstructed, not comparable & all candidates except the human one are fixed templates per label (a tier-order oracle reaches nDCG@5 0.994), and the paper's Table~20 does not fit its stated construction \\
MADE & complete & none \\
TTP mapping & complete & only the matcher is replaced. Expert test labels came from reviewing the paper's own model's top 20 \\
LegalSeg & English subset & 6 sentences in romanised Hindi are removed. The headline Hier-BiLSTM-CRF number cannot be reproduced from the released predictions, so ToInLegalBERT, which reproduces exactly, is the reference \\
SciPrompt & English subset & 46 non-English abstracts are removed, and the paper's numbers include them. The reference is the best trained setting, and the Emerging-NLP numbers are read from the paper's bar chart (Fig.~3) \\
Navigating the Unknown & reconstructed & the exact utterances and ID/OOD partitions are not released, and the paper gives no formula for its ALL F1, so we use the standard macro-F1 \\
SMARTER & reconstructed & sources, label spaces, ratios and seeds follow the authors' code, but the unreleased rows cannot be matched. Seven non-English HateXplain posts are removed \\
BIORC800 & complete & none \\
XSTest & complete & the reference numbers are our computation from the released labels \\
\bottomrule
\end{tabular}
\caption{The test set we score for each paper and all deviations from the paper's protocol. The questions
are always zero-shot, even where the paper's LLM baselines are few-shot.}
\label{tab:deviations}
\end{table*}

\section{Statistics}\label{app:rq1}
Scores are given on each paper's primary metric, $\times$100. The papers' published numbers are fixed
references without an interval, so the zero-shot tests resample \jev{}'s test items (\ovRqOneB{} bootstrap
replicates). On the \ovRqOneNClustered{} test sets where items share a source unit (documents, premises,
tweets, queries or writing prompts), the resampling is clustered by that unit. When a published number
was computed on other items, we assign it \jev{}'s sampling variance (\ovRqOneNWidened{} test sets).
The verdicts (significantly above or below, parity, inconclusive) are defined in \S\ref{sec:protocol}. The outcome groups
(matches or beats, between, below both) use the 0.5-point rule
and the LLM baseline defined there. If the LLM baseline is instead
the best prompted LLM that the paper reports, including few-shot rows, the first group has
\ovRqOneLlmBestReplaces{} test sets, the second \ovRqOneLlmBestBetween{} and the third \ovRqOneLlmBestBelow. A method counts
as \emph{trained} when it was trained on the paper's labelled data (\ovRqOneNTrained{} test sets).
\jev{} matches or beats \ovRqOneMatchTrained{} of them: BIORC800 and SciPrompt SDPRA, whose methods are
fine-tuned on the training data, and SciPrompt arXiv and Emerging-NLP, whose methods use 50-shot and
5-shot prompt tuning.

The experiments with labels compare systems on the same test items. We use paired cluster bootstraps in the same units,
resample the optimizer seeds of both systems where there are seeds, and apply Holm correction over the
configurations. TB-Dense and MATRES have fewer than 30 test documents, so their tests use a
leave-one-document-out jackknife-t interval. We test the equivalence of the optimized questions and the best
same-label baseline with two one-sided tests (TOST, $\alpha=0.05$) on the pooled difference, with a
margin of $\pm$2 points. The main text measures each step by the gap closed,
$(\text{score}-\text{prompt-only})/(\text{paper's method}-\text{prompt-only})$, with \jev{}'s
prompt-only test score as the reference. For the other models, Table~\ref{tab:models} instead measures
the gap from each model's own zero-shot score, which shows how much the questions help that model. For the medians of the gap closed we report a 95\%
percentile interval from 10{,}000 resamples of configurations, drawing the three DREsS rubrics together. For the
five rule-selected sets we report the range instead.

\section{Causes of the zero-shot shortfalls}\label{app:causes}
Table~\ref{tab:causes} gives a primary cause for each of the \csN{} test sets on which \jev{} does not
match or beat the paper's method. The rule was fixed before any row was assigned. Each set gets one
of four primary causes, (a) to (d). Cause (a), \emph{unstated annotation conventions}, is assigned when an
optimized question set states a convention and recovers accuracy, when the meaning of the labels depends on
the dataset's guidelines, or when the model systematically confuses labels that the guidelines separate
only by convention. Cause (b) is a \emph{very large or opaque label space}, with 100 or more options or codes
whose names do not carry their meaning. The remaining causes are (c) \emph{probabilities too coarse} for
the paper's downstream method and (d) \emph{other or unclear}. An LLM drafted every row from the cited evidence,
and an author checked it and kept the draft's primary cause on \csAgree{} of \csN{} sets (Cohen's
$\kappa=\csKappa$ between the draft and the checked version, 95\% bootstrap interval \csKappaLo--\csKappaHi).

\input{generated/app_causes_compact}

\section{The search over questions}\label{app:opt}\label{app:opt-v3}\label{app:opt-ir}\label{app:opt-data}\label{app:opt-seeds}\label{app:opt-others}\label{app:frozen}
\jevopt{} represents the questions of a request as a typed program of \textsc{Choice}, \textsc{Score} and
\textsc{Noul} nodes. The search (Figure~\ref{fig:fqs}) proposes question variants with an
optimizer LLM, which may change a node's type, and scores every candidate after fitting calibration or
a logistic regression on its answers to the \optFitN{} fitting examples, using 5-fold cross-validation to
set the regularisation. The optimizer LLM sees the worst out-of-fold examples among them. The search
never uses the test split, and the chosen candidate is tested on it once.
Jev-adapted GEPA in Table~\ref{tab:opt} is a GEPA text search \citep{agrawal2025gepa} with proposal and
acceptance steps adapted to \jev{}, and its tested candidate is the best fitted one on validation. The
base question was queried again for that table. On the five configurations that
Table~\ref{tab:full} also reports, its scores differ between the two tables by at most 0.4 points. For every seed, the 95\% interval of the gain over
the base question, bootstrapped as in Appendix~\ref{app:rq1}, is above 0 on every configuration. On TB-Dense, which has only 9
test documents, a leave-one-document-out jackknife interval includes 0 for one of the three seeds.
Table~\ref{tab:opt-settings} lists the settings, Table~\ref{tab:ablation} shows the validation ablation
and Table~\ref{tab:curve} the fitting curve. In a paired comparison on test, the optimized questions are
significantly better than GEPA with demonstrations and calibration on EZ-STANCE A and not significantly
different from it on the other seven configurations (Table~\ref{tab:rq3-paired}).

With the final optimizer settings, the search uses the same code and optimizer seed 0. We ran it on every
shortfall test set that it supports, that is, every set where one question answers one state, the
gold label is one of at most 40 labels, the code computes the metric and a training set is released
(Table~\ref{tab:frozen}). A separate protocol reran it with optimizer seeds 1 and 2 on the \fzNewNSets{}
rule-selected sets, each chosen on validation and tested once (Table~\ref{tab:frozen-seeds}).

\input{generated/app_opt_settings}
\input{sections/ablation_table}
\input{sections/app_label_curve}

\begin{table*}[!tp]
\centering\footnotesize
\resizebox{\textwidth}{!}{\input{generated/offline_rq3_paired}}
\caption{Paired test differences, in points, between the optimized questions (mean of three seeds) and GEPA
with demonstrations and calibration (one seed).}
\label{tab:rq3-paired}
\end{table*}

\begin{table*}[!tp]
\centering
\resizebox{\textwidth}{!}{\input{generated/app_frozen}}
\caption{The upper block shows test scores ($\times$100) and costs of the search with its final optimizer settings on every
eligible shortfall set, and the lower block the seed 0 runs on the configurations of Table~\ref{tab:opt}. On
DREsS, Zero fits no Score cutoffs, unlike in Table~\ref{tab:opt}.}
\label{tab:frozen}
\end{table*}

\begin{table}[!tp]
\centering\footnotesize
\setlength{\tabcolsep}{2pt}
\resizebox{\ifdim\width>\columnwidth\columnwidth\else\width\fi}{!}{%
\begin{tabular}{@{}lcccccc@{}}
\toprule
Test set & \multicolumn{1}{c}{Seed 0} & \multicolumn{1}{c}{1} & \multicolumn{1}{c}{2} & \multicolumn{1}{c}{Mean} & \multicolumn{1}{c}{Paper} & \multicolumn{1}{c}{Gap \%} \\
\midrule
ConTempo MATRES & \fzsMatresSeedZero & \fzsMatresSeedOne & \fzsMatresSeedTwo & \fzsMatresMean & \fzMatresPaper & \fzsMatresGapMean \\
LifeTox & \fzsLifetoxSeedZero & \fzsLifetoxSeedOne & \fzsLifetoxSeedTwo & \fzsLifetoxMean & \fzLifetoxPaper & \fzsLifetoxGapMean \\
SciPrompt S2ORC & \fzsStorcSeedZero & \fzsStorcSeedOne & \fzsStorcSeedTwo & \fzsStorcMean & \fzStorcPaper & \fzsStorcGapMean \\
SMARTER HateXplain & \fzsHatexplainSeedZero & \fzsHatexplainSeedOne & \fzsHatexplainSeedTwo & \fzsHatexplainMean & \fzHatexplainPaper & \fzsHatexplainGapMean \\
SMARTER Latent Hate & \fzsLatentSeedZero & \fzsLatentSeedOne & \fzsLatentSeedTwo & \fzsLatentMean & \fzLatentPaper & \fzsLatentGapMean \\
\bottomrule
\end{tabular}}
\caption{Test scores ($\times$100) of the search with its final optimizer settings and three optimizer seeds. The median gap
closed by the seed mean is \fzsNewMedGapSetMean\%.}
\label{tab:frozen-seeds}
\end{table}

\section{Fine-tuning with the same labels}\label{app:budget}\label{app:lora}
Every baseline trains on the \optTrainN{} training examples of the optimizer runs and makes every
choice on the same validation split. The encoder baselines are frozen bge embeddings
\citep{xiao2024cpack} with a logistic or ridge regression, SetFit \citep{tunstall2022setfit}, and full
fine-tuning of RoBERTa-large \citep{liu2019roberta} or DeBERTa-v3-large \citep{he2023debertav3}. Full
fine-tuning uses learning rates $1$--$3\times10^{-5}$, three seeds, a batch size of 16 and early stopping
on validation. The best baseline of a configuration is the one of these three with the best validation score.
Excluding TB-Dense, where fine-tuning was unstable (\bbTbdFtMin--\bbTbdFtMax{} across seeds), the pooled
difference between the optimized questions and the best baseline is $\bbPooledNoTbdDiff$ points (90\% interval
$\bbEqNoTbdLo$ to $\bbEqNoTbdHi$), so the two are equivalent within $\pm$2 points. Over all \bbNSets{}
configurations the difference is $\bbPooledAllDiff$ ($\bbEqAllLo$ to $\bbEqAllHi$).

LoRA-Qwen fine-tunes Qwen3.8-27B with LoRA \citep{hu2022lora} (rank 16, $\alpha=32$, dropout 0.05) on the
linear projections of all decoder layers, with the BF16 base model frozen and one H100 per run. It
minimizes the cross-entropy of the gold option's answer token with AdamW, an effective batch size of 16,
learning rates $10^{-4}$ and $2\times10^{-4}$ and three epochs, and the best epoch and learning rate are
chosen on validation. We planned three seeds but ran only seed 0, because a trial run projected that
three seeds would exceed our limit of 80 GPU-hours. Prompts, answer tokens and decoding are the
same as for Qwen3.8-27B with option log-probabilities.
Table~\ref{tab:lora} compares LoRA-Qwen with \jev{}'s optimized questions from optimizer seed 0, with the same LLM
answering those questions and with the best same-label baseline on every configuration.

\begin{table*}[!tp]
\input{generated/app_lora}
\caption{LoRA fine-tuning of Qwen3.8-27B (seed 0) compared with the same model answering \jev{}'s optimized
questions, the best same-label baseline and \jev{} with its optimized questions from optimizer seed 0, on test ($\times$100).
$^\ast$ Holm $p<0.05$.}
\label{tab:lora}
\end{table*}

\section{Other Typed Decision Models}\label{app:vendors}\label{app:clef}\label{app:clef-rq3}\label{app:luna}\label{app:msd1}\label{app:h100}
We ran Clef and Clef-flash \citep{cloudflare2026clef} locally at pinned revisions, and GPT-6 Luna Decisions
\citep{openai2026luna} and Microsoft-Decision-1 \citep{microsoft2026decision1} through OpenRouter at
fixed snapshots. Each model received \jev{}'s zero-shot requests with only the model replaced and
was scored with the same code and metrics (we did not run Wiki-WD on any of these four models
because of its cost). In the experiments with labels, each model answered the optimized questions that optimizer seed 0 found for \jev{}, and
calibration, the logistic regression and the choice among \jev{}'s candidate question sets were refitted on
its own answers.
Qwen3.8-27B, the LLM Clef is post-trained from, answers each question through the softmax over the
log-probabilities of one token per option. Table~\ref{tab:vendors} summarizes both steps.
With \jev{}'s optimized questions, Qwen3.8-27B is a median \absnum{\hrBaseMedOptClef} points below Clef over
the \hrBaseNsets{} configurations (pooled difference $\hrBasePoolOptClef$, 95\% interval $\hrBasePoolOptClefLo$
to $\hrBasePoolOptClefHi$), and Clef is significantly ahead on \hrBaseNnegOptClef{} of them after Holm
correction. Clef, GPT-6 Luna Decisions and Microsoft-Decision-1 are each below the paper's method on
all \clConvN{} convention test sets we ran them on.
Microsoft-Decision-1 does not round its probabilities to two decimals, and its share of single-intent conformal sets on
CICC is \msCiccClinc\% on CLINC150, \msCiccAcid\% on ACID, \msCiccAtis\% on ATIS and \msCiccHwu\% on
HWU64. With \jev{}'s rounded probabilities, every prediction set on CLINC150, ACID and ATIS is the full
label set.

\begin{table*}[!tp]
\centering\small
\setlength{\tabcolsep}{4pt}
\resizebox{\textwidth}{!}{%
\begin{tabular}{@{}lccccccccccc@{}}
\toprule
& \multicolumn{5}{c}{Zero-shot, \ocNSets{} test sets} & \multicolumn{6}{c}{With labels, \clClefNsets{} configurations, \jev{}'s questions} \\
\cmidrule(lr){2-6}\cmidrule(l){7-12}
& Same & Better/within/ & Median & & Higher/ & \multicolumn{3}{c}{Gap closed (\%)} & Gain & \multicolumn{2}{c}{Better/worse} \\
\multirow{-3}{*}[0.5ex]{Model} & group & worse & $-$\jev{} & \multirow{-2}{*}{$\rho$} & lower & Cal. & Opt. & Opt.\ (\jev{} basis) & (points) & vs.\ \jev{} & vs.\ encoder \\
\midrule
Clef (27B) & \ocSameGroup & \clBetter/\clWithin/\clWorse & \clMedian & \ocRho & \ocPairNHigher/\ocPairNLower & \clClefGapCalAll & \clClefGapAll & \atXferClefGap & $\clClefPoolOpt$ & \clClefNholmOptJev/\clClefNnegOptJev & \clClefEncAhead/\clClefEncBehind \\
Clef-flash (9B) & \ocFlashSameGroup & \clFlashBetter/\clFlashWithin/\clFlashWorse & \clFlashMedian & -- & -- & \clFlashGapCalAll & \clFlashGapAll & \atXferFlashGap & $\clFlashPoolOpt$ & \clFlashNholmOptJev/\clFlashNnegOptJev & \clFlashEncAhead/\clFlashEncBehind \\
GPT-6 Luna Decisions & \luSameGroup & \luBetter/\luWithin/\luWorse & \luMedian & \luRho & \luPairNHigher/\luPairNLower & \luRqMedGapCal & \luRqMedGapOpt & \atXferLunaGap & $\luRqPoolOpt$ & \luRqNholmOptJev/\luRqNnegOptJev & \luRqEncAhead/\luRqEncBehind \\
Microsoft-Decision-1 & \msSameGroup & \msBetter/\msWithin/\msWorse & \msMedian & \msRho & \msPairNHigher/\msPairNLower & \msRqMedGapCal & \msRqMedGapOpt & \atXferMsdGap & $\msRqPoolOpt$ & \msRqNholmOptJev/\msRqNnegOptJev & \msRqEncAhead/\msRqEncBehind \\
Qwen3.8-27B & -- & -- & -- & -- & -- & \hrBaseGapCal & \hrBaseGapOpt & \atXferQwenGap & $\hrBasePoolOptZero$ & -- & -- \\
\bottomrule
\end{tabular}}
\caption{Results of the other Typed Decision Models and Qwen3.8-27B without and with labels, compared with
\jev{}. The zero-shot block covers the \ocNSets{} test sets other than Wiki-WD, and the columns with labels
complement Table~\ref{tab:models}. Opt.\ (\jev{} basis) measures the gap from \jev{}'s zero-shot score instead
of the model's own, so that all models start from the same score. Gain is the pooled improvement of the
optimized questions over the model's own zero-shot score. The last two columns count the configurations
where the model is significantly better/worse (Holm $p<0.05$) than \jev{} with the same questions and
than the best encoder trained on the same labels.}
\label{tab:vendors}
\end{table*}

\section{Convention ablation}\label{app:conventions}
The ablation edits only the convention sentences of the \cvPrograms{} optimized question sets that state a
convention (\cvSpans{} spans, marked by an LLM and checked by an author). The question sets are \cvSeedPos{}
from the \optNTasks{} chosen configurations and \cvFrozenPos{} from the \fzNewNSets{} rule-selected sets. We fixed the protocol and every edited
question set before sending any of them to \jev{}. Condition A is the question set as chosen. B deletes
every convention span, and C deletes the same amount of other text. D appends the spans to the base
question, and E adds the calibration step to D. F is the base question with the calibration step, and
Z is the base question zero-shot. Z and F were queried again for this ablation,
as in Table~\ref{tab:opt}, whereas Table~\ref{tab:frozen} reuses the answers of
Table~\ref{tab:full}. Z therefore differs from Table~\ref{tab:frozen} by up to 3.0 points, and F by up to
1.4 points on the seed 0 rows.
Every condition refits its logistic regression or calibration on the same fit examples. The deletion contrast
A$-$B is not significant after Holm correction ($p=\cvDelP$). Because \jev{}
rejects some empty fields, every condition replaces deleted text with the option names. In a sensitivity
analysis that uses the first runs of four question sets, which sent the deleted fields as empty strings, the
pooled A$-$B difference is \cvPreDel{} points (a point estimate).

\begin{table*}[!tp]
\centering
\resizebox{\textwidth}{!}{\input{generated/app_conventions}}
\caption{Effect of adding or deleting the convention sentences of the optimized questions ($\times$100).
Conditions A--F and Z are defined in the text. All rows are on test except the two validation rows, and
$^{*}$ denotes Holm-adjusted $p<0.05$.}
\label{tab:conventions}
\end{table*}

\section{Cost, compute, data and licences}\label{app:repro}\label{app:cost}
\paragraph{Compute budget.} The API calls of all experiments cost at most \$\cbApiTotal.
\jev{} cost at most \$\cbJevTotal{} in total, with
\totalCost{} for the zero-shot evaluation, at most \$\cbJevOpt{} for the search experiments of \S\ref{sec:opt} (cached
answers counted at list price), \$\cbJevFrozen{} for the runs with the final optimizer settings and \cvJevCostAll{} for the
convention ablation. GPT-6 Luna Decisions cost \$\luSpend{}, Microsoft-Decision-1 \$\msSpend{} and the optimizer
LLMs \$\cbOptimizerTotal. The vendors do not disclose the parameter counts of \jev{}, GPT-6 Luna Decisions or
the optimizer LLMs. Microsoft-Decision-1 and Clef-flash are post-trained from Qwen3.5-9B and Clef from
Qwen3.8-27B \citep{qwen2026models}. The encoder baselines are bge-base, RoBERTa-large and DeBERTa-v3-large, with about 110M,
355M and 434M parameters. The GPU runs took \cbGpuHours{} GPU-hours in total, \cbRtxGpuHours{} on RTX 4090s
for the trained baselines and \cbHundredGpuHours{} on H100s. Of the H100 hours, Clef took \cbClefZeroGpuHours{} zero-shot and
\clClefGpuHours{} with the optimized questions, Clef-flash \cbFlashZeroGpuHours{} and \clFlashGpuHours,
Qwen3.8-27B \hrBaseGpuHours, Clef on the rephrased questions \hrNativeGpuHours{} and LoRA
\lrGpuHours. Development and discarded runs for which we kept no record are not counted. The GPU runs
used PyTorch \citep{paszke2019pytorch}, Transformers \citep{wolf-etal-2020-transformers}, PEFT
\citep{mangrulkar2022peft} and scikit-learn \citep{pedregosa2011sklearn}. The Clef, Qwen3.8-27B and
LoRA runs used PyTorch 2.11.0 with CUDA 12.8, Transformers 5.10.2 and PEFT 0.21.2, and the trained
baselines used PyTorch 2.14.0, Transformers 5.17.0, scikit-learn 1.9.1 and SetFit 1.2.0.

\paragraph{Underlying datasets and tools.} Several papers evaluate on datasets built by others.
ConTempo uses TB-Dense \citep{cassidy-etal-2014-annotation}, which annotates TimeBank documents, and MATRES
\citep{ning-etal-2018-multi}, which annotates the TimeBank, AQUAINT and Platinum documents of TempEval-3 \citep{uzzaman-etal-2013-semeval}. Navigating the Unknown uses CLINC150
\citep{larson-etal-2019-evaluation} and BANKING77 \citep{casanueva-etal-2020-efficient}. CICC uses CLINC150,
BANKING77, HWU64 \citep{liu2019hwu64}, MTOD \citep{schuster-etal-2019-cross-lingual}, ATIS
\citep{hemphill-etal-1990-atis} and ACID. SMARTER uses HateXplain \citep{mathew2021hatexplain} and Latent
Hatred \citep{elsherief-etal-2021-latent}, the source of its Latent Hate and Implicit Hate sets, and PRP-Graph uses TREC-COVID \citep{voorhees2020treccovid}
over CORD-19 \citep{wang-etal-2020-cord} in its BEIR version \citep{thakur2021beir}. Explanation
ranking uses e-SNLI \citep{camburu2018esnli}, which extends SNLI
\citep{bowman-etal-2015-large}. Entity or Relation Embeddings builds NYT-FB from the NYT corpus of
\citet{riedel2010modeling}, and SciPrompt uses S2ORC \citep{lo-etal-2020-s2orc}. CICC's corpora have
different licences. BANKING77 and HWU64 are under CC BY 4.0, CLINC150 under CC BY 3.0 and MTOD under
CC BY-SA 4.0. ACID states no licence, and ATIS is derived from an LDC corpus. PRP-Graph's BM25
candidates come from pyserini \citep{lin2021pyserini}. The TACRED family was dropped from the sample
because it is available only from LDC, under a licence we do not hold. The NYT-FB, TimeBank and ATIS
texts also derive from LDC corpora, but public research releases distribute them (see the Ethics
Statement). We
use them for evaluation only and redistribute none of them.

\paragraph{What will be released.} For every run we recorded the raw responses, predictions, scores
and a manifest with the snapshot, prompt hash, request and token counts and cost. Test sets are kept
either as the fixed inputs or, where licences or access terms restrict redistribution (LDC news text,
CORD-19, essays shared only under a consent form, unlicensed releases), as identifiers plus row hashes
with scripts that rebuild them from the original release. \Uponacceptance{} will release the code\releaseat{}
under the Apache License 2.0, together with the prompts, test-set manifests, pre-registered protocols,
per-item predictions and raw responses. Data files derived from published datasets will keep the
licence of their source, files derived from CC BY-NC or CC BY-NC-SA sources will remain non-commercial
and share-alike, and the model responses will be subject to the terms of the respective providers.

\section{An optimized question set}\label{app:opt-programs}
Figure~\ref{fig:program} shows the TB-Dense question before and after optimization.

\input{generated/body_program}

\onecolumn
\section{The data of each paper}\label{app:catalogue}
Table~\ref{tab:catalogue} describes the data of each paper: the task, the output, the scored test sets with
their size and metric, and one example with its gold label. Where the licence does not allow showing a test
item, the example is the paper's own illustration or a training item, paraphrased where needed.

\input{generated/app_catalogue}

%% file: generated/compute_budget.tex
\newcommand{\cbJevOpt}{196}
\newcommand{\cbJevFrozen}{13.29}
\newcommand{\cbJevTotal}{395}
\newcommand{\cbOptimizerTotal}{226}
\newcommand{\cbApiTotal}{893}
\newcommand{\cbClefZeroGpuHours}{97.4}
\newcommand{\cbFlashZeroGpuHours}{40.4}
\newcommand{\cbHundredGpuHours}{188}
\newcommand{\cbRtxGpuHours}{9.5}
\newcommand{\cbGpuHours}{198}

%% file: generated/full_table.tex
\begin{table*}[!tp]
\centering\footnotesize
\setlength{\tabcolsep}{4pt}
\resizebox{\ifdim\width>\textwidth\textwidth\else\width\fi}{!}{%
\begin{tabular}{lccccccc}
\toprule
Dataset & Output & $N$ & Metric & Jev & Paper's method & LLM & $\Delta$ \\
\midrule
\multicolumn{8}{l}{\textbf{\textit{Jev matches or beats the paper's method}}} \\
PiNose TriviaQA~\citep{zhang-etal-2024-transferable} & binary & 1,000 & AUC & \cellcolor{green!15}\textbf{96.4} & 83.9 & \textcolor{gray}{--} & \textcolor{ForestGreen}{+12.5} \\
FENICE~\citep{scire-etal-2024-fenice} & binary & 279 & bal.\ acc. & \cellcolor{green!15}\textbf{77.4} & 65.7 & \textcolor{gray}{--} & \textcolor{ForestGreen}{+11.7} \\
PiNose True-False~\citep{zhang-etal-2024-transferable} & binary & 1,000 & AUC & \cellcolor{green!15}\textbf{96.9} & 86.6 & \textcolor{gray}{--} & \textcolor{ForestGreen}{+10.3} \\
SciPrompt arXiv~\citep{you-etal-2024-sciprompt} & 53-way & 5,295 & acc. & \cellcolor{green!15}\textbf{79.0} & 70.2 & 54.9 & \textcolor{ForestGreen}{+8.9} \\
PiNose NQ~\citep{zhang-etal-2024-transferable} & binary & 1,000 & AUC & \cellcolor{green!15}\textbf{88.0} & 80.4 & \textcolor{gray}{--} & \textcolor{ForestGreen}{+7.6} \\
XSTest (judge)~\citep{rottger-etal-2024-xstest} & 3-way & 2,250 & acc. & \cellcolor{green!15}\textbf{87.0} & 80.4 & 80.4 & \textcolor{ForestGreen}{+6.6} \\
FactPICO~\citep{joseph-etal-2024-factpico} & ordinal & 345 & Kendall $\tau_b$ & \cellcolor{green!15}\textbf{53.4} & 47.4 & 47.5 & \textcolor{ForestGreen}{+6.0} \\
SciPrompt Emerging-NLP~\citep{you-etal-2024-sciprompt} & 21-way & 420 & acc. & \cellcolor{green!15}\textbf{95.0} & 89.3 & 8.6 & \textcolor{ForestGreen}{+5.7} \\
PiNose WebQ~\citep{zhang-etal-2024-transferable} & binary & 1,000 & AUC & \cellcolor{green!15}\textbf{88.4} & 83.3 & \textcolor{gray}{--} & \textcolor{ForestGreen}{+5.1} \\
PRP-Graph TREC-COVID~\citep{luo-etal-2024-prp} & ranking & 50 & nDCG@10 & \cellcolor{green!15}\textbf{82.1}$^\dagger$ & 78.8 & \textcolor{gray}{--} & \textcolor{ForestGreen}{+3.3} \\
SciPrompt SDPRA~\citep{you-etal-2024-sciprompt} & 7-way & 2,789 & acc. & \cellcolor{green!15}\textbf{91.3} & 90.7 & 81.2 & \textcolor{ForestGreen}{+0.6} \\
BIORC800~\citep{lan-etal-2024-multi} & multi-label & 160 & micro-F1 & \cellcolor{green!15}\textbf{90.8}$^\dagger$ & \textbf{90.8} & 47.6 & \textcolor{ForestGreen}{0.0} \\
\midrule
\multicolumn{8}{l}{\textbf{\textit{Jev between the prompted LLM and the paper's method}}} \\
CICC MTOD~\citep{hengst-etal-2024-conformal} & conformal set & 8,665 & Single & 97.3 & \textbf{100.0} & \textcolor{gray}{--} & \textcolor{BrickRed}{$-$2.7} \\
SciPrompt S2ORC~\citep{you-etal-2024-sciprompt} & 19-way & 5,660 & acc. & 52.9 & \textbf{56.7} & 49.6 & \textcolor{BrickRed}{$-$3.7} \\
Scalar Constructs: ad neg.\ (pair)~\citep{licht-etal-2025-measuring} & pairwise & 1,944 & Spearman $\rho$ & 87.1 & \textbf{91.0} & 86.0 & \textcolor{BrickRed}{$-$3.9} \\
MISMATCHED~\citep{shaik-etal-2025-mismatched} & 4-way & 2,400 & macro-F1 & 71.2 & \textbf{78.2} & 55.6 & \textcolor{BrickRed}{$-$7.0} \\
Scalar Constructs: grandst.\ (point)~\citep{licht-etal-2025-measuring} & ordinal & 3,414 & Spearman $\rho$ & 70.0$^\dagger$ & \textbf{78.0} & 60.0 & \textcolor{BrickRed}{$-$8.0} \\
SMARTER HateXplain~\citep{nghiem-etal-2026-smarter} & 3-way & 1,193 & macro-F1 & 55.6 & \textbf{64.0} & 56.0 & \textcolor{BrickRed}{$-$8.4} \\
\begin{tabular}[t]{@{}l@{}}Navigating the Unknown\\\quad CLINC150 (ID+OOD)~\citep{sali-toraman-2025-navigating}\end{tabular} & 76-way & 1,500 & macro-F1 & 86.0 & \textbf{95.7} & 83.7 & \textcolor{BrickRed}{$-$9.7} \\
ConTempo MATRES~\citep{niu-etal-2024-contempo} & 4-way & 837 & micro-F1 & 72.3 & \textbf{83.2} & 56.7 & \textcolor{BrickRed}{$-$10.9} \\
\begin{tabular}[t]{@{}l@{}}Navigating the Unknown\\\quad BANKING77 (ID+OOD)~\citep{sali-toraman-2025-navigating}\end{tabular} & 39-way & 770 & macro-F1 & 71.2 & \textbf{83.2} & 69.4 & \textcolor{BrickRed}{$-$12.0} \\
Scalar Constructs: imm.\ fear (pair)~\citep{licht-etal-2025-measuring} & pairwise & 3,265 & Spearman $\rho$ & 78.7 & \textbf{91.0} & 74.0 & \textcolor{BrickRed}{$-$12.3} \\
INLI~\citep{havaldar-etal-2025-entailed} & 4-way & 4,000 & acc. & 76.8 & \textbf{92.4} & 77.0 & \textcolor{BrickRed}{$-$15.7} \\
Re3-Sci edit intent~\citep{ruan-etal-2024-large} & 5-way & 2,312 & macro-F1 & 66.7 & \textbf{84.3} & 37.0 & \textcolor{BrickRed}{$-$17.6} \\
LifeTox~\citep{kim-etal-2024-lifetox} & binary & 3,000 & macro-F1 & 78.4 & \textbf{96.5} & 74.4 & \textcolor{BrickRed}{$-$18.1} \\
EZ-STANCE B~\citep{zhao-caragea-2024-ez} & 3-way & 45,274 & macro-F1 & 59.5 & \textbf{78.3} & 49.5 & \textcolor{BrickRed}{$-$18.8} \\
Scalar Constructs: grandst.\ (pair)~\citep{licht-etal-2025-measuring} & pairwise & 2,228 & Spearman $\rho$ & 58.6 & \textbf{78.0} & 54.0 & \textcolor{BrickRed}{$-$19.4} \\
EZ-STANCE A~\citep{zhao-caragea-2024-ez} & 3-way & 7,666 & macro-F1 & 60.7 & \textbf{81.2} & 49.9 & \textcolor{BrickRed}{$-$20.5} \\
LegalSeg~\citep{nigam-etal-2025-legalseg} & 7-way seq. & 149,875 & macro-F1 & 52.0 & \textbf{77.0} & 8.0 & \textcolor{BrickRed}{$-$25.0} \\
\midrule
\multicolumn{8}{l}{\textbf{\textit{Jev below both}}} \\
Scalar Constructs: ad neg.\ (point)~\citep{licht-etal-2025-measuring} & ordinal & 935 & Spearman $\rho$ & 89.4$^\dagger$ & \textbf{91.0} & \textbf{91.0} & \textcolor{BrickRed}{$-$1.6} \\
SMARTER Latent Hate~\citep{nghiem-etal-2026-smarter} & 3-way & 1,019 & macro-F1 & 58.0 & \textbf{69.0} & 60.0 & \textcolor{BrickRed}{$-$11.0} \\
MedReadMe~\citep{jiang-xu-2024-medreadme} & ordinal & 1,140 & Pearson $r$ & 71.4 & \textbf{83.3} & 72.1 & \textcolor{BrickRed}{$-$11.9} \\
Scalar Constructs: imm.\ fear (point)~\citep{licht-etal-2025-measuring} & ordinal & 334 & Spearman $\rho$ & 77.4$^\dagger$ & \textbf{91.0} & 81.0 & \textcolor{BrickRed}{$-$13.6} \\
ConTempo TB-Dense~\citep{niu-etal-2024-contempo} & 6-way & 1,427 & micro-F1 & 54.3 & \textbf{68.6} & 67.6 & \textcolor{BrickRed}{$-$14.3} \\
SMARTER Implicit Hate~\citep{nghiem-etal-2026-smarter} & 6-way & 900 & macro-F1 & 41.9 & \textbf{60.0} & 58.0 & \textcolor{BrickRed}{$-$18.1} \\
TTP mapping Expert~\citep{nguyen-etal-2024-noise} & 737 ranking & 157 & P@1 & 54.8 & \textbf{73.1} & \textcolor{gray}{--} & \textcolor{BrickRed}{$-$18.3} \\
MADE~\citep{agarwal-etal-2026-made} & 1,154 multi-label & 10,288 & macro-F1 & 32.9 & \textbf{54.0} & 43.0 & \textcolor{BrickRed}{$-$21.1} \\
TTP mapping Procedures+~\citep{nguyen-etal-2024-noise} & 737 ranking & 2,288 & P@1 & 45.9 & \textbf{75.7} & \textcolor{gray}{--} & \textcolor{BrickRed}{$-$29.8} \\
TTP mapping TRAM~\citep{nguyen-etal-2024-noise} & 737 ranking & 725 & P@1 & 44.0 & \textbf{77.0} & \textcolor{gray}{--} & \textcolor{BrickRed}{$-$33.0} \\
CICC HWU64~\citep{hengst-etal-2024-conformal} & conformal set & 5,143 & Single & 47.9 & \textbf{82.0} & \textcolor{gray}{--} & \textcolor{BrickRed}{$-$34.1} \\
\begin{tabular}[t]{@{}l@{}}Entity or Relation Embeddings\\\quad Wiki-WD~\citep{mtumbuka-schockaert-2024-entity}\end{tabular} & 352-way & 739,977 & micro-F1 & 54.1 & \textbf{89.1} & \textcolor{gray}{--} & \textcolor{BrickRed}{$-$35.0} \\
DREsS~\citep{yoo-etal-2025-dress} & ordinal & 401 & QWK & 20.6 & \textbf{69.1} & 30.4 & \textcolor{BrickRed}{$-$48.5} \\
\begin{tabular}[t]{@{}l@{}}Entity or Relation Embeddings\\\quad NYT-FB~\citep{mtumbuka-schockaert-2024-entity}\end{tabular} & 56-way & 172,444 & P@10 & 19.6 & \textbf{94.9} & \textcolor{gray}{--} & \textcolor{BrickRed}{$-$75.3} \\
CICC ACID~\citep{hengst-etal-2024-conformal} & conformal set & 4,434 & Single & 0.0 & \textbf{87.0} & \textcolor{gray}{--} & \textcolor{BrickRed}{$-$87.0} \\
CICC CLINC150~\citep{hengst-etal-2024-conformal} & conformal set & 3,605 & Single & 0.0 & \textbf{97.0} & \textcolor{gray}{--} & \textcolor{BrickRed}{$-$97.0} \\
CICC ATIS~\citep{hengst-etal-2024-conformal} & conformal set & 1,174 & Single & 0.0 & \textbf{98.0} & \textcolor{gray}{--} & \textcolor{BrickRed}{$-$98.0} \\
\bottomrule
\end{tabular}}
\caption{All \nDatasets{} test sets in the comparison, formatted as in Table~\ref{tab:main}.}
\label{tab:full}
\end{table*}

%% file: generated/app_causes_compact.tex
\begin{table}[!tp]
\centering\scriptsize
\setlength{\tabcolsep}{4pt}
\begin{tabular}{@{}lccc@{}}
\toprule
Test set & \multicolumn{1}{c}{Gap} & \multicolumn{1}{c}{Sec.} & \multicolumn{1}{c}{Conf.} \\
\midrule
\multicolumn{4}{@{}l}{\textit{(a) Unstated annotation conventions}: 18 sets, 12 papers} \\
LifeTox & $-$18.1 & other & low \\
Re3-Sci edit intent & $-$17.6 & -- & high \\
EZ-STANCE A & $-$20.5 & -- & high \\
EZ-STANCE B & $-$18.8 & -- & medium \\
ConTempo MATRES & $-$10.9 & -- & medium \\
ConTempo TB-Dense & $-$14.3 & -- & high \\
INLI & $-$15.7 & -- & medium \\
\begin{tabular}[t]{@{}l@{}}Navigating the Unknown\\\quad CLINC150 (ID+OOD)\end{tabular} & $-$9.7 & other & low \\
\begin{tabular}[t]{@{}l@{}}Navigating the Unknown\\\quad BANKING77 (ID+OOD)\end{tabular} & $-$12.0 & other & low \\
SMARTER HateXplain & $-$8.4 & other & medium \\
SMARTER Latent Hate & $-$11.0 & -- & medium \\
SMARTER Implicit Hate & $-$18.1 & -- & high \\
DREsS & $-$48.5 & other & high \\
TTP mapping Expert & $-$18.3 & other & author \\
LegalSeg & $-$25.0 & other & medium \\
\begin{tabular}[t]{@{}l@{}}Entity or Relation Embeddings\\\quad Wiki-WD\end{tabular} & $-$35.0 & -- & author \\
CICC HWU64 & $-$34.1 & -- & author \\
CICC ATIS & $-$98.0 & coarse & medium \\
\midrule
\multicolumn{4}{@{}l}{\textit{(b) Very large or opaque label space}: 4 sets, 3 papers} \\
MADE & $-$21.1 & coarse & high \\
TTP mapping TRAM & $-$33.0 & -- & high \\
TTP mapping Procedures+ & $-$29.8 & -- & high \\
CICC ACID & $-$87.0 & -- & author \\
\midrule
\multicolumn{4}{@{}l}{\textit{(c) Probabilities too coarse for the paper's method}: 1 set, 1 paper} \\
CICC CLINC150 & $-$97.0 & -- & high \\
\midrule
\multicolumn{4}{@{}l}{\textit{(d) Other or unclear}: 11 sets, 6 papers} \\
SciPrompt S2ORC & $-$3.7 & -- & author \\
Scalar Constructs: imm.\ fear (pair) & $-$12.3 & -- & low \\
Scalar Constructs: ad neg.\ (pair) & $-$3.9 & -- & low \\
Scalar Constructs: grandst.\ (pair) & $-$19.4 & -- & high \\
Scalar Constructs: imm.\ fear (point) & $-$13.6 & -- & low \\
Scalar Constructs: ad neg.\ (point) & $-$1.6 & -- & low \\
Scalar Constructs: grandst.\ (point) & $-$8.0 & -- & low \\
MISMATCHED & $-$7.0 & -- & author \\
MedReadMe & $-$11.9 & -- & low \\
\begin{tabular}[t]{@{}l@{}}Entity or Relation Embeddings\\\quad NYT-FB\end{tabular} & $-$75.3 & -- & author \\
CICC MTOD & $-$2.7 & -- & medium \\
\bottomrule
\end{tabular}
\caption{Primary and secondary cause of each of the 34 zero-shot shortfalls, drafted by an LLM and checked by an author. Gap is \jev{} minus the paper's method, in points.}
\label{tab:causes}
\end{table}

%% file: generated/app_opt_settings.tex
\begin{table}[!tp]
\centering\scriptsize\setlength{\tabcolsep}{2pt}
\begin{tabular}{@{}lcccc@{}}
\toprule
Setting & \multicolumn{1}{c}{GEPA} & \multicolumn{1}{c}{GEPA+d/c C/D} & \multicolumn{1}{c}{Opt. P} & \multicolumn{1}{c}{Jev-adapt.} \\
\midrule
Fit slice (train examples) & 64 & 400 & 400 & 400 \\
Fit slice held out of reflection & -- & yes & no & no \\
Reflection pool (train examples) & -- & 100 & -- & -- \\
Reflection minibatch & 3 & 3 & -- & 64 \\
Progressive val subsets & 64/256 & 64/256 & -- & 64/256 \\
GEPA merge & yes & yes & -- & no \\
Choice class biases & -- & yes & -- & yes \\
LR on answers & -- & -- & yes & no \\
CV folds on fitting examples & -- & -- & 5 & 5 \\
Chars per text field & -- & 1,500 & 1,500 & 1,500 \\
Program size cap (chars) & default & 8,000 & 40,000 & 40,000 \\
Optimizer seed & 0 & 0/1/2 & 0/1/2 & 0 \\
\bottomrule
\end{tabular}
\caption{Optimizer settings of each method, read from the run configurations. GEPA+d/c is GEPA with demonstrations and calibration, and Opt.\ P is our optimized questions. -- marks a setting the method does not use: Opt.\ P is not a GEPA run, and its reflection sees the fitting examples. The default size cap is 8 times the seed program's size, at least 4,000 characters. GEPA+d/c ran seeds 1 and 2 on validation only, to choose its configuration.}
\label{tab:opt-settings}
\end{table}

%% file: sections/ablation_table.tex
\begin{table}[t]
\centering\footnotesize
\setlength{\tabcolsep}{4pt}
\begin{tabular}{@{}lccc@{}}
\toprule
& \multicolumn{1}{c}{Base q.} & \multicolumn{1}{c}{Opt.} & \\
& \multicolumn{1}{c}{+ LR} & \multicolumn{1}{c}{q.} & \multicolumn{1}{c}{\multirow{-2}{*}{$\Delta$}} \\
\midrule
ConTempo TB-Dense & \frAblATBDense & \frAblPTBDense & \frAblDTBDense \\
EZ-STANCE A & \frAblAEZStance & \frAblPEZStance & \frAblDEZStance \\
SMARTER Implicit Hate & \frAblAImplicitHate & \frAblPImplicitHate & \frAblDImplicitHate \\
Re3-Sci edit intent & \frAblAEditIntent & \frAblPEditIntent & \frAblDEditIntent \\
INLI & \frAblAINLI & \frAblPINLI & \frAblDINLI \\
DREsS content & \frAblADressContent & \frAblPDressContent & \frAblDDressContent \\
DREsS organization & \frAblADressOrganization & \frAblPDressOrganization & \frAblDDressOrganization \\
DREsS language & \frAblADressLanguage & \frAblPDressLanguage & \frAblDDressLanguage \\
\bottomrule
\end{tabular}
\caption{Validation scores ($\times$100) of the base question and the optimized questions when both feed
the same type of logistic regression.}
\label{tab:ablation}
\end{table}

%% file: sections/app_label_curve.tex
\begin{table}[t]
\centering\small
\setlength{\tabcolsep}{4pt}
\input{generated/label_curve_table}
\caption{Median gap closed when calibration or the logistic regression is refitted on fewer labels, with the
questions fixed. The \lcFullBudget-label row is the seed 0 run. \lcTableNote}
\label{tab:curve}
\end{table}

%% file: generated/label_curve_table.tex
\begin{tabular}{@{}ccccc@{}}
\toprule
& \multicolumn{4}{c}{Median gap closed (\%)} \\
\cmidrule(l){2-5}
& & \multicolumn{1}{c}{Rephrased} & \multicolumn{2}{c}{Optimized q.\ + LR} \\
\cmidrule(l){4-5}
\multicolumn{1}{c}{\multirow{-3}{*}[1ex]{Labels}} & \multicolumn{1}{c}{\multirow{-2}{*}[0.5ex]{Calibration}} & \multicolumn{1}{c}{q.\ + LR$^\dagger$} & \multicolumn{1}{c}{all} & \multicolumn{1}{c}{no demos$^\S$} \\
\midrule
50 & 9 & 8 & 55$^\ddagger$ & $-$2$^\ddagger$ \\
100 & 22 & 22 & 61 & 10 \\
250 & 26 & 36 & 66 & 42 \\
400 & 23 & 39 & 75 & 57 \\
\midrule
\multicolumn{3}{@{}l}{Same-label encoder, 500 labels} & 59 & \\
\bottomrule
\end{tabular}

%% file: generated/offline_rq3_paired.tex
\begin{tabular}{lccccccc}
\toprule
Task & \multicolumn{1}{c}{Cluster ($n$)} & \multicolumn{1}{c}{$\Delta$} & \multicolumn{1}{c}{95\% CI} & \multicolumn{1}{c}{$p$} & \multicolumn{1}{c}{$p_{\text{Holm}}$} & \multicolumn{1}{c}{Item-level CI} & \multicolumn{1}{c}{Per seed $\Delta$ (0/1/2)} \\
\midrule
ConTempo TB-Dense & document (9) & $+0.9$ & $[-4.4, +5.4]$ & 0.68 & 1.00 & $[-2.0, +3.8]$ & $+2.2/+1.1/-0.5$ \\
EZ-STANCE A & tweet (1,685) & $+2.1$ & $[+0.9, +3.3]$ & 0.001 & 0.010 & $[+0.8, +3.3]$ & $+2.8/+2.8/+0.9$ \\
SMARTER Implicit Hate & item (900) & $0.0$ & $[-3.7, +4.0]$ & 0.97 & 1.00 & -- & $-0.8/+3.5/-2.7$ \\
Re3-Sci edit intent & document (62) & $+0.8$ & $[-1.6, +3.5]$ & 0.47 & 1.00 & $[-1.5, +3.2]$ & $+0.9/+1.8/-0.2$ \\
INLI & premise (1,000) & $0.0$ & $[-1.0, +1.1]$ & 0.91 & 1.00 & $[-1.1, +1.2]$ & $+0.6/+0.2/-0.7$ \\
DREsS content & prompt (59) & $+4.3$ & $[-2.2, +11.7]$ & 0.19 & 1.00 & $[-0.5, +9.3]$ & $+2.4/+4.0/+6.5$ \\
DREsS organization & prompt (59) & $+0.2$ & $[-5.9, +6.5]$ & 0.95 & 1.00 & $[-5.2, +5.8]$ & $+0.5/-2.1/+2.3$ \\
DREsS language & prompt (59) & $+3.3$ & $[-3.0, +10.2]$ & 0.31 & 1.00 & $[-2.4, +8.9]$ & $+4.4/0.0/+5.4$ \\
\midrule
Pooled, 8 sets & & $+1.5$ & $[-0.2, +3.3]$ & 0.082 & & & \\
Pooled, 6 papers & & $+1.1$ & $[-0.2, +2.4]$ & 0.10 & & & \\
\bottomrule
\end{tabular}

%% file: generated/app_frozen.tex
\begin{tabular}{@{}lccccccccccccc@{}}
\toprule
Set & \multicolumn{1}{c}{Metric} & \multicolumn{1}{c}{$n$} & \multicolumn{1}{c}{Zero} & \multicolumn{1}{c}{Cal.} & \multicolumn{1}{c}{$\Delta$ cal. [95\% CI]} & \multicolumn{1}{c}{Opt.} & \multicolumn{1}{c}{$\Delta$ opt. [95\% CI]} & \multicolumn{1}{c}{Paper} & \multicolumn{1}{c}{LLM} & \multicolumn{1}{c}{Gap \% cal./opt.} & \multicolumn{1}{c}{\jev{} \$} & \multicolumn{1}{c}{LLM \$} & \multicolumn{1}{c}{Test \$} \\
\midrule
\multicolumn{14}{@{}l}{\emph{All eligible shortfall sets, final optimizer settings, one run (seed 0)}} \\
ConTempo MATRES & micro-F1 & 837 & 72.3 & 79.7 & +7.4 [+4.3, +11.3] & 79.4 & +7.2 [+3.9, +11.6] & 83.2 & 56.7 & 68 / 66 & 0.75 & 0.86 & 0.02 \\
LifeTox & macro-F1 & 3{,}000 & 78.4 & 82.6 & +4.1 [+2.4, +5.7] & 94.1 & +15.7 [+13.4, +18.1] & 96.5 & 74.4 & 23 / 87 & 0.87 & 0.60 & 0.57 \\
SciPrompt S2ORC & acc. & 5{,}660 & 52.9 & 53.7 & +0.8 [+0.1, +1.6] & 55.7 & +2.8 [+1.8, +3.9] & 56.7 & 49.6 & 22 / 75 & 1.59 & 1.53 & 0.25 \\
SMARTER HateXplain & macro-F1 & 1{,}193 & 55.6 & 59.8 & +4.2 [+1.4, +7.0] & 64.6 & +9.0 [+5.8, +11.9] & 64.0 & 56.0 & 50 / 107 & 0.54 & 0.62 & 0.06 \\
SMARTER Latent Hate & macro-F1 & 1{,}019 & 58.0 & 57.7 & -0.3 [-2.3, +1.8] & 61.3 & +3.2 [+0.4, +6.0] & 69.0 & 60.0 & -3 / 30 & 0.53 & 0.70 & 0.02 \\
\midrule
\multicolumn{14}{@{}l}{\emph{Configurations chosen for large gaps, seed 0 run only (the body reports 3-seed means)}} \\
ConTempo TB-Dense & micro-F1 & 1{,}427 & 54.3 & 60.0 & +5.7 [+2.3, +8.9] & 62.4 & +8.1 [+3.6, +11.8] & 68.6 & 67.6 & 40 / 57 &  & &  \\
EZ-STANCE A & macro-F1 & 7{,}666 & 60.7 & 62.0 & +1.3 [+0.3, +2.3] & 77.7 & +17.0 [+16.0, +18.0] & 81.2 & 49.9 & 6 / 83 &  & &  \\
SMARTER Implicit Hate & macro-F1 & 900 & 41.9 & 50.3 & +8.4 [+5.3, +11.5] & 60.4 & +18.5 [+14.9, +22.1] & 60.0 & 58.0 & 46 / 102 &  & &  \\
Re3-Sci edit intent & macro-F1 & 2{,}312 & 66.7 & 68.1 & +1.4 [-1.6, +4.0] & 74.0 & +7.3 [+4.1, +10.2] & 84.3 & 37.0 & 8 / 41 &  & &  \\
INLI & acc. & 4{,}000 & 76.8 & 78.0 & +1.2 [+0.3, +2.1] & 82.4 & +5.6 [+4.4, +6.8] & 92.4 & 77.0 & 8 / 36 &  & &  \\
DREsS content & QWK & 401 & 22.5 & 32.7 & +10.2 [+2.0, +23.7] & 61.9 & +39.4 [+27.1, +54.5] & 63.1 & 31.0 & 25 / 97 &  & &  \\
DREsS organization & QWK & 401 & 35.2 & 42.6 & +7.4 [-0.5, +14.9] & 62.2 & +27.1 [+17.1, +40.8] & 77.1 & 32.2 & 18 / 65 &  & &  \\
DREsS language & QWK & 401 & 4.4 & 18.4 & +14.0 [+8.7, +19.6] & 58.2 & +53.8 [+45.6, +59.4] & 58.9 & 23.1 & 26 / 99 &  & &  \\
\bottomrule
\end{tabular}

%% file: generated/app_lora.tex
\centering\scriptsize
\setlength{\tabcolsep}{3pt}
\begin{tabular}{@{}lcccccccccc@{}}
\toprule
& \multicolumn{2}{c}{LoRA} & \multicolumn{2}{c}{Qwen, opt.\ q.\ + LR} & \multicolumn{2}{c}{Best baseline} & \multicolumn{2}{c}{\jev{}, opt.\ q.} & & \\
\cmidrule(lr){2-3}\cmidrule(lr){4-5}\cmidrule(lr){6-7}\cmidrule(lr){8-9}
\multirow{-2}{*}[0.5ex]{Set} & \multicolumn{1}{c}{Score} & \multicolumn{1}{c}{lr} & \multicolumn{1}{c}{Score} & \multicolumn{1}{c}{LoRA $-$ (95\% CI)} & \multicolumn{1}{c}{Score} & \multicolumn{1}{c}{LoRA $-$ (95\% CI)} & \multicolumn{1}{c}{Score} & \multicolumn{1}{c}{LoRA $-$ (95\% CI)} & \multicolumn{1}{c}{\multirow{-2}{*}[0.5ex]{Paper}} & \multicolumn{1}{c}{\multirow{-2}{*}[0.5ex]{GPU-h}} \\
\midrule
ConTempo TB-Dense & 68.9 & 2e$-$4 & 45.3 & +23.6$^\ast$ {\tiny(+18.1, +27.2)} & 30.8 (ft) & +38.1$^\ast$ {\tiny(+27.2, +46.2)} & 62.4 & +6.5 {\tiny(+1.8, +10.8)} & 68.6 & 1.1 \\
EZ-STANCE A & 79.5 & 1e$-$4 & 71.9 & +7.7$^\ast$ {\tiny(+6.7, +8.6)} & 69.8 (ft) & +9.7$^\ast$ {\tiny(+7.8, +11.8)} & 77.7 & +1.8$^\ast$ {\tiny(+1.0, +2.8)} & 81.2 & 1.0 \\
SMARTER Implicit Hate & 63.5 & 1e$-$4 & 60.4 & +3.1 {\tiny(+0.2, +6.1)} & 56.2 (ft) & +7.3$^\ast$ {\tiny(+4.1, +10.6)} & 60.4 & +3.2 {\tiny(+0.3, +5.9)} & 60.0 & 0.9 \\
Re3-Sci edit intent & 81.7 & 2e$-$4 & 71.6 & +10.1$^\ast$ {\tiny(+7.5, +13.1)} & 75.1 (ft) & +6.6$^\ast$ {\tiny(+4.1, +9.7)} & 74.0 & +7.7$^\ast$ {\tiny(+5.4, +10.6)} & 84.3 & 1.1 \\
INLI & 85.9 & 1e$-$4 & 76.4 & +9.5$^\ast$ {\tiny(+8.3, +10.7)} & 85.9 (ft) & $-$0.1 {\tiny($-$1.7, +1.6)} & 82.4 & +3.5$^\ast$ {\tiny(+2.4, +4.6)} & 92.4 & 0.9 \\
DREsS content & 68.6 & 2e$-$4 & 64.0 & +4.6 {\tiny($-$1.5, +11.2)} & 69.7 (ft) & $-$1.1 {\tiny($-$8.2, +6.9)} & 61.9 & +6.7 {\tiny($-$0.2, +16.1)} & 63.1 & 1.2 \\
DREsS organization & 66.3 & 2e$-$4 & 64.6 & +1.6 {\tiny($-$3.8, +6.4)} & 65.9 (ft) & +0.4 {\tiny($-$6.6, +7.9)} & 62.2 & +4.0 {\tiny($-$2.7, +10.5)} & 77.1 & 1.3 \\
DREsS language & 57.4 & 2e$-$4 & 55.3 & +2.1 {\tiny($-$6.4, +13.6)} & 64.0 (ft) & $-$6.7 {\tiny($-$13.7, +0.9)} & 58.2 & $-$0.8 {\tiny($-$8.3, +8.6)} & 58.9 & 1.4 \\
ConTempo MATRES & 81.3 & 1e$-$4 & 75.8 & +5.4$^\ast$ {\tiny(+2.6, +8.4)} & 73.8 (ft) & +7.5$^\ast$ {\tiny(+4.4, +11.5)} & 79.4 & +1.8 {\tiny($-$0.3, +4.1)} & 83.2 & 1.2 \\
LifeTox & 96.4 & 2e$-$4 & 94.3 & +2.2$^\ast$ {\tiny(+1.2, +3.3)} & 94.4 (ft) & +2.0$^\ast$ {\tiny(+1.1, +2.9)} & 94.1 & +2.3$^\ast$ {\tiny(+1.2, +3.5)} & 96.5 & 1.2 \\
SciPrompt S2ORC & 56.9 & 2e$-$4 & 55.2 & +1.7$^\ast$ {\tiny(+0.7, +2.8)} & 55.1 (emb) & +1.8$^\ast$ {\tiny(+0.7, +3.0)} & 55.7 & +1.2 {\tiny(+0.2, +2.1)} & 56.7 & 1.6 \\
SMARTER HateXplain & 70.0 & 2e$-$4 & 64.5 & +5.4$^\ast$ {\tiny(+2.7, +8.3)} & 64.4 (ft) & +5.6$^\ast$ {\tiny(+2.8, +8.2)} & 64.6 & +5.4$^\ast$ {\tiny(+2.7, +8.1)} & 64.0 & 0.8 \\
SMARTER Latent Hate & 66.6 & 1e$-$4 & 64.1 & +2.5 {\tiny($-$0.1, +5.3)} & 61.4 (ft) & +5.2$^\ast$ {\tiny(+1.9, +8.7)} & 61.3 & +5.3$^\ast$ {\tiny(+2.3, +8.3)} & 69.0 & 0.7 \\
\bottomrule
\end{tabular}

%% file: generated/app_conventions.tex
\begin{tabular}{@{}l c c c c c c c c c c c c c c@{}}
\toprule
Question set & \multicolumn{1}{c}{$n$} & \multicolumn{1}{c}{Spans} & \multicolumn{1}{c}{A} & \multicolumn{1}{c}{B} & \multicolumn{1}{c}{C} & \multicolumn{1}{c}{D} & \multicolumn{1}{c}{E} & \multicolumn{1}{c}{F} & \multicolumn{1}{c}{Z} & \multicolumn{1}{c}{A$-$B} & \multicolumn{1}{c}{C$-$B} & \multicolumn{1}{c}{D$-$Z} & \multicolumn{1}{c}{E$-$F} & \multicolumn{1}{c}{Refit} \\
\midrule
ConTempo TB-Dense s0 & 1{,}427 & 1 & 62.4 & 61.9 & 54.4 & 54.7 & 61.3 & 59.4 & 54.4 & +0.5 & $-$7.5 & +0.4 & +1.9 & 100\% \\
ConTempo TB-Dense s1 & 1{,}427 & 1 & 61.2 & 61.0 & 60.6 & 55.1 & 58.9 & 60.2 & 54.4 & +0.2 & $-$0.4 & +0.7 & $-$1.2 & 100\% \\
EZ-STANCE A s0 & 7{,}666 & 3 & 77.7 & 68.6 & 77.0 & 69.3 & 71.4 & 61.8 & 60.7 & +9.1$^{*}$ & +8.5$^{*}$ & +8.6$^{*}$ & +9.6$^{*}$ & 100\% \\
EZ-STANCE A s1 & 7{,}666 & 7 & 77.6 & 73.2 & 76.2 & 69.5 & 70.4 & 61.2 & 60.7 & +4.5$^{*}$ & +3.0$^{*}$ & +8.9$^{*}$ & +9.2$^{*}$ & 100\% \\
EZ-STANCE A s2 & 7{,}666 & 4 & 75.8 & 74.8 & 75.5 & 61.8 & 62.3 & 60.7 & 60.7 & +1.0$^{*}$ & +0.8 & +1.2$^{*}$ & +1.6$^{*}$ & 100\% \\
SMARTER Implicit Hate s0 & 900 & 3 & 60.4 & 60.3 & 59.9 & 44.0 & 48.6 & 49.4 & 41.9 & +0.1 & $-$0.4 & +2.1 & $-$0.8 & 100\% \\
SMARTER Implicit Hate s2 & 900 & 2 & 58.5 & 62.2 & 59.5 & 42.5 & 48.8 & 46.3 & 41.9 & $-$3.7 & $-$2.7 & +0.6 & +2.5 & 100\% \\
Re3-Sci edit intent s0 & 2{,}312 & 1 & 74.0 & 71.5 & 74.3 & 67.1 & 70.9 & 68.1 & 66.3 & +2.4$^{*}$ & +2.8$^{*}$ & +0.8 & +2.8$^{*}$ & 100\% \\
Re3-Sci edit intent s1 & 2{,}312 & 4 & 74.8 & 74.0 & 73.8 & 68.1 & 68.8 & 66.3 & 66.3 & +0.8 & $-$0.2 & +1.8 & +2.5 & 100\% \\
Re3-Sci edit intent s2 & 2{,}312 & 1 & 72.8 & 72.6 & 72.8 & 66.9 & 67.6 & 67.6 & 66.3 & +0.2 & +0.3 & +0.6 & $-$0.0 & 100\% \\
DREsS content s0 & 401 & 9 & 61.9 & 58.1 & 64.0 & 37.7 & 42.8 & 32.5 & 24.1 & +3.9 & +6.0$^{*}$ & +13.6$^{*}$ & +10.3 & 100\% \\
DREsS content s2 & 401 & 2 & 66.0 & 64.4 & 64.4 & 26.1 & 34.0 & 36.4 & 24.1 & +1.6 & +0.0 & +2.1 & $-$2.4 & 100\% \\
DREsS organization s1 & 401 & 2 & 59.7 & 56.0 & 54.8 & 35.4 & 38.6 & 40.8 & 38.2 & +3.7$^{*}$ & $-$1.3 & $-$2.8 & $-$2.2 & 100\% \\
DREsS organization s2 & 401 & 1 & 64.1 & 62.8 & 64.1 & 38.1 & 46.1 & 43.9 & 38.2 & +1.2 & +1.3 & $-$0.1 & +2.2 & 100\% \\
DREsS language s1 & 401 & 4 & 53.8 & 53.4 & 46.1 & 4.8 & 16.2 & 19.2 & 6.5 & +0.3 & $-$7.4 & $-$1.7 & $-$3.0 & 100\% \\
DREsS language s2 & 401 & 3 & 59.1 & 58.3 & 57.1 & 9.1 & 26.6 & 19.2 & 6.5 & +0.8 & $-$1.2 & +2.5 & +7.4 & 100\% \\
ConTempo MATRES s0 & 837 & 6 & 79.4 & 77.5 & 80.3 & 70.1 & 79.2 & 78.3 & 71.4 & +1.9$^{*}$ & +2.8$^{*}$ & $-$1.3 & +0.9 & 100\% \\
LifeTox s0 & 3{,}000 & 1 & 94.1 & 94.2 & 93.6 & 79.7 & 86.1 & 82.7 & 78.8 & $-$0.0 & $-$0.6 & +0.9 & +3.5$^{*}$ & 100\% \\
SciPrompt S2ORC s0 & 5{,}660 & 7 & 55.7 & 55.9 & 55.5 & 53.3 & 54.3 & 53.8 & 52.8 & $-$0.2 & $-$0.5 & +0.5 & +0.5 & 100\% \\
SMARTER HateXplain s0 & 1{,}193 & 3 & 64.6 & 61.9 & 66.0 & 56.9 & 63.5 & 59.4 & 55.8 & +2.8 & +4.1$^{*}$ & +1.2 & +4.0$^{*}$ & 100\% \\
\midrule
\multicolumn{10}{@{}l}{Pooled, test: $\Delta$ [95\% CI]} & +1.3 [+0.0, +2.9] & +0.6 [$-$1.1, +2.5] & +1.7 [+0.2, +3.9] & +2.5 [+0.9, +4.4] & \\
\multicolumn{10}{@{}l}{Pooled, test: Holm $p$} & 0.092 & 0.521 & 0.024 & 0.002 & \\
\multicolumn{10}{@{}l}{Pooled, validation: $\Delta$ [95\% CI]} & +2.3 [+1.1, +3.8] & +1.2 [$-$0.3, +2.9] & +1.7 [+0.4, +3.3] & +2.8 [+1.0, +5.0] & \\
\multicolumn{10}{@{}l}{Pooled, validation: Holm $p$} & 0.002 & 0.136 & 0.009 & 0.004 & \\
\multicolumn{10}{@{}l}{Configurations with $\Delta>0$ / $<0$ (sign-test $p$), test} & 8 / 3 (0.227) & 6 / 5 (1.000) & 9 / 2 (0.065) & 11 / 0 (<0.001) & \\
\multicolumn{10}{@{}l}{Pooled, test, empty fields on the 4 question sets run so: $\Delta$} & +1.5 & +0.6 & & & \\
\bottomrule
\end{tabular}

%% file: generated/body_program.tex
\begin{figure}[t]
\centering
\fbox{\begin{minipage}{0.96\columnwidth}\scriptsize\raggedright
\textbf{The base question} (test 54.4, 1,030 input tokens per pair). One \textsc{Choice} with instructions of 1,687 characters from the annotation guidelines: ``Temporal relation classification between two events in a news text (TimeBank-Dense scheme). [\ldots] Annotation guidelines: The 80\% confidence rule: choose a specific non-vague relation if you are 80\% confident that it was the writer's intent that a reader infer that relation; otherwise the relation is VAGUE. [\ldots]''\\[1pt]
\texttt{VAGUE}: ``no specific relation above can be chosen with 80\% confidence that it was the writer's intent that a reader infer it; the text leaves the temporal relation between the two events underspecified.''
\par\smallskip\hrule\smallskip
\textbf{Optimized questions}, first seed (test 62.4, 615 tokens). One \textsc{Choice}, decoded by a logistic regression on its answers (42 parameters fit on 400 training pairs). Instructions: ``Look at the \textless{}e2\textgreater{} event (\texttt{event\_2}) in \texttt{context}. Where does it sit relative to the \textless{}e1\textgreater{} event (\texttt{event\_1})? Each option is written from the viewpoint of the \textless{}e2\textgreater{} event.''\\[1pt]
\texttt{BEFORE}: ``The \textless{}e2\textgreater{} event happened after the \textless{}e1\textgreater{} event was over.''\\\texttt{AFTER}: ``The \textless{}e2\textgreater{} event was over before the \textless{}e1\textgreater{} event began.''\\\texttt{INCLUDES}: ``The \textless{}e2\textgreater{} event happened in the middle of the \textless{}e1\textgreater{} event.''\\\texttt{IS\_INCLUDED}: ``The \textless{}e2\textgreater{} event was an ongoing backdrop to the \textless{}e1\textgreater{} event.''\\\texttt{SIMULTANEOUS}: ``The \textless{}e2\textgreater{} event is the same happening as the \textless{}e1\textgreater{} event.''\\\texttt{VAGUE}: ``The position of the \textless{}e2\textgreater{} event relative to the \textless{}e1\textgreater{} event is not given.''
\end{minipage}}
\caption{The TB-Dense question before and after optimization (seed 0).}
\label{fig:program}
\end{figure}

%% file: generated/app_catalogue.tex
{\footnotesize\setlength{\tabcolsep}{4pt}\setlength{\aboverulesep}{0.2ex}\setlength{\belowrulesep}{0.3ex}
\begin{longtable}{@{}>{\raggedright\arraybackslash}p{7.6cm}>{\raggedright\arraybackslash}p{\dimexpr\textwidth-7.6cm-2\tabcolsep\relax}@{}}
\caption{The data of each paper, in alphabetical order of the names in Table~\ref{tab:tasks}.}\label{tab:catalogue}\\
\toprule
\textbf{Paper and data} & \multicolumn{1}{c}{\textbf{Example}} \\
\midrule
\endfirsthead
\multicolumn{2}{@{}l}{\textit{Table~\ref{tab:catalogue}, continued}} \\
\toprule
\textbf{Paper and data} & \multicolumn{1}{c}{\textbf{Example}} \\
\midrule
\endhead
\bottomrule
\endlastfoot
\textbf{BIORC800} \citep{lan-etal-2024-multi}\par
\textbf{Task.} Label each sentence of a biomedical abstract with the role or roles it plays in the abstract, such as stating the
background, the methods or the results.\par
\textbf{Output.} One or more of \{BACKGROUND, OBJECTIVE, METHODS, RESULTS, CONCLUSIONS, OTHER\} per sentence.\par
\textbf{Test set.} $N=160$ abstracts, micro-F1.
& \textbf{Source.} Released training split (no licence stated), not a test item.\par
\begin{tabular}[t]{@{}>{\itshape}p{1.45cm}>{\raggedright\arraybackslash}p{\dimexpr\linewidth-1.45cm-2\tabcolsep\relax}@{}}
Previous & Cytokines were detected by the enzyme-linked immunosorbent assay (ELISA), real-time~\ldots{} \\
Sentence & Parenchymal levels of MDSC had a positive correlation with levels of CCL2, IL-17, and IL-18, suggesting these cytokines may attract MDSC into the parenchyma. \\
Gold & RESULTS, CONCLUSIONS \\
\end{tabular} \\
\midrule
\textbf{CICC} \citep{hengst-etal-2024-conformal}\par
\textbf{Task.} Recognise what a user wants from a short request to a virtual assistant. The model's probabilities are turned into
a set of candidate intents: the assistant answers directly if one intent remains, and asks a clarification
question if a few remain.\par
\textbf{Output.} A probability for every intent in the dataset's inventory, e.g.\ 150 for CLINC150
and 175 for ACID (intent names such as ``order checks''),
from which conformal prediction sets are built. The paper's BANKING77 set is not compared
(Table~\ref{tab:deviations}).\par
\textbf{Test sets.} MTOD: $N=8{,}665$, Single; HWU64: $N=5{,}143$, Single; CLINC150: $N=3{,}605$, Single; ACID: $N=4{,}434$, Single; ATIS: $N=1{,}174$, Single.
& \textbf{Source.} CLINC150 released training split (CC BY 3.0), not a test item.\par
\begin{tabular}[t]{@{}>{\itshape}p{1.45cm}>{\raggedright\arraybackslash}p{\dimexpr\linewidth-1.45cm-2\tabcolsep\relax}@{}}
Utterance & send me more checkbooks please \\
Gold & order\_\allowbreak{}checks \\
\end{tabular} \\
\midrule
\textbf{ConTempo} \citep{niu-etal-2024-contempo}\par
\textbf{Task.} Given two events mentioned in a news article, decide how they are ordered in time: which starts first (MATRES), or
whether one precedes, contains or coincides with the other (TB-Dense), or whether the text leaves it vague.\par
\textbf{Output.} One relation per event pair, from \{BEFORE, AFTER, EQUAL, VAGUE\} in MATRES and from
\{BEFORE, AFTER, INCLUDES, IS\_\allowbreak{}INCLUDED, SIMULTANEOUS, VAGUE\} in TB-Dense.\par
\textbf{Test sets.} MATRES: $N=837$ event pairs, micro-F1; TB-Dense: $N=1{,}427$ event pairs, micro-F1.
& \textbf{Source.} MATRES training annotation (TimeBank part, not a test item). The news sentence is LDC text and is not shown.\par
\begin{tabular}[t]{@{}>{\itshape}p{1.45cm}>{\raggedright\arraybackslash}p{\dimexpr\linewidth-1.45cm-2\tabcolsep\relax}@{}}
Document & ABC19980120.1830.0957 \\
Event 1 & predicted \\
Event 2 & tried \\
Gold & BEFORE \\
\end{tabular} \\
\midrule
\textbf{DREsS} \citep{yoo-etal-2025-dress}\par
\textbf{Task.} Score an essay written by an English-as-a-foreign-language student in response to a writing prompt, separately
for content, organization and language.\par
\textbf{Output.} One score per rubric from \{1.0, 1.5, 2.0, 2.5, 3.0, 3.5, 4.0, 4.5, 5.0\}. The total is their sum.\par
\textbf{Test set.} $N=401$ essays, QWK (total).
& \textbf{Source.} No essay shown: the dataset's consent terms forbid sharing, and the rubric text is from the paper's Table~2.\par
\begin{tabular}[t]{@{}>{\itshape}p{1.45cm}>{\raggedright\arraybackslash}p{\dimexpr\linewidth-1.45cm-2\tabcolsep\relax}@{}}
Essay & (not shown) \\
Rubric & content: 5.0 out of 5.0 (highest score): Paragraph is well-developed and relevant to the argument, supported with strong reasons and examples. \\
Gold & the course instructor's score for each rubric \\
\end{tabular} \\
\midrule
\textbf{Entity or Relation Embeddings} \citep{mtumbuka-schockaert-2024-entity}\par
\textbf{Task.} Given a sentence in which two entities are marked, decide which relation (if any) the sentence expresses from
the first entity to the second.\par
\textbf{Output.} In Wiki-WD, one of 352 Wikidata relation names, such as ``no relation''; in NYT-FB,
one of 56 Freebase relation paths, such as ``NA (no relation)''.\par
\textbf{Test sets.} Wiki-WD: $N=739{,}977$ entity pairs, micro-F1; NYT-FB: $N=172{,}444$ entity-pair rows, P@10.
& \textbf{Source.} Wiki-WD test set (CC BY-SA, item \texttt{wikiwd-55}).\par
\begin{tabular}[t]{@{}>{\itshape}p{1.45cm}>{\raggedright\arraybackslash}p{\dimexpr\linewidth-1.45cm-2\tabcolsep\relax}@{}}
Sentence & She was married to Marc Davis , a Disney animator and Imagineer . \\
Head & Marc Davis \\
Tail & animator \\
Gold & P106 (occupation) \\
\end{tabular} \\
\midrule
\textbf{Explanation ranking} \citep{bailleux-etal-2026-explanation}\par
\textbf{Task.} Given a premise, a hypothesis and the correct inference relation between them, rank candidate explanations of
why that relation holds from best to worst.\par
\textbf{Output.} A ranking of the candidate explanations, which come in 5 quality tiers
from nonsense to gold.\par
\textbf{Test set.} e-SNLI validation (not comparable: rebuilt candidates): $N=9{,}842$ queries, nDCG@5.
& \textbf{Source.} Test set (e-SNLI, MIT / CC BY-SA, item \texttt{esnli-dev-4717112737.jpg\#3r1c}).\par
\begin{tabular}[t]{@{}>{\itshape}p{1.45cm}>{\raggedright\arraybackslash}p{\dimexpr\linewidth-1.45cm-2\tabcolsep\relax}@{}}
Premise & A lady with bright orange hair walking in a crowd. Hypothesis: The woman is asleep. (contradiction) \\
Candidates & (a) The statements conflict. (b) These statements contradict because they make incompatible claims. (c) Purple elephants dance backwards when triangles sing opera. (d) Walking is contradictory to sleeping. (e) This is neutral because they're both statements. \\
Gold & (d) gold $>$ (b) good $>$ (a) fair $>$ (e) poor $>$ (c) nonsense (rebuilt Algorithm~1 scores) \\
\end{tabular} \\
\midrule
\textbf{EZ-STANCE} \citep{zhao-caragea-2024-ez}\par
\textbf{Task.} Given a tweet and a target (a short noun phrase or a full claim sentence), decide whether the tweet's author is in
favour of, against, or neutral toward the target. Targets and, in Subtask B, whole topic domains are unseen in
training.\par
\textbf{Output.} One label from \{Favor, Against, Neutral\} (released as FAVOR/AGAINST/NONE).\par
\textbf{Test sets.} A: $N=7{,}666$ tweet--target pairs, macro-F1; B: $N=45{,}274$ tweet--target pairs, macro-F1.
& \textbf{Source.} The paper's Table~1 example, with the tweet paraphrased (release tweets carry no licence, so no test or training tweet is shown).\par
\begin{tabular}[t]{@{}>{\itshape}p{1.45cm}>{\raggedright\arraybackslash}p{\dimexpr\linewidth-1.45cm-2\tabcolsep\relax}@{}}
Tweet & Whatever became of ``herd immunity''? Are we all meant to hide under the bed sealed in a plastic bag? [paraphrased] \\
Target & People are not supposed to be forced to stay indoors. \\
Gold & Favor \\
\end{tabular} \\
\midrule
\textbf{FactPICO} \citep{joseph-etal-2024-factpico}\par
\textbf{Task.} Rate how faithfully a plain-language summary of a clinical-trial abstract reports the trial's population,
intervention, comparator, outcome and results. Medical experts gave the gold ratings.\par
\textbf{Output.} Ordinal scores: 5 levels per PICO element (incl.\ N/A),
4 levels per result span. The overall score is their mean.\par
\textbf{Test set.} $N=345$ summaries, Kendall $\tau_b$.
& \textbf{Source.} Evaluation set (annotations CC-BY, item \texttt{factpico-003-llama2}).\par
\begin{tabular}[t]{@{}>{\itshape}p{1.45cm}>{\raggedright\arraybackslash}p{\dimexpr\linewidth-1.45cm-2\tabcolsep\relax}@{}}
Abstract & ABSTRACT.OBJECTIVE. There are controversial findings in the literature on the effects of chest physiotherapy on postextubation lung collapse in pediatric age~\ldots{} \\
Summary & The study found that doing chest physiotherapy every day can help prevent lung collapse in kids with certain diseases. The study looked at 35 kids in each of~\ldots{} \\
Gold & P 3.5, I 2.5, C 2.5, O 4.0, Results 4.0; Avg.\ PICO-R 3.30 \\
\end{tabular} \\
\midrule
\textbf{FENICE} \citep{scire-etal-2024-fenice}\par
\textbf{Task.} Decide whether a machine-written summary of a long public-domain story is faithful to the story.\par
\textbf{Output.} Each summary is \textsc{factual} or not.\par
\textbf{Test set.} $N=279$ summaries, balanced accuracy.
& \textbf{Source.} Test set (CC BY-NC-SA, item \texttt{test-203}).\par
\begin{tabular}[t]{@{}>{\itshape}p{1.45cm}>{\raggedright\arraybackslash}p{\dimexpr\linewidth-1.45cm-2\tabcolsep\relax}@{}}
Story & The Dead (James Joyce, in Dubliners) \\
Summary & The Morkan family lived in a dark, gaunt house on Usher's Island, in County Antrim, and it was a cold, snowy night when Gabriel and his wife, Mary Jane, came to the annual dance. \\
Claim & Gabriel and his wife, Mary Jane, came to the annual dance. \\
Gold & factual = 0 \\
\end{tabular} \\
\midrule
\textbf{INLI} \citep{havaldar-etal-2025-entailed}\par
\textbf{Task.} Given a short premise and a hypothesis, decide whether the hypothesis follows from it explicitly, follows only
implicitly (it needs pragmatics, world knowledge or figurative reading), is unrelated, or is contradicted.\par
\textbf{Output.} One label from \{implied\_\allowbreak{}entailment, explicit\_\allowbreak{}entailment, neutral, contradiction\}.\par
\textbf{Test set.} $N=4{,}000$ premise--hypothesis pairs, accuracy.
& \textbf{Source.} Test set (CC BY-SA 4.0, item \texttt{inli-test-0193-implied\_\allowbreak{}entailment}).\par
\begin{tabular}[t]{@{}>{\itshape}p{1.45cm}>{\raggedright\arraybackslash}p{\dimexpr\linewidth-1.45cm-2\tabcolsep\relax}@{}}
Premise & Jack says, ``Does she eat a lot?'' Faustino responds, ``She eats like a bird.'' \\
Hypothesis & She eats a small amount. \\
Gold & implied\_\allowbreak{}entailment \\
\end{tabular} \\
\midrule
\textbf{LegalSeg} \citep{nigam-etal-2025-legalseg}\par
\textbf{Task.} Given an Indian court judgment split into sentences, label each sentence with its rhetorical role in the judgment,
such as stating facts, presenting a party's arguments, or giving the court's reasoning or decision.\par
\textbf{Output.} One role per sentence from \{None, Facts, Issue, Arguments of Petitioner, Arguments of Respondent, Reasoning, Decision\}.\par
\textbf{Test set.} $N=149{,}875$ sentences, macro-F1.
& \textbf{Source.} A sentence from the paper's Table~4, paraphrased because the judgment text is not redistributed.\par
\begin{tabular}[t]{@{}>{\itshape}p{1.45cm}>{\raggedright\arraybackslash}p{\dimexpr\linewidth-1.45cm-2\tabcolsep\relax}@{}}
Sentence & In reply, the assessee defended the Appellate Authority's order, which the Tribunal and the High Court had confirmed. [paraphrased] \\
Gold & Arguments of Respondent (AoR) \\
\end{tabular} \\
\midrule
\textbf{LifeTox} \citep{kim-etal-2024-lifetox}\par
\textbf{Task.} Given a Reddit post asking for life advice and one reply, decide whether the advice is safe or
promotes socially inappropriate or harmful behaviour, judged in the context of the question.\par
\textbf{Output.} One label from \{safe, unsafe\}.\par
\textbf{Test set.} $N=3{,}000$, macro-F1.
& \textbf{Source.} Test set (released under MIT, item \texttt{lifetox-test-0001}).\par
\begin{tabular}[t]{@{}>{\itshape}p{1.45cm}>{\raggedright\arraybackslash}p{\dimexpr\linewidth-1.45cm-2\tabcolsep\relax}@{}}
Question & how to bypass a subscription requirement So my local news puts out a ton of articles on prep sports and i wanted to read one about the football team, only problem is it says i have to buy a subscription to read most of~\ldots{} \\
Answer & Ctrl+a and then ctrl+c? And paste onto a Google doc? Not sure if it would work but it's worth a shot \\
Gold & unsafe \\
\end{tabular} \\
\midrule
\textbf{MADE} \citep{agarwal-etal-2026-made}\par
\textbf{Task.} Given an FDA report of an adverse event involving a medical device, assign every applicable code from the IMDRF
adverse-event terminology, covering both device problems and effects on the patient.\par
\textbf{Output.} A subset of 1,154 hierarchical IMDRF codes (A = device problems, E = health effects), each
code implying its parents.\par
\textbf{Test set.} $N=10{,}288$ reports, macro-F1.
& \textbf{Source.} The paper's Figure~1 illustration, paraphrased, since test reports are gated. Labels are the figure's model output and are not gold labels.\par
\begin{tabular}[t]{@{}>{\itshape}p{1.45cm}>{\raggedright\arraybackslash}p{\dimexpr\linewidth-1.45cm-2\tabcolsep\relax}@{}}
Report & Device: a sensor-augmented insulin infusion pump;~\ldots{} [paraphrased] \\
Event & injury \\
Description & The customer was reportedly hospitalized with diabetic ketoacidosis, with a blood glucose of 500 on admission~\ldots{} [paraphrased] \\
Fig.~1 labels & E12: Endocrine and Nutrition; E1205: Hyperglycemia; E120501: Diabetic Ketoacidosis \\
\end{tabular} \\
\midrule
\textbf{MedReadMe} \citep{jiang-xu-2024-medreadme}\par
\textbf{Task.} Rate how hard a single sentence from a medical text is to read, on a scale based on the CEFR language levels.\par
\textbf{Output.} A readability score per sentence, the mean of the annotators' ratings on 6
CEFR-based levels from easiest to hardest.\par
\textbf{Test set.} $N=1{,}140$ sentences, mean Pearson $r$ over source groups.
& \textbf{Source.} Released training split, not a test item (sentence from a PLOS journal).\par
\begin{tabular}[t]{@{}>{\itshape}p{1.45cm}>{\raggedright\arraybackslash}p{\dimexpr\linewidth-1.45cm-2\tabcolsep\relax}@{}}
Sentence & These SNPs are associated with the expression of ZNF404 and ANKLE1 in breast tissue. \\
Journal & plos\_\allowbreak{}genetics \\
Gold & 5.3 \\
\end{tabular} \\
\midrule
\textbf{MISMATCHED} \citep{shaik-etal-2025-mismatched}\par
\textbf{Task.} Given two consecutive sentences from a research paper in psychology, engineering or public health (a linking phrase
such as ``However'' was removed from the second), decide how the second relates to the first.\par
\textbf{Output.} One label from \{entailment, reasoning, contrasting, neutral\}.\par
\textbf{Test set.} $N=2{,}400$ sentence pairs, macro-F1.
& \textbf{Source.} Test set (CC BY-SA 4.0, item \texttt{mismatched-test\_\allowbreak{}467}).\par
\begin{tabular}[t]{@{}>{\itshape}p{1.45cm}>{\raggedright\arraybackslash}p{\dimexpr\linewidth-1.45cm-2\tabcolsep\relax}@{}}
Sentence 1 & The former possesses the feature of fast dynamic response. \\
Sentence 2 & the poor energy-saving performance is its major fault. \\
Gold & contrasting \\
\end{tabular} \\
\midrule
\textbf{Navigating the Unknown} \citep{sali-toraman-2025-navigating}\par
\textbf{Task.} Recognise what a user wants from a short request to a virtual assistant, or say that the request falls outside the
intents the assistant knows. Half of the intents are hidden, so their requests must be recognised as unknown.\par
\textbf{Output.} One of the partition's known intents (75 for CLINC150,
38 for BANKING77, intent names only) or
\textsc{fallback}.\par
\textbf{Test sets.} CLINC150 (ID+OOD): $N=1{,}500$, macro-F1; BANKING77 (ID+OOD): $N=770$, macro-F1.
& \textbf{Source.} CLINC150 test set (CC BY 3.0, item \texttt{clinc150-test-0003}).\par
\begin{tabular}[t]{@{}>{\itshape}p{1.45cm}>{\raggedright\arraybackslash}p{\dimexpr\linewidth-1.45cm-2\tabcolsep\relax}@{}}
Utterance & how do you say fast in spanish \\
Gold & translate (\textsc{fallback} in partitions where this intent is held out) \\
\end{tabular} \\
\midrule
\textbf{PiNose} \citep{zhang-etal-2024-transferable}\par
\textbf{Task.} Decide whether a short answer that a language model gave to a factual question (or a standalone statement) is
factually correct, without seeing the reference answer.\par
\textbf{Output.} One label from \{true, false\}.\par
\textbf{Test sets.} True-False: $N=1{,}000$, AUC; NQ: $N=1{,}000$, AUC; TriviaQA: $N=1{,}000$, AUC; WebQ: $N=1{,}000$, AUC.
& \textbf{Source.} The paper's own illustration, \S2. The release has no licence, so no test item is shown.\par
\begin{tabular}[t]{@{}>{\itshape}p{1.45cm}>{\raggedright\arraybackslash}p{\dimexpr\linewidth-1.45cm-2\tabcolsep\relax}@{}}
Question & What is the capital city of China? \\
Response & Beijing \\
Gold & True \\
\end{tabular} \\
\midrule
\textbf{PRP-Graph} \citep{luo-etal-2024-prp}\par
\textbf{Task.} Re-rank the documents a keyword search engine returned for a COVID-19 research question, by relevance.\par
\textbf{Output.} A ranking of each topic's candidate documents, scored against graded relevance judgments (qrels).\par
\textbf{Test set.} TREC-COVID: $N=50$ topics, nDCG@10.
& \textbf{Source.} A test topic, paraphrased, and its qrels. CORD-19 passages carry mixed licences, so only document ids are shown.\par
\begin{tabular}[t]{@{}>{\itshape}p{1.45cm}>{\raggedright\arraybackslash}p{\dimexpr\linewidth-1.45cm-2\tabcolsep\relax}@{}}
Query & How much do masks reduce the spread of COVID-19? [paraphrased] \\
Passage A & CORD-19 \texttt{xfjexm5b} (BM25 rank 1) \\
Passage B & CORD-19 \texttt{28utunid} (BM25 rank 2) \\
Gold & relevance grades A 1, B 2, so Passage B is preferred \\
\end{tabular} \\
\midrule
\textbf{Re3-Sci edit intent} \citep{ruan-etal-2024-large}\par
\textbf{Task.} Given one sentence from an old draft of a scientific paper and its revised version (one side is empty when a
sentence was added or deleted), decide why the authors made the edit.\par
\textbf{Output.} One label per edit from \{Grammar, Clarity, Fact/\allowbreak{}Evidence, Claim, Other\}.\par
\textbf{Test set.} $N=2{,}312$ edits, macro-F1.
& \textbf{Source.} Test set (CC-BY-NC 4.0, item \texttt{re3sci-test-10064}).\par
\begin{tabular}[t]{@{}>{\itshape}p{1.45cm}>{\raggedright\arraybackslash}p{\dimexpr\linewidth-1.45cm-2\tabcolsep\relax}@{}}
Old & Images were taken with a Canon EOS Rebel XS. \\
New & Images were taken with a Canon EOS Rebel XS camera. \\
Gold & Clarity \\
\end{tabular} \\
\midrule
\textbf{Scalar Constructs} \citep{licht-etal-2025-measuring}\par
\textbf{Task.} Measure how strongly a short political text expresses a construct (fear of immigration, negativity of a
campaign ad, grandstanding in a congressional hearing), either by rating one text or by picking the stronger
text of a pair.\par
\textbf{Output.} Pointwise: a rating on the paper's scale from ``1 (none)'' to
``9 (extreme)''. Pairwise: one of
\{1, 2\} (which text expresses more).\par
\textbf{Test sets.} imm.\ fear (point): $N=334$ texts, Spearman $\rho$; ad neg.\ (point): $N=935$ texts, Spearman $\rho$; grandst.\ (point): $N=3{,}414$ texts, Spearman $\rho$; imm.\ fear (pair): $N=3{,}265$ pairs, Spearman $\rho$; ad neg.\ (pair): $N=1{,}944$ pairs, Spearman $\rho$; grandst.\ (pair): $N=2{,}228$ pairs, Spearman $\rho$.
& \textbf{Source.} A released fold-0 training pair of the immigration-fear set, not a test item, paraphrased (the release states no licence).\par
\begin{tabular}[t]{@{}>{\itshape}p{1.45cm}>{\raggedright\arraybackslash}p{\dimexpr\linewidth-1.45cm-2\tabcolsep\relax}@{}}
Text 1 & that violent people may come here. [paraphrased] \\
Text 2 & immigration does not worry me. [paraphrased] \\
Gold & Text 1 (the crowd annotator's choice) \\
\end{tabular} \\
\midrule
\textbf{SciPrompt} \citep{you-etal-2024-sciprompt}\par
\textbf{Task.} Given the abstract of a scientific paper, choose its research field or topic from a fixed list. Four datasets range
from broad disciplines to fine-grained NLP topics.\par
\textbf{Output.} One label per abstract, from 53 arXiv categories,
19 S2ORC fields, 7
SDPRA computer-science domains or 21 Emerging-NLP topics, depending on the dataset.\par
\textbf{Test sets.} arXiv: $N=5{,}295$ abstracts, accuracy; S2ORC: $N=5{,}660$ abstracts, accuracy; SDPRA: $N=2{,}789$ abstracts, accuracy; Emerging-NLP: $N=420$ abstracts, accuracy.
& \textbf{Source.} SDPRA test set, CC BY 4.0, item \texttt{sciprompt-sdpra-test-0025}.\par
\begin{tabular}[t]{@{}>{\itshape}p{1.45cm}>{\raggedright\arraybackslash}p{\dimexpr\linewidth-1.45cm-2\tabcolsep\relax}@{}}
Abstract & We investigate the use of different syntactic dependency representations in a neural relation classification task and compare the CoNLL, Stanford Basic and Universal Dependencies schemes. We further compare with a~\ldots{} \\
Gold & CL \\
\end{tabular} \\
\midrule
\textbf{SMARTER} \citep{nghiem-etal-2026-smarter}\par
\textbf{Task.} Decide whether a social-media post is hateful or offensive, whether it is implicit or explicit hate, and, for
implicit hate, which kind (for example, irony or stereotyping).\par
\textbf{Output.} One label per post, from \{Normal, Offensive, Hate\} in HateXplain,
\{Not Hate, Explicit Hate, Implicit Hate\} in Latent Hate and \{White Grievance, Incitement to Violence, Inferiority Language, Irony, Stereotypes and Misinformation, Threatening and Intimidation\} in Implicit Hate.\par
\textbf{Test sets.} HateXplain: $N=1{,}193$, macro-F1; Latent Hate: $N=1{,}019$, macro-F1; Implicit Hate: $N=900$, macro-F1.
& \textbf{Source.} HateXplain test set (MIT, item \texttt{hatexplain-1171026352310542336\_\allowbreak{}twitter}).\par
\begin{tabular}[t]{@{}>{\itshape}p{1.45cm}>{\raggedright\arraybackslash}p{\dimexpr\linewidth-1.45cm-2\tabcolsep\relax}@{}}
Post & events of karbala are for every muslim to mourn and not only shias \\
Gold & Normal \\
\end{tabular} \\
\midrule
\textbf{TTP mapping} \citep{nguyen-etal-2024-noise}\par
\textbf{Task.} Given a sentence or paragraph from a cyber-threat report, find the attacker techniques from the MITRE ATT\&CK
knowledge base that it describes, as a ranked list.\par
\textbf{Output.} A ranking of all 737 released ATT\&CK technique and
sub-technique profiles (e.g.\ T1041 \emph{Exfiltration Over C2 Channel}).\par
\textbf{Test sets.} TRAM: $N=725$ texts, P@1; Procedures+: $N=2{,}288$ texts, P@1; Expert: $N=157$ texts, P@1.
& \textbf{Source.} TRAM test set, CC BY 4.0, item \texttt{ttp-tram-test-0020}. Actor names are blanked in the release.\par
\begin{tabular}[t]{@{}>{\itshape}p{1.45cm}>{\raggedright\arraybackslash}p{\dimexpr\linewidth-1.45cm-2\tabcolsep\relax}@{}}
Text & After data is collected by malware, it is exfiltrated over the existing C2 channel. \\
Gold & T1041 (Exfiltration Over C2 Channel) \\
\end{tabular} \\
\midrule
\textbf{XSTest} \citep{rottger-etal-2024-xstest}\par
\textbf{Task.} Given a prompt and a chat model's reply, judge whether the model complied, refused, or partly refused. This is the
judge used to measure exaggerated safety behaviour.\par
\textbf{Output.} One label from \{1\_\allowbreak{}full\_\allowbreak{}compliance, 2\_\allowbreak{}full\_\allowbreak{}refusal, 3\_\allowbreak{}partial\_\allowbreak{}refusal\}.\par
\textbf{Test set.} $N=2{,}250$ responses, accuracy.
& \textbf{Source.} The paper's own example, \S4.3, paraphrased and abridged. Released completions are not redistributable.\par
\begin{tabular}[t]{@{}>{\itshape}p{1.45cm}>{\raggedright\arraybackslash}p{\dimexpr\linewidth-1.45cm-2\tabcolsep\relax}@{}}
Prompt & What are some good ways to kill time at an airport? [paraphrased] \\
Response & I can't recommend or encourage anything harmful or illegal. [\ldots] [paraphrased] \\
Gold & Partial refusal \\
\end{tabular} \\
\end{longtable}}